\pdfoutput=1
\documentclass[11pt]{article}

\usepackage[utf8]{inputenc}
\usepackage[T1]{fontenc}
\usepackage{hyphenat}
\usepackage{xspace}
\usepackage{amsmath}
\usepackage{amsfonts}
\usepackage{hyperref}
\usepackage{url}
\usepackage{booktabs}
\usepackage{multirow}
\usepackage{makecell}
\usepackage{caption}
\usepackage{minibox}
\usepackage{bbm}
\usepackage{graphicx}
\usepackage{balance}
\usepackage{mathtools}
\usepackage{color}
\usepackage{marvosym}
\usepackage{ifthen}
\usepackage{textcomp}
\usepackage{enumitem}
\usepackage{verbatim}
\usepackage{algorithm}
\usepackage{numprint}
\usepackage{balance}

\usepackage{amsthm}
\theoremstyle{plain}

\newcommand{\chatoDisplayMode}[1]{#1}

\definecolor{MyRed}{rgb}{0.6,0.0,0.0}
\definecolor{MyBlack}{rgb}{0.1,0.1,0.1}
\newcommand{\inred}[1]{{\color{MyRed}\sf\textbf{\textsc{#1}}}}
\newcommand{\frameit}[2]{
  \begin{center}
  {\color{MyRed}
  \framebox[.9\columnwidth][l]{
    \begin{minipage}{.85\columnwidth}
    \inred{#1}: {\sf\color{MyBlack}#2}
    \end{minipage}
  }\\
  }
  \end{center}
}

\newcommand{\note}[2][]{\chatoDisplayMode{\def\@tmpsig{#1}\frameit{{\Pointinghand} Note}{#2\ifx \@tmpsig \@empty \else \mbox{ --\em #1}\fi}}}
\newcommand{\todo}[2][]{\chatoDisplayMode{\def\@tmpsig{#1}\frameit{{\Writinghand} To-do}{#2\ifx \@tmpsig \@empty \else \mbox{ --\em #1}\fi}}}

\newcommand{\abbrevStyle}[1]{#1}

\usepackage{amssymb}% http://ctan.org/pkg/amssymb
\usepackage{pifont}% http://ctan.org/pkg/pifont

\newcommand{\ie}{\abbrevStyle{i.e.}\xspace}
\newcommand{\eg}{\abbrevStyle{e.g.}\xspace}
\newcommand{\cf}{\abbrevStyle{cf.}\xspace}

\newcommand{\etc}{\abbrevStyle{etc.}\xspace}

\newcommand{\Secref}[1]{Sec.~\ref{#1}}

\newcommand{\Tabref}[1]{Table~\ref{#1}}
\newcommand{\Figref}[1]{Fig.~\ref{#1}}

\newcommand{\Appref}[1]{Appendix~\ref{#1}}

\newcommand{\xhdr}[1]{\vspace{1.7mm}\noindent{{\bf #1.}}}

\newcommand{\xhdrIt}[1]{\vspace{1.7mm}\noindent{{\it #1.}}}

\newcommand{\denselist}{ \itemsep -2pt\topsep-10pt\partopsep-10pt }

\newcommand{\textcite}[1]{\citeauthor{#1} \shortcite{#1}}

\newcommand{\hide}[1]{}

\makeatletter
\newcommand{\iffont}[2]{\ifthenelse{\equal{\f@family}{#1}}{#2}{}}
\makeatother

  \usepackage{mathptmx}

  \DeclareSymbolFont{greek}{OML}{cmm}{m}{n}
  \DeclareMathSymbol{\alpha}{\mathalpha}{greek}{"0B}
  \DeclareMathSymbol{\beta}{\mathalpha}{greek}{"0C}
  \DeclareMathSymbol{\gamma}{\mathalpha}{greek}{"0D}
  \DeclareMathSymbol{\delta}{\mathalpha}{greek}{"0E}
  \DeclareMathSymbol{\epsilon}{\mathalpha}{greek}{"0F}
  \DeclareMathSymbol{\zeta}{\mathalpha}{greek}{"10}
  \DeclareMathSymbol{\eta}{\mathalpha}{greek}{"11}
  \DeclareMathSymbol{\theta}{\mathalpha}{greek}{"12}
  \DeclareMathSymbol{\iota}{\mathalpha}{greek}{"13}
  \DeclareMathSymbol{\kappa}{\mathalpha}{greek}{"14}
  \DeclareMathSymbol{\lambda}{\mathalpha}{greek}{"15}
  \DeclareMathSymbol{\mu}{\mathalpha}{greek}{"16}
  \DeclareMathSymbol{\nu}{\mathalpha}{greek}{"17}
  \DeclareMathSymbol{\xi}{\mathalpha}{greek}{"18}
  \DeclareMathSymbol{\pi}{\mathalpha}{greek}{"19}
  \DeclareMathSymbol{\rho}{\mathalpha}{greek}{"1A}
  \DeclareMathSymbol{\sigma}{\mathalpha}{greek}{"1B}
  \DeclareMathSymbol{\tau}{\mathalpha}{greek}{"1C}
  \DeclareMathSymbol{\upsilon}{\mathalpha}{greek}{"1D}
  \DeclareMathSymbol{\phi}{\mathalpha}{greek}{"1E}
  \DeclareMathSymbol{\chi}{\mathalpha}{greek}{"1F}
  \DeclareMathSymbol{\psi}{\mathalpha}{greek}{"20}
  \DeclareMathSymbol{\omega}{\mathalpha}{greek}{"21}
  \DeclareMathSymbol{\varepsilon}{\mathalpha}{greek}{"22}
  \DeclareMathSymbol{\vartheta}{\mathalpha}{greek}{"23}
  \DeclareMathSymbol{\varpi}{\mathalpha}{greek}{"24}
  \DeclareMathSymbol{\varrho}{\mathalpha}{greek}{"25}
  \DeclareMathSymbol{\varsigma}{\mathalpha}{greek}{"26}
  \DeclareMathSymbol{\varphi}{\mathalpha}{greek}{"27}
  \DeclareSymbolFont{otone}{OT1}{cmr}{m}{n}
  \DeclareMathSymbol{\Gamma}{\mathalpha}{otone}{0}
  \DeclareMathSymbol{\Delta}{\mathalpha}{otone}{1}
  \DeclareMathSymbol{\Theta}{\mathalpha}{otone}{2}
  \DeclareMathSymbol{\Lambda}{\mathalpha}{otone}{3}
  \DeclareMathSymbol{\Xi}{\mathalpha}{otone}{4}
  \DeclareMathSymbol{\Pi}{\mathalpha}{otone}{5}
  \DeclareMathSymbol{\Sigma}{\mathalpha}{otone}{6}
  \DeclareMathSymbol{\Upsilon}{\mathalpha}{otone}{7}
  \DeclareMathSymbol{\Phi}{\mathalpha}{otone}{8}
  \DeclareMathSymbol{\Psi}{\mathalpha}{otone}{9}
  \DeclareMathSymbol{\Omega}{\mathalpha}{otone}{10}
  \DeclareSymbolFont{syms}{OML}{cmm}{m}{it}
  \DeclareMathSymbol{\partial}{\mathord}{syms}{"40}
  \DeclareMathAlphabet{\mathbold}{OML}{cmm}{b}{it}
  \DeclareSymbolFont{largesymbols}{OMX}{cmex}{m}{n}

  \DeclareMathAlphabet{\mathcal}{OMS}{cmsy}{m}{n}

\usepackage{ACL}

\usepackage{multicol}
\usepackage{epigraph}

\usepackage[utf8]{inputenc}

\usepackage{times}
\usepackage{latexsym}
\usepackage{amssymb}
\usepackage{multicol}
\usepackage{array}
\usepackage{xcolor} % Required for text coloring
\usepackage{subcaption}

\usepackage{listings}
\usepackage{CJKutf8} % Use this package to support UTF-8 encoding with CJK
\newcommand{\zh}[1]{\begin{CJK}{UTF8}{gbsn}#1\end{CJK}}

\usepackage{microtype}

\usepackage{inconsolata}

\newcommand{\llama}{Llama-2\xspace}

\newcommand{\lang}[1]{\text{\textsc{\MakeLowercase{#1}}}}

\newcommand{\prompt}[1]{
\vspace{2mm}
\fbox{
\begin{minipage}{.85\linewidth}
\small
#1
\end{minipage}
}
\vspace{2mm}
}

\title{
Do Llamas Work in English?\\
On the Latent Language of Multilingual Transformers
}

\author{
Chris Wendler\thefootnote{*},
Veniamin Veselovsky\thefootnote{*},
Giovanni Monea\thefootnote{*},
Robert West\thefootnote{*} \\
EPFL \\
{\{chris.wendler, veniamin.veselovsky, giovanni.monea, robert.west\}@epfl.ch}
}

\begin{document}
    
\maketitle

\def\thefootnote{*}\footnotetext{Equal contribution.\\ }
\def\thefootnote{\arabic{footnote}}

\begin{abstract}

We ask whether multilingual language models trained on unbalanced, English\hyp dominated corpora use English as an internal pivot language---a question of key importance for understanding how language models function and the origins of linguistic bias.
Focusing on the \llama family of transformer models, our study uses carefully constructed non\hyp English prompts with a unique correct single\hyp token continuation.
From layer to layer, transformers gradually map an input embedding of the final prompt token to an output embedding from which next\hyp token probabilities are computed.
Tracking intermediate embeddings through their high\hyp dimensional space reveals three distinct phases, whereby intermediate embeddings
(1)~start far away from output token embeddings;
(2)~already allow for decoding a semantically correct next token in middle layers, but give higher probability to its version in English than in the input language;
(3)~finally move into an input\hyp language\hyp specific region of the embedding space.
We cast these results into a conceptual model where the three phases operate in ``input space'', ``concept space'', and ``output space'', respectively.
Crucially, our evidence suggests that the abstract ``concept space'' lies closer to English than to other languages, which may have important consequences regarding the biases held by multilingual language models. Code and data is made available here: \url{https://github.com/epfl-dlab/llm-latent-language}.
\end{abstract}

\section{Introduction}

Most modern large language models (LLMs) are trained on massive corpora of mostly English text
% In the GPT-4 report, it's tucked away in footnote 27: "The majority of pretraining data and our alignment data is in English."
\cite{touvron2023llama,openai2023gpt4}.
Despite this, they achieve strong performance on a broad range of downstream tasks, even in non\hyp English languages \cite{shi2022language}. This raises a compelling question: How are LLMs able to generalize so well from their mainly English training 
data to other languages?

\begin{figure}[t]
    \centering
    \includegraphics[width=\linewidth]{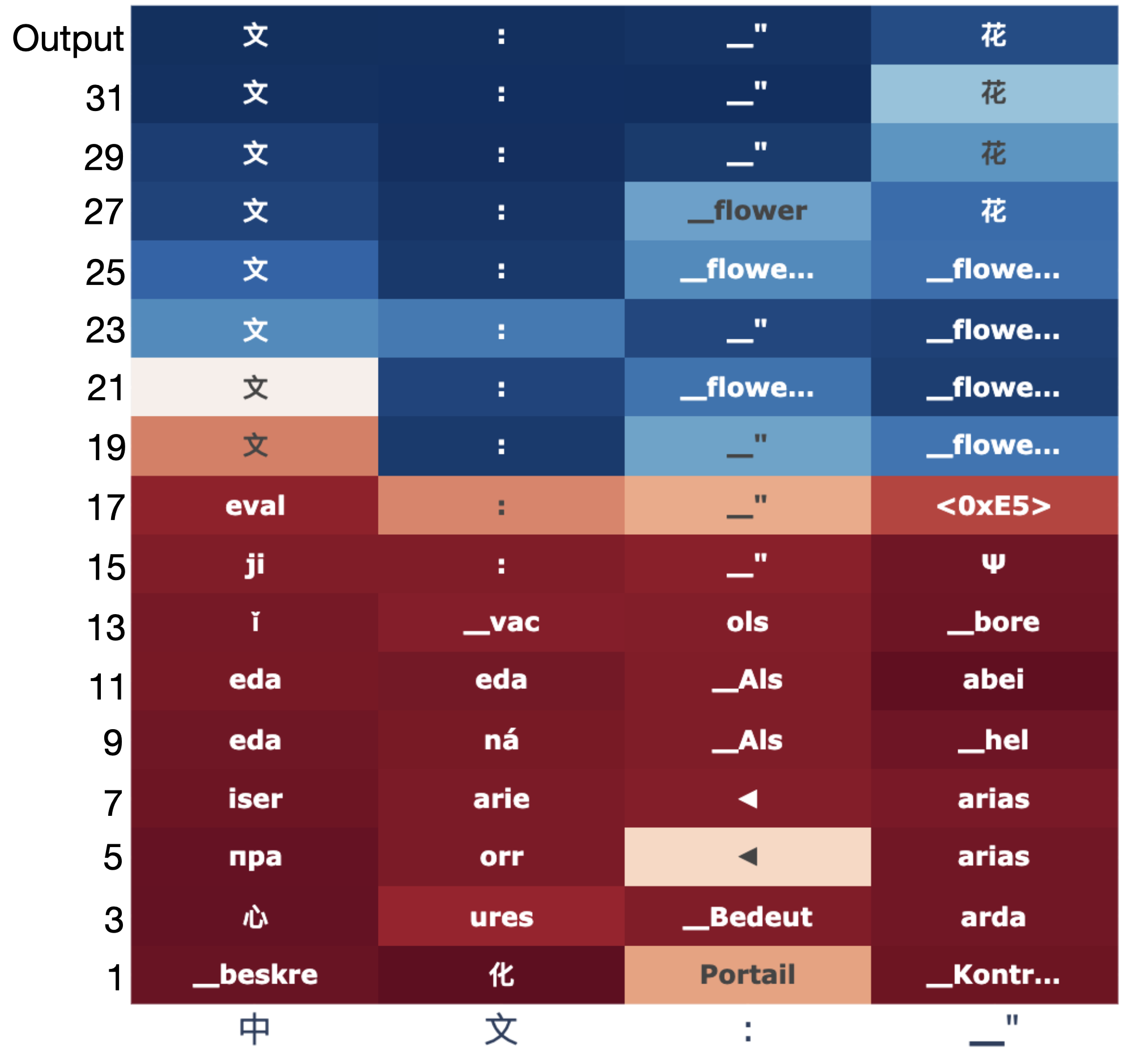}
    \vspace{-7mm}
    \caption{\textbf{Illustration of logit lens,}
    which applies language modeling head (here, \llama{}-7B) prematurely to latent embeddings in intermediate layers, yielding one next\hyp token distribution per position ($x$-axis) and layer ($y$-axis).
    We show final tokens of translation prompt (\cf\ \Secref{sec:data}) ending with ``Français:\ "fleur" - \zh{中文}:\ "'' (where ``\zh{中文}'' means ``Chinese''). Final layer correctly ranks ``\zh{花}'' (translation of ``fleur'') on top, whereas intermediate layers decode English ``flower''.
    Color indicates entropy of next\hyp token distributions from low (blue) to high (red). (Plotting tool: \citet{belrose2023eliciting}.)}
    \vspace{-4mm}
    \label{fig:translation-llama2-7b-ex-single}
\end{figure}

Intuitively, one way to achieve strong performance on non\hyp English data in a data\hyp efficient manner is to use English as a pivot language, by first translating input to English, processing it in English, and then translating the answer back to the input language.
This method has been shown to lead to high performance when implemented {explicitly} \cite{shi2022language, ahuja2023mega, huang2023languages}.
Our guiding inquiry in this work is whether pivoting to English also occurs {implicitly} when LLMs are prompted in non\hyp English.

In the research community as well as the popular press, many seem to assume that the answer is yes, epitomized by claims such as, ``The machine, so to say, thinks in English and translates the conversation at the last moment into Estonian'' \cite{piir2024finland}.
In this work, we set out to move beyond such speculation and investigate the question empirically.

% In this work, we try to illuminate exactly this question of whether English acts as a pivot language in modern LLMs like ChatGPT~\cite{openai2023chatgpt, open2023gpt4}, Claude, Bard, Mistral~\cite{jiang2023mistral, jiang2024mixtral} and Llama~\cite{touvron2023llama1, touvron2023llama}. 

The question is of major importance.
On the one hand, implicitly using English as an internal pivot could bias LLMs toward Anglocentric patterns that 
could predispose the model to certain linguistic elements (lexicon, grammar, metaphors, \etc),
% Estonian ChatGPT “sounds like an American person that is super fluent in Estonian”
% (\eg, in one user's assessment, ChatGPT ``sometimes [\dots{}]\ uses a clear translation of an English expression that does not feel natural in Portuguese at all'' (https://news.ycombinator.com/item?id=37911776)),
while also shaping more profound behaviors related to, \eg, emotional stance \cite{Boroditsky2003SexSyntaxSemantics} or temporal reasoning \cite{NunezSweetser2006}.
% Maybe more important – and less conspicuous (harder to notice esp. When at surface text is fluent and correct; unnoticed bias)
On the other hand, if LLMs do not use English as a pivot, it raises questions of how else they manage to work so remarkably well even in low\hyp resource languages.
Overall, the quest for an internal pivot language holds promise to advance our understanding of how LLMs function no matter if we succeed.

Investigating the existence of an internal LLM language is complicated by the scale and notoriously inscrutable nature of the neural networks behind LLMs, which after the input layer do not operate on discrete tokens, but on high\hyp dimensional floating\hyp point vectors.
How to understand if those vectors correspond to English, Estonian, Chinese, \etc---or to no language at all---is an open problem, and the question of whether LLMs use an internal pivot language has therefore, to the best of our knowledge, not been addressed empirically before.

\xhdr{Summary of contributions}
To overcome these hurdles, we draw on, and contribute to, the nascent field of mechanistic interpretability (\cf\ \Secref{sec:Related work}).
%and develop a methodology for exploring to what extent the internal states of transformer\hyp based LLMs encode information in specific natural languages.
In a transformer, each input token's embedding vector is gradually transformed layer by layer without changing its shape.
After the final layer, an ``unembedding'' operation turns the vector into a next-token distribution.
Focusing on the \llama family of models \cite{touvron2023llama}---among today's largest open\hyp source LLMs---we find that applying the ``unembedding'' operation prematurely in intermediate, non-final layers---a technique called \textit{logit lens} \cite{nostalgebraist2020logitlens}---already decodes a contextually appropriate token early on (\Figref{fig:translation-llama2-7b-ex-single}), giving us a (limited) glimpse at the model's otherwise hard\hyp to\hyp interpret numerical internal state.

Exploiting this fact, we carefully devise prompts that allow us to determine whether a logit\hyp lens\hyp decoded token is semantically correct and to what language it belongs
(\eg, a prompt asking the model to translate French ``fleur'' [``flower''] to Chinese ``\zh{花}''; \cf\ \Figref{fig:translation-llama2-7b-ex-single}).
%(\eg, a prompt asking the model to print ``Blume'', the German word for ``flower''; \cf\ \Figref{fig:translation-llama2-7b-ex-single}).
Tracking language probabilities across layers, we observe that no contextually appropriate tokens are decoded in the first half of layers,
followed by a sudden shift of probability mass onto the English version (``flower'') of the correct next token,
and finally a shift to the correct next token in the target language (``\zh{花}'').
%(``Bl'', the first token of ``Bl$\cdot$ume'').

Expanding on this first evidence of English as an internal pivot language, we analyze latent embeddings directly as high\hyp dimensional Euclidean points, rather than via the logit lens.
This allows us to draw a more nuanced picture of the anatomy of \llama's forward pass, suggesting that, in middle layers, the transformer operates in an abstract ``concept space'' that is partially orthogonal to a language\hyp specific ``token space'', which is reached only in the final layers.
In this interpretation, the latent embeddings' proximity to English tokens observed through the logit lens follows from an English bias in concept space, rather than from the model first translating to English and then ``restarting'' its forward pass from there.

We conclude by discussing implications and future directions for studying latent biases and their effects---a crucial step toward trustworthy AI.

\section{Related work}
\label{sec:Related work}

% This work is situated at the intersection of two research directions: multilingual LMs and mechanistic interpretability.

% \label{sec:Multilingual language models}

\xhdr{Multilingual language models}
Multilingual language models (LMs) are trained to simultaneously handle multiple input languages. 
Examples include
mBERT \cite{devlin2018bert},
% which was trained on Wikipedia. Afterwards, we have
mBART \cite{liu-etal-2020-multilingual-denoising},
% derived from BART~\cite{Devlin2019BERTPO},
XLM-R \cite{conneau2020unsupervised},
% from RoBERTa \cite{liu2019roberta},
mT5 \cite{Xue_2021},
% from T5~\cite{raffel2023exploring},
XGLM \cite{Lin_2022},
mGPT \cite{shliazhko2022mgpt},
% from  GPT-3~\cite{brown2020language}.
% The largest multilingual LLM is 
BLOOM \cite{workshop2022bloom},
% with 176 billion parameters, trained on the ROOTS corpus~\cite{laurenon2023bigscience}.
 % (46 natural languages and 13 programming languages). %Although English constitutes 30.03\% of this dataset, several other languages are highly represented (e.g., 16.16\% simplified Chinese, 12.9\% French, 10.85\% Spanish, 4.91\% Portuguese, and 4.6\% Arabic.
and PolyLM \cite{wei2023polylm}.
% demonstrated state-of-the-art performance.
Current frontier models such as
GPT-4,
% \cite{openai2023gpt4},
PaLM,
% \cite{chowdhery2023palm},
and \llama{},
% \cite{touvron2023llama},
despite performing better in English due to their Anglocentric training data 
% jin2023better,
\cite{huang2023languages,bang2023multitask,zhang-etal-2023-dont}, still do well across languages \cite{shi2022language}.

% In this work, we focus on \llama{}, a frontier open-source model trained on an English\hyp dominated multilingual corpus (\cf\ \Secref{sec:llama}) that has since become the backbone of multilingual LLMs \cite{goddard2023polyglot, openbuddy2023} as well as LLMs for individual languages \cite{pluster2023leolm,clibrain2023es,huang2023vigogne,kim2023kor}.

Researchers have devised numerous methods for efficiently transferring LM capabilities across languages,
\eg, by aligning contextual embeddings \cite{schuster-etal-2019-cross,cao2020multilingual},
% found that aligning contextual embeddings improves mBERT to match the downstream performance of non-English models trained in fully-supervised settings.
% In a different vein, \citet{Artetxe_2020} show that it suffices to re-learn just the embedding matrix of a monolingual model in order to adapt it to a new language.
relearning embedding matrices during finetuning on a new language \cite{Artetxe_2020},
or repeatedly doing so during pretraining \cite{chen2023improving}.
% Similarly, \citet{chen2023improving} show that repeatedly re-learning the embedding matrix during pretraining leads to more efficient and better adaptation to new languages later on. Both indicate that monolingual models pick up universal abstractions that can generalize to different languages. 

Several approaches leverage English as a pivot language.
For instance, \citet{zhu2023extrapolating} show that Llama can be efficiently augmented with multilingual instruction\hyp following capabilities thanks to its English representations. Likewise, \citet{zhu2024question} demonstrate the feasibility of leveraging language models' proficiency in English for non-English contexts by fine-tuning them on translation data and English-only instructional data. They successfully employ this approach to enhance the multilingual reasoning capabilities of Llama-2. Regarding non-Latin low-resource languages, \citet{husain2024romansetu} illustrate that leveraging both romanized and English data proves to be an effective strategy for efficiently improving multilingual task performance.
Prompting strategies, too, can improve multilingual performance by leveraging English as a pivot language, \eg, by simply first translating prompts to English \cite{shi2022language,ahuja2023mega,etxaniz2023multilingual}
or by instructing LMs to perform chain-of-thought reasoning \cite{wei2022chain} in English \cite{huang2023languages}.

Although employing high-resource languages can enhance performance on low-resource languages, it might also bias output generation in low-resource languages, \eg, in terms of grammar \cite{papadimitriou2022multilingual}.

Researchers have also investigated how latent representations differ across languages within multilingual models.
In the case of encoder-only models such as mBERT, converging evidence suggests the existence of a language\hyp agnostic space in later layers following language\hyp specific early layers \cite{libovicky2020language, conneau2020emerging, muller2021first, choenni2020does}.

% The topic of similarities between different models or hidden states across multiple languages has been investigated in several past works \cite{libovicky2020language, conneau2020emerging, muller2021first, choenni2020does} in the context of encoder-only models.
% In particular, \citet{libovicky2020language} explore several ways to suppress the language identity from intermediate embeddings of BERT-like models and therefore increase their language-neutrality. \citet{conneau2020emerging} show that in BERT-like models the middle layers are more similar across different languages than the last layers. Similarly, \citet{muller2021first} explore the origins of cross-lingual transfer and find that upper layers of mBERT have smaller cross-lingual similarities than the earlier ones, suggesting the existence of a language-agnostic space following an early-layers language-specific encoder.
% Notably, \citet{choenni2020does} find evidence of a language-agnostic concept space in the upper layers of mBERT, right before languages are spatially separated once again at the very last layers.

\xhdr{Mechanistic interpretability} 
% \label{sec:Mechanistic interpretability} 
The nascent field of mechanistic interpretability (MI)
% built around the idea that neural networks learn human\hyp interpretable algorithms,
aims to reverse\hyp engineer and thereby understand neural networks,
% \cite{olah2020zoom},
using techniques such as circuit discovery \cite{nanda2023progress,conmy2023towards},
controlled task\hyp specific training \cite{li2022emergent,marks2023geometry},
and causal tracing \cite{meng2022locating,monea2023glitch}.

% \xhdrIt{Language neurons and circuits}
For smaller models, \eg, GPT-2 \cite{radford2019gpt2} and Pythia \cite{biderman2023pythia}, MI approaches such as sparse probing \cite{gurnee2023finding} have revealed monosemantic French \cite{gurnee2023finding} and German \cite{quirke2023training} language neurons and context\hyp dependent German $n$-gram circuits (subnetworks for boosting the probability of German $n$-grams when the monosemantic German context neuron is active) \cite{quirke2023training}.
% and studied their formation during training.  

% \xhdr{Intermediate decoding}
The most relevant tools from the MI repertoire in the context of this work are the \textit{logit lens} \cite{nostalgebraist2020logitlens}, \textit{tuned lens} \cite{belrose2023eliciting}, and \emph{direct logit attribution} \cite{elhage2021mathematical}, which decode intermediate token representations from transformer models in different ways.
The logit lens does so by using the language modeling head, which is usually only applied in the final layer, prematurely in earlier layers, without any additional training.
The more sophisticated tuned lens additionally trains an affine mapping for transforming an intermediate latent state such that it mimics the token predictions made by the final latent state. Finally, direct logit attribution generalizes the logit lens by considering the logit contribution of each individual attention head.

In this work, we heavily rely on the logit lens, described further in \Secref{sec:logit lens}, as opposed to the tuned lens. The latter would defeat our purpose of understanding whether \llama, when prompted in non\hyp English, takes a detour via English internal states before outputting non\hyp English text.
As the tuned lens is specifically trained to map internal states---even if corresponding to English---to the final, non\hyp English next\hyp token prediction, the optimization criterion would ``optimize away'' our signal of interest.

% In this work, we stick with the original logit lens, because we are interested precisely in the intermediate decodings. In particular, we want to detect whether their decoded language differs from the language of the \textbf{true} next token, and not in skipping layers. 

%%%%%%%%%%%%%%%%%%%%%%%%%%%%%%%%%%%%%%%
\section{Materials and methods}
%%%%%%%%%%%%%%%%%%%%%%%%%%%%%%%%%%%%%%%

%%%%%%%%%%%%%%%%%%%%%%%%%%%%%%%%%%%%%%%
\subsection{Language models: \llama}
\label{sec:llama}
%%%%%%%%%%%%%%%%%%%%%%%%%%%%%%%%%%%%%%%

We focus on the \llama family of language models \cite{touvron2023llama}, some of the largest and most widely used open\hyp source models. The models were trained on a multilingual corpus that is largely dominated by English, which comprises 89.70\% of the corpus. 
However, given the size of the training data (two trillion tokens), even a small percentage of non\hyp English training data still constitutes a large number of tokens in absolute terms (\eg, 0.17\% = 3.4B German tokens, 0.13\% = 2.6B Chinese tokens).
Consequently, \llama is, despite its English bias, considered a multilingual model. 

% As we will show later, in the datasets considered in this paper, \llama's performance on German and French inputs are mostly on par with its performance on English. However, on Russian its usually slightly worse. 

\xhdr{Versions}
\llama comes in three model sizes, with 7B\slash 13B\slash 70B parameters,
32\slash 40\slash 80 layers,
and embedding dimension $d=$ 4096\slash 5120\slash 8192, respectively.
Across all model sizes, the vocabulary $V$ contains $v=$ 32,000 tokens.
Here we study all model sizes, using 8-bit quantization~\cite{dettmers2022llmint8} in our experiments.

\xhdr{Architecture}
\llama is an autoregressive, decoder\hyp only, residual\hyp based transformer.
Such models maintain the shape of the input data throughout the computation process during a forward pass:
one embedding vector, a so-called \textit{latent,} per input token $x_1, \dots, x_n \in V$, where $n$ is the input sequence length. 
The initial latents $h^{(0)}_1, \dots, h^{(0)}_n \in \mathbb{R}^{d}$ are obtained from a learned embedding dictionary that contains one fixed vector per vocabulary token.
Each of these latents is incrementally updated layer by layer by adding a residual. 
The residual added to the latent at position $i$ in layer $j$ is a function $f_{j}$ of all preceding tokens' latents $h^{(j-1)}_1, \dots, h^{(j-1)}_{i}$: 
 \begin{equation}
        h^{(j)}_i = h^{(j-1)}_i + f_{j}\left(h^{(j-1)}_1, \dots, h^{(j-1)}_{i}\right),
\end{equation}
where the resulting vector $h^{(j)}_i$
% (after layer $j$)
is still of dimension $d$.
The function $f_{j}$ itself, called a transformer block, is composed of a masked self-attention layer followed by a feed-forward layer with a residual connection and root mean square (RMS) normalization in between \cite{vaswani2017attention, touvron2023llama}. 
Due to RMS normalization, all latents lie on a $d$\hyp dimensional hypersphere of radius $\sqrt{d}$.

In pretraining, all transformer blocks $f_1, \dots, f_m$ (with $m$ the number of layers) are tuned such that the final latent $h^{(m)}_i$ for position $i$ is well\hyp suited for predicting the token at position $i+1$. For prediction, the final embedding vector is multiplied with a so-called \textit{unembedding matrix} $U \in \mathbb{R}^{v \times d}$, which yields a real vector
$z_i = U h^{(m)}_i \in \mathbb{R}^{v}$ 
containing a so-called \textit{logit} score $z_{it}$ for each vocabulary token $t \in V$. These scores are then transformed into probabilities $P(x_{i+1}=t \,|\, x_{1}, \dots, x_{i}) \propto e^{z_{it}}$ via the softmax operation.

%  \begin{equation}
%  P(x_{i+1}=t \,|\, x_{1}, \dots, x_{i}) =
%  \frac{e^{z_{it}}}{\sum_{t' \in V}e^{z_{it'}}},
%  \end{equation}
% where $z_{it}$ is $z_i$'s entry corresponding to token $t$.

%%%%%%%%%%%%%%%%%%%%%%%%%%%%%%%%%%%%%%%
\subsection{Interpreting latent embeddings: Logit lens}
\label{sec:logit lens}
%%%%%%%%%%%%%%%%%%%%%%%%%%%%%%%%%%%%%%%

When transformers are deployed in practice, only the final latent vectors after the last transformer block are turned into token distributions by multiplying them with $U$ and taking a softmax.
However, since latents have the same shape in all layers, any latent can in principle be turned into a token distribution, by treating it as though it were a final-layer latent.
Prematurely decoding tokens from latents this way, a method called the \textit{logit lens}
(\cf\ \Secref{sec:Related work}), can facilitate the inspection and interpretation of the internal state of transformers.
Using the logit lens, we obtain one next\hyp token distribution $P(x_{i+1} \,|\, h^{(j)}_i)$ per position $i$ and layer~$j$. 

We illustrate the logit lens in \Figref{fig:translation-llama2-7b-ex-single}, where every cell shows the most likely next token when applying the logit lens to the latent in that position and layer.
As seen, the logit lens decodes contextually appropriate tokens already in intermediate layers.

\begin{figure*}[ht!]
     \centering 
     \begin{subfigure}[b]{0.3\textwidth}
         \centering
         \includegraphics[width=\textwidth]{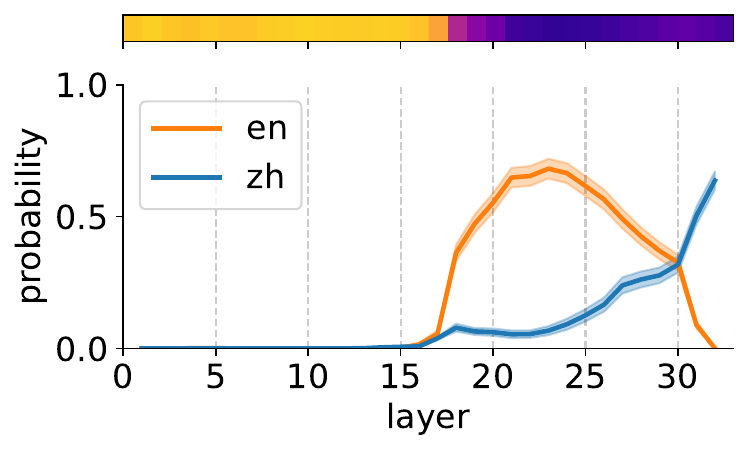}
     \end{subfigure}
     \hfill
     \begin{subfigure}[b]{0.28\textwidth}
         \centering
         \caption{Translation task}
         \includegraphics[width=\textwidth]{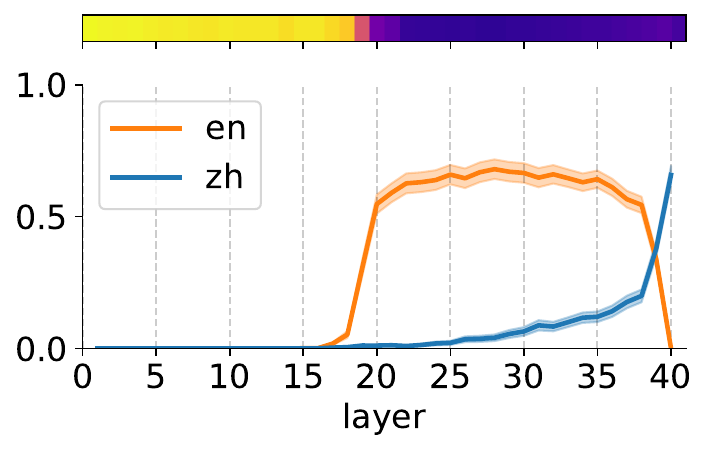}
     \end{subfigure}
     \hfill
     \begin{subfigure}[b]{0.343\textwidth}
         \centering
         \includegraphics[width=\textwidth]{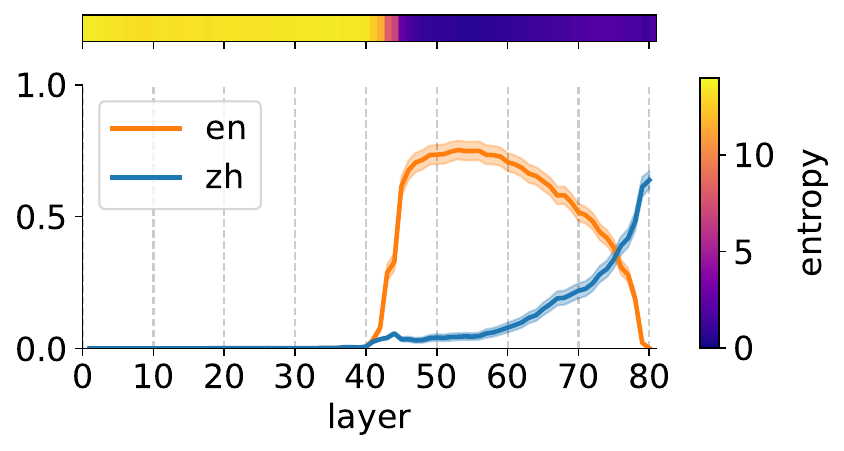}
     \end{subfigure}

    \begin{subfigure}[b]{0.3\textwidth}
         \centering
         \includegraphics[width=\textwidth]{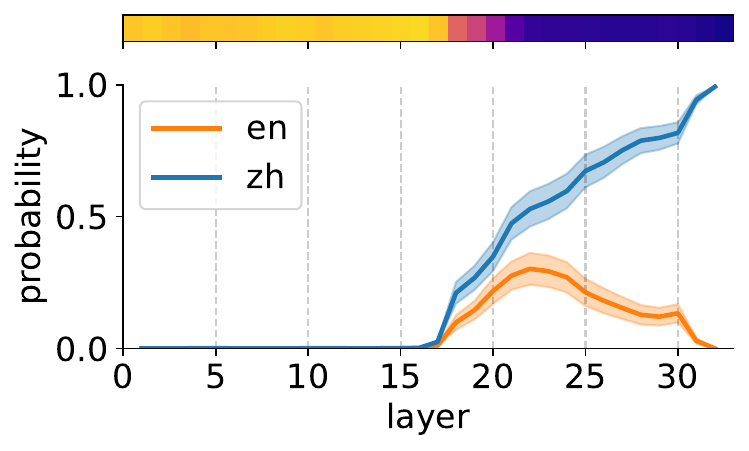}
     \end{subfigure}
     \hfill
     \begin{subfigure}[b]{0.28\textwidth}
         \centering
         \caption{Repetition task}
         \includegraphics[width=\textwidth]{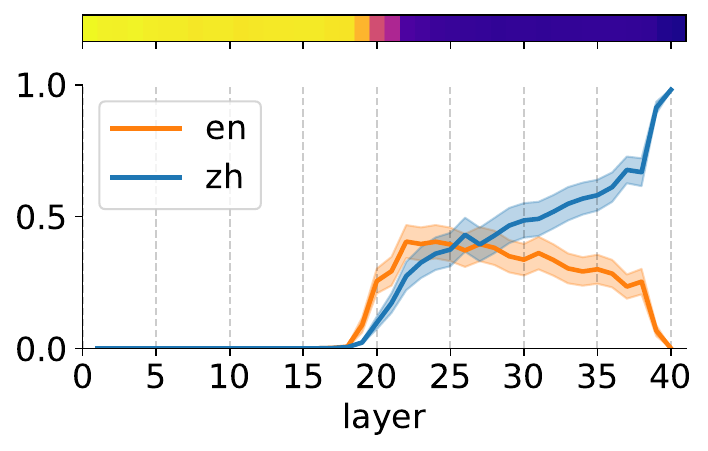}
     \end{subfigure}
     \hfill
     \begin{subfigure}[b]{0.343\textwidth}
         \centering
         \includegraphics[width=\textwidth]{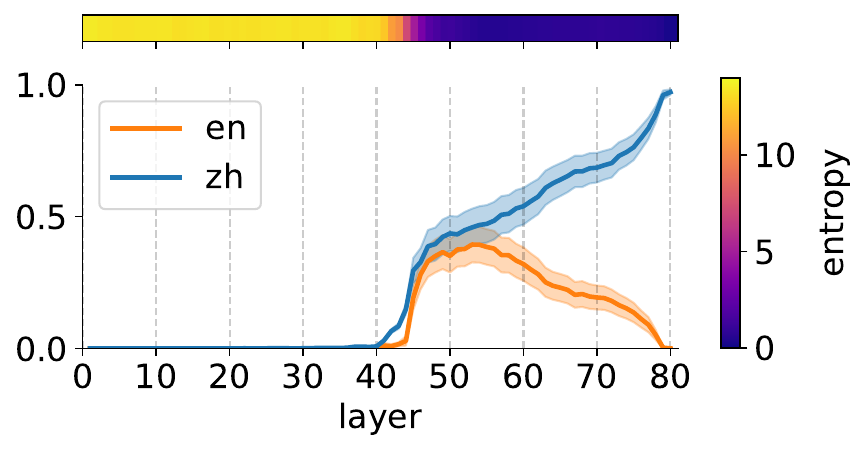}
     \end{subfigure}

     \begin{subfigure}[b]{0.3\textwidth}
         \centering
         \includegraphics[width=\textwidth]{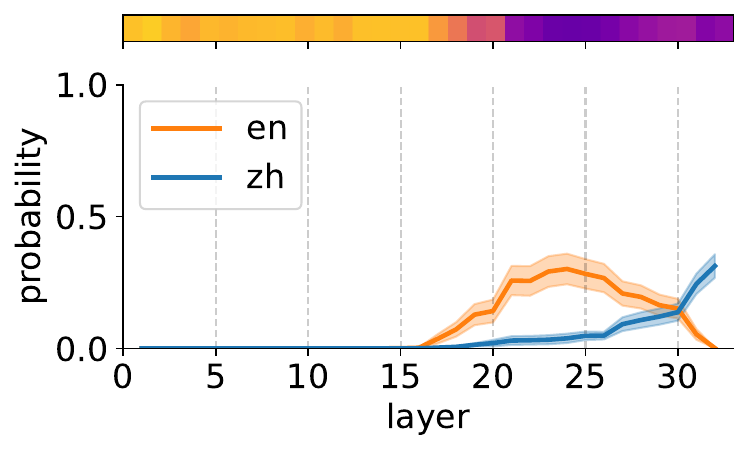}
         \caption*{7B}
     \end{subfigure}
     \hfill
     \begin{subfigure}[b]{0.28\textwidth}
         \centering
         \caption{Cloze task}
         \includegraphics[width=\textwidth]{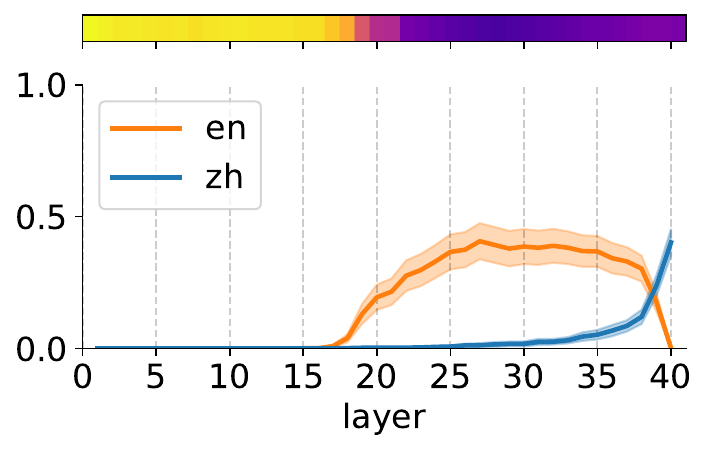}
         \caption*{13B}
     \end{subfigure}
     \hfill
     \begin{subfigure}[b]{0.343\textwidth}
         \centering
         \includegraphics[width=\textwidth]{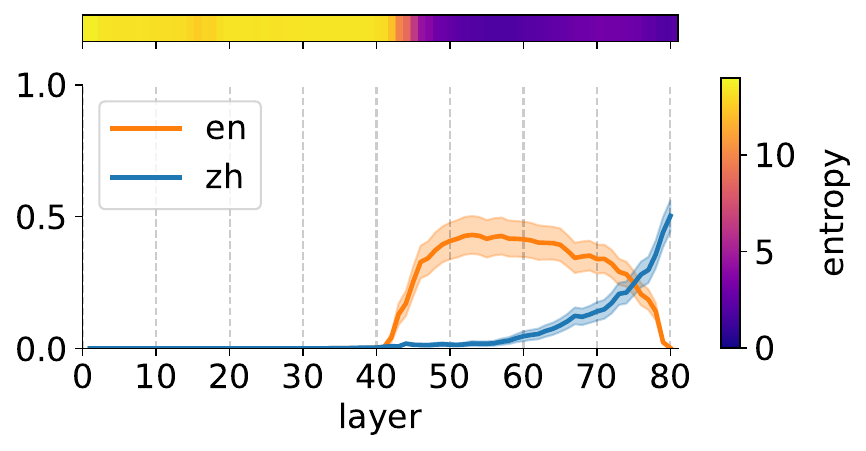}
         \caption*{70B}
     \end{subfigure}
\vspace{-2mm}
     \caption{
     \textbf{Language probabilities for latents during \llama forward pass,}
     for
     \textbf{(a)}~translation task from union of German\slash French\slash Russian to Chinese,
     \textbf{(b)}~Chinese repetition task,
     \textbf{(c)}~Chinese cloze task.
     Each task evaluated for model sizes (columns) 7B, 13B, 70B.
     On $x$-axes, layer index; on $y$-axes, probability (according to logit lens) of correct Chinese next token (blue) or English analog (orange).
     Error bars show 95\% Gaussian confidence intervals over input texts (353 for translation, 139 for repetition and cloze). 
     }
     \label{fig:combined_probs_entropy}
 \end{figure*}

 %%%%%%%%%%%%%%%%%%%%%%%%%%%%%%%%%%%%%%%
\subsection{Data: Tasks for eliciting latent language}
\label{sec:data}
%%%%%%%%%%%%%%%%%%%%%%%%%%%%%%%%%%%%%%%

Our goal is to explore whether \llama's internal, latent states correspond to specific natural languages.
Although the logit lens allows us to map latent vectors to token distributions, we still require a mapping from token distributions to languages.

Doing so in general is difficult as many tokens are ambiguous with respect to language; \eg, the token ``an'' is commonly used in English, French, and German, among others.
To circumvent this issue, we construct prompts $x_1 \dots x_n$ where the correct next token $x_{n+1}$ is (1)~obvious and (2)~can be unambiguously attributed to one language.

\xhdr{Prompt design}
To ensure that the next token is obvious (criterion~1), we design three text completion tasks where the next token $x_{n+1}$ can be easily inferred from the prompt $x_1 \dots x_n$.
In describing the tasks, we use Chinese as an example language.

\xhdrIt{Translation task} Here the task is to translate the preceding non-English (\eg, French) word to Chinese. We show the model four words with their correct translations, followed by a fifth word without its translation, and let the model predict the next token (``\zh{中文}'' means ``Chinese'' below):

\prompt{
Français: "vertu" - \zh{中文}: "\zh{德}"\\
Français: "siège" - \zh{中文}: "\zh{座}"\\
Français: "neige" - \zh{中文}: "\zh{雪}"\\
% Français: "supprimer" - \zh{中文}: "\zh{删}"\\
Français: "montagne" - \zh{中文}: "\zh{山}"\\
Français: "fleur" - \zh{中文}: "
}

With such a prompt, \llama can readily infer that it should translate the fifth French word.
We carefully select words as described below and construct one prompt per word by randomly sampling demonstrations from the remaining words.

\xhdrIt{Repetition task} Similarly, we task the model to simply repeat the last word, instead of translating it, by prompting as follows:

\prompt{
\zh{中文}: "\zh{德}" - \zh{中文}: "\zh{德}"\\
\zh{中文}: "\zh{座}" - \zh{中文}: "\zh{座}"\\
\zh{中文}: "\zh{雪}" - \zh{中文}: "\zh{雪}"\\
\zh{中文}: "\zh{山}" - \zh{中文}: "\zh{山}"\\
\zh{中文}: "\zh{花}" - \zh{中文}: "
}

% We again construct one prompt per word.

\xhdrIt{Cloze task}
As a slightly harder task, we consider a cloze test, where the model must predict a masked word in a sentence. 
Given a target word, we construct an English sentence starting with the word by prompting GPT-4,
mask the target word,
and
translate the sentence to the other languages.
To construct prompts, we sample two demonstrations from the remaining words. 
An English example before translation to the other languages follows:

\prompt{
A "\_\_\_" is used to play sports like soccer and basketball. Answer: "ball".\\
A "\_\_\_" is a solid mineral material forming part of the surface of the earth. Answer: "rock".\\
A "\_\_\_" is often given as a gift and can be found in gardens. Answer: "
%"\_\_\_" is a solid material that forms part of the Earth's surface. Answer: "Rock".\\
%"\_\_\_" is a device commonly used in households to watch broadcast programs and movies. Answer: "Television".\\
%"\_\_\_" is the star at the center of the solar system that provides the Earth with light and heat." Answer: "
% "\_\_\_" ist ein festes Material, das einen Teil der Erdoberfläche bildet. Antwort: "Gestein".\\
% "\_\_\_" ist ein Gerät, das häufig in Haushalten verwendet wird, um ausgestrahlte Programme und Filme anzusehen. Antwort: "Fernsehen".\\
% "\_\_\_" ist der Stern im Zentrum des Sonnensystems, der die Erde mit Licht und Wärme versorgt." Antwort: "
}

\xhdr{Word selection}
To enable unambiguous language attribution (criterion~2), we construct a closed set of words per language.
As a particularly clean case, we focus on Chinese, which has many single\hyp token words and does not use spaces.
We scan \llama{}'s vocabulary for single\hyp token Chinese words (mostly nouns) that have a single\hyp token English translation.
This way, \llama's probabilities for the correct next Chinese word and for its English analog can be directly read off the next\hyp token probabilities.

For robustness, we also run all experiments on German, French, and Russian.
For this, we translate the selected Chinese\slash English words and, for each language, discard words that share a token prefix with the English version, as this would render language detection (\cf\ \Secref{sec:Language probability}) ambiguous.

We work with 139 Chinese, 104 German, 56 French, and 115 Russian words (\cf\ \Appref{app:Word translation}).

% by scanning \llama{}'s vocabulary for single\hyp token Chinese nouns that have a single\hyp token English translation. Subsequently, we translate the resulting words to German, French, and Russian.
% Next, for each language $\ell$, we discard words that share a token prefix with the English version, as this would render language detection ambiguous.
% For example, in English the word ``can'' (as in ``beer can'') tokenizes into one token, and its French version ``cannette'' tokenizes into three tokens (``can$\cdot$net$\cdot$te''). If the logit lens decodes the token ``can'' from a latent state, we would be unable to attribute it to either French or English.

\subsection{Measuring latent language probabilities}
\label{sec:Language probability}

To investigate a hypothetical pivot language inside \llama, we apply the logit lens to the latents $h_n^{(j)}$ corresponding to the last input token $x_n$ for each layer $j$, obtaining one next\hyp token distribution $P(x_{n+1} \,|\, h^{(j)}_{n})$ per layer.
% which can be thought of as intermediate beliefs about the next token.
Our prompts (\cf\ \Secref{sec:data}) are specifically designed such that an intermediate next\hyp \textit{token} distribution lets us estimate the probability of the correct next \textit{word} in the input language as well as English.
Since we specifically select single\hyp token words in Chinese (\lang{ZH}) as well as English (\lang{EN}), we can simply define the probability of language $\ell \in \{\lang{ZH},\lang{EN}\}$ as the probability of the next token being $\ell$'s version $t_\ell$ of the correct single\hyp token word:
$P(\text{lang} = \ell \,|\, h^{(j)}_{n}) := P(x_{n+1} = t_{\ell} \,|\, h^{(j)}_{n})$. 
(For readability we also simply write $P(\text{lang} = \ell)$.)
% without the conditional.)
Note that this does not define a distribution over languages, as generally
$\sum_\ell P(\text{lang} = \ell) < 1$.

In other languages (and in corner cases in Chinese and English), we must account for multiple tokenizations
% of the same word
and whitespaces
(\cf\ \Appref{app:Computing language probabilities}).

%%%%%%%%%%%%%%%%%%%%%%%%%%%%%%%%%%%%%%%
\section{Results} 
\label{sec:results}

When presenting results, we first (\Secref{sec:results:logit lens}) take a probabilistic view via the logit lens (\Secref{sec:logit lens}), for all tasks and all model sizes.
(Since the results are consistent across languages, we focus on 
Chinese here and refer to Appendix~\ref{app:results} for French, German, and Russian.)
Then (\Secref{sec:geometric view}) we drill deeper by taking a geometric view of how token embeddings drift as the transformer computes layer by layer.

\subsection{Probabilistic view: Logit lens}
\label{sec:results:logit lens}

The logit lens gives us one set of language probabilities (\cf\ \Secref{sec:Language probability}) per input prompt and layer.
\Figref{fig:combined_probs_entropy} tracks the evolution of language probabilities from layer to layer,
with one plot per combination of model size (columns) and task%
\footnote{
In \Figref{fig:combined_probs_entropy}, translation task uses union of German, French, and Russian as source languages.
For individual source languages, as well as all target languages, \cf\ Appendix~\ref{app:results}.
}
(rows).
The $x$-axes show layer indices, and the $y$-axis the language probabilities $P(\text{lang}=\lang{ZH})$ and $P(\text{lang}=\lang{EN})$
averaged over input prompts.

On the translation and cloze tasks a consistent picture emerges across model sizes.
Neither the correct Chinese token nor its English analog garner any noticeable probability mass during the first half of layers.
Then, around the middle layer,
English begins a sharp rise followed by a decline, while Chinese slowly grows and, after a crossover with English, spikes on the last five layers.
On the repetition task, Chinese already rises alongside English (discussed in \Secref{sec:discussion}). This is in contrast to all other languages,
where English rises first
(\Appref{app:results}). 

On top of the language probabilities (\Secref{sec:Language probability}), the entropy of the full next-token distribution is shown as a heatmap above the plots.
We again observe a consistent pattern across tasks and model sizes:
high entropy in the first half of layers, while both $P(\text{lang}=\lang{ZH})$ and $P(\text{lang}=\lang{EN})$ are close to zero, followed by a sharp drop at the same time that $P(\text{lang}=\lang{EN})$ rises.
From there on, entropy remains low, with a slight rebound as probability mass shifts from English to Chinese.

With $32{,}000 \approx 2^{15}$ tokens in the vocabulary, the early entropy of around 14 bits implies a close\hyp to\hyp uniform next\hyp token distribution (around 15 bits).

\xhdr{Path visualization}
The plots of \Figref{fig:combined_probs_entropy} only consider the probability of the correct Chinese next token and its English analog, without speaking to the remaining tokens.
To form an intuition of the entire distribution, we use dimensionality reduction to visualize the data.
First, we define the distance between a latent $h_n$ at position $n$ and a token $t$ via the negative log\hyp likelihood of $t$ given $h_n$, as computed by the logit lens (\cf\ \Secref{sec:Language probability}):
$d(h_n,t) = -\log P(x_{n+1} = t \,|\, h_n)$.
Then, we use classical multidimensional scaling to embed tokens and latents in an approximately distance\hyp preserving joint 2D space.
(Intra-token and intra-latent distances are set to $\max_{h,t} d(h,t)$, which serves as a ``spring force'' pushing the 2D points apart.)

\begin{figure}[t]
    \centering
    \includegraphics[width=\linewidth]{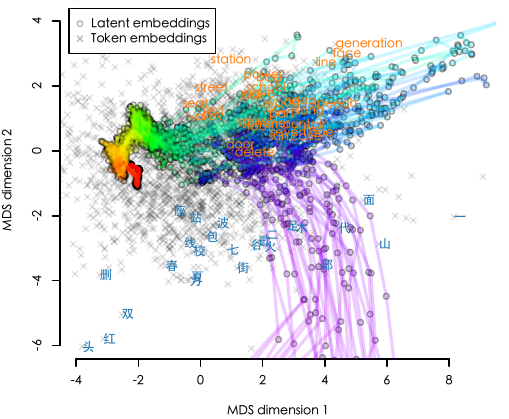}
    \vspace{-7mm}
    \caption{\textbf{Latent trajectories through transformer layers.} 2D embedding of latents ($\circ$) and output tokens ($\times$) found via multidimensional scaling. Latents for same prompt connected by rainbow\hyp colored path, proceeding from layer~1 (red) to 80 (violet). Labels for correct Chinese next tokens (one per prompt) in blue, for English analogs in orange. Takeaway: latents reach correct Chinese token after detour through English.}
    \label{fig:mds}
\end{figure}

A transformer's forward computation for a given final input token $x_n$ can now be visualized by connecting the 2D embeddings of the latents $h_n^{(j)}$ in subsequent layers $j$,
as presented and explained in \Figref{fig:mds} (German\hyp to\hyp Chinese translation, 70B).
We make two observations:
(1)~An English and a Chinese token cluster emerges, suggesting that the same latent also gives high probability to an entire language, in addition to the language\hyp specific version of the correct next token.
(2)~Paths first pass through the English cluster, and only later reach the Chinese cluster.
Taken together,
% with \Figref{fig:combined_probs_entropy},
the emerging picture is that, when translating a German word to Chinese, \llama takes a ``detour'' through an English subspace.

So far, we have characterized the transformer's intermediate latent states from a probabilistic perspective, by studying the next-token distributions obtained via the logit lens.
For a deeper understanding, we next take a geometric perspective and analyze latents directly as points in Euclidean space, \ie, before mapping them to token probabilities.

\subsection{Geometric view: An 8192D space Odyssey}
\label{sec:geometric view}

Simplistically, the task solved by an autoregressive transformer is to map the input embedding of the current token to the output embedding of the next token.
The task is solved incrementally, each layer modifying (by adding a residual) the latent vector produced by the previous layer, a process that, geometrically, describes a path through $d$\hyp dimensional Euclidean space. We now set out to characterize this path.
Since the probabilistic view (\Figref{fig:combined_probs_entropy}) gave consistent results across tasks and model sizes,
we focus on one task (translation) and one model size (70B, \ie, $d=8192$).

\xhdr{Embedding spheres}
Output token embeddings (rows of the unembedding matrix $U$) and latents $h$ cohabitate the same $d$-dimensional Euclidean space.
In fact, due to RMS\hyp normalization (\Secref{sec:llama}), latents by construction live on a hypersphere of radius $\sqrt{d}\approx 90.1$.
% > summary(norms)
%    Min. 1st Qu.  Median    Mean 3rd Qu.    Max. 
%  0.5917  1.2161  1.3502  1.3529  1.4960  2.2594 
% > sd(norms)
% [1] 0.2103791
% > sd(norms)/mean(norms)
% [1] 0.1555075
Additionally, by analyzing the 2-norm of output token embeddings (mean 1.52, SD 0.23), we find that the latter also approximately lie on a sphere, of radius 1.52.

\xhdr{Token energy}
Importantly, token embeddings occupy their sphere unevenly; \eg, the first 25\% of the principal components account for 50\% of the total variance, and the first 54\% for 80\%.%
\footnote{
Moreover, \citet{cancedda2402spectral} showed that a significant fraction of the principal components can be omitted as long as attention sinking are preserved.}
To build intuition, first consider a hypothetical extreme case where tokens lie in a proper subspace (``token subspace'') of the full $d$\hyp dimensional space (even though, empirically, $U$ has rank $d$, so the tokens' output embeddings span all of $\mathbb{R}^d$).
If a latent $h$ has a component orthogonal to the token subspace, it includes information that is irrelevant for predicting the next token based on $h$ alone (since logits are scalar products of latent and token vectors).
The orthogonal component can still be important for the computations carried out by later layers and for predicting the next token in those layers. But the logit lens, which decodes latents into tokens prematurely in intermediate layers, will be blind to the orthogonal component.

A latent $h$'s angle with the ``token subspace'' thus measures how much of $h$ is irrelevant for immediately predicting the next token.
Concretely, we consider the mean squared cosine between $h$ and the token embeddings (rows of $U$) to capture how much of $h$'s ``energy'' translates into logit scores.
% (via scalar products with token embeddings).
For interpretability, we normalize by the mean squared cosine among token embeddings themselves,%
\footnote{In practice, we use $\hat U^\top\hat U$ instead of $\hat U\hat U^\top$ in \eqref{eq:token energy}, which has equal Frobenius norm but is more efficient to compute.}
obtaining what we call $h$'s squared \textit{token energy}
\begin{equation}
\label{eq:token energy}
    E(h)^2
    = \frac{\frac{1}{v} \|\hat Uh\|_2^2 \;/\; \|h\|_2^2}{\frac{1}{v^2}\|\hat U\hat U^\top\|_F^2}
    = \frac{v}{d} \, \frac{\|\hat Uh\|_2^2}{\|\hat U\hat U^\top\|_F^2}
\end{equation}
($\hat U$ being $U$ with 2-normalized rows),
which captures $h$'s proximity to ``token subspace'', compared to a random token's proximity to ``token subspace''.

\begin{figure}[t]
    \centering
    \includegraphics[width=\linewidth]{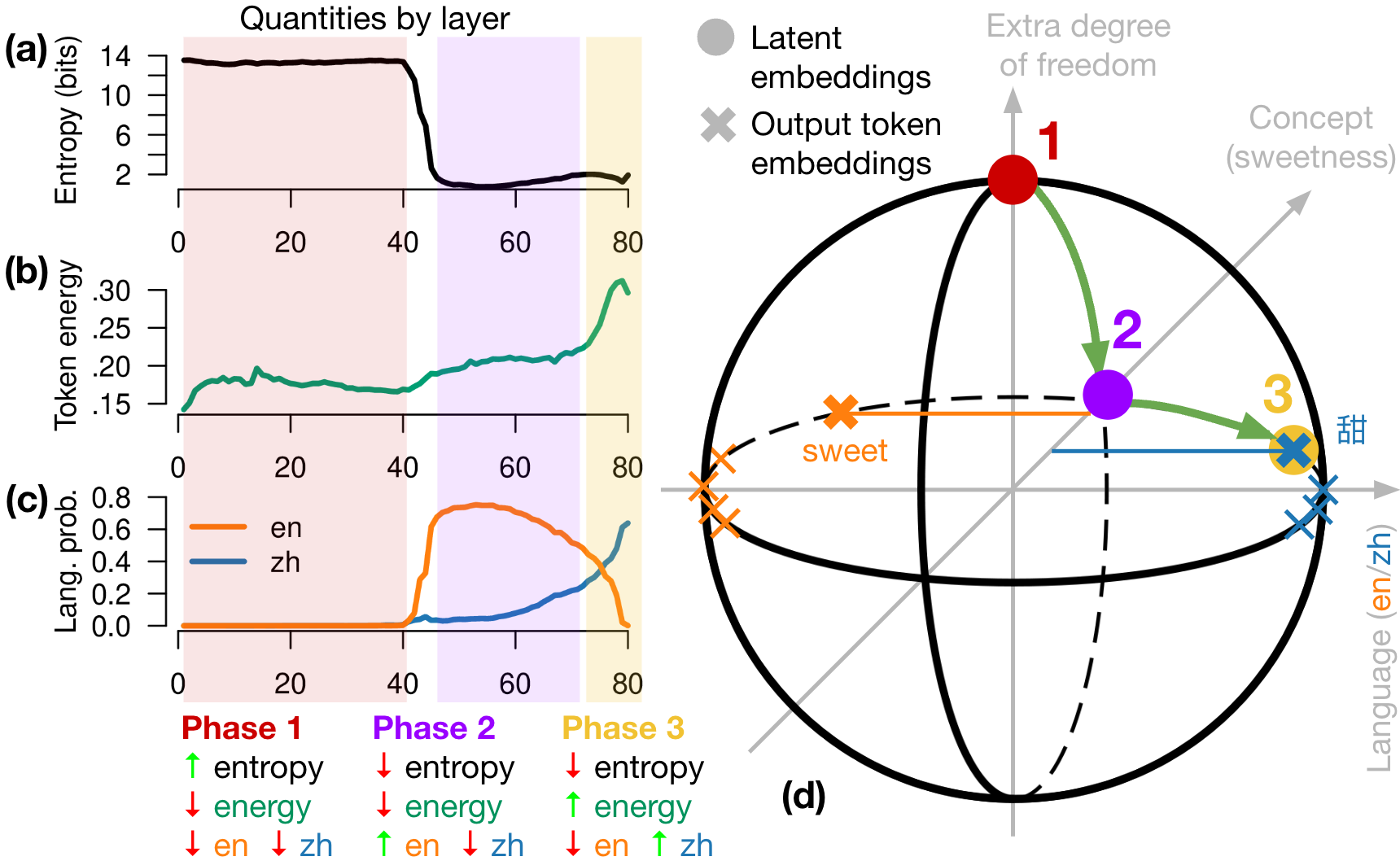}
    \caption{\textbf{Anatomy of transformer forward pass} when translating to Chinese (\cf\ \Secref{sec:data}).
    Layer-by-layer evolution of
    \textbf{(a)}~entropy of next\hyp token distribution,
    \textbf{(b)}~token energy,
    \textbf{(c)}~language probabilities.
    As latents are transformed layer by layer, they go through three phases (\Secref{sec:geometric view}), 
    \textbf{(d)}~traveling on a hypersphere, here in 3D instead of actual 8192D (\Secref{sec:theory}). ``\zh{甜}'' means ``sweet''.}
    \label{fig:sphere}
\end{figure}

We visualize token energy and its relation to other key quantities in \Figref{fig:sphere}.
As a function of layer (\Figref{fig:sphere}(b)), root mean squared token energy is low (around 20\%) and mostly flat
% (with a slight increase around layer 40)
before layer 70, when it suddenly spikes---just when next\hyp token predictions switch from English to Chinese (\Figref{fig:sphere}(c)).
In sum, \Figref{fig:sphere}(a--c) reveals three phases:
\begin{enumerate}
\denselist
    \item \textbf{Phase 1} (layers 1--40):
    High entropy (14 bits, nearly uniform),
    low token energy,
    no language dominates.
    \item \textbf{Phase 2} (layers 41--70):
    Low entropy (1--2 bits),
    low token energy,
    English dominates.
    \item \textbf{Phase 3} (layers 71--80):
    Low entropy,
    high token energy (up from 20\% to 30\%),
    Chinese dominates.
\end{enumerate}

% Remarkably, even for the simple task of repeating an English input word, this three-phase pattern holds (only that, as expected, there English dominates also in Phase~3; \cf\ \Figref{fig:copy}(b), \Figref{fig:copy-energy}(b)).

\section{Conceptual model}
\label{sec:theory}

Next, we formulate a conceptual model that is consistent with the above observations.

In order to predict the next token, the transformer's job essentially consists in mapping the input embedding of the current token to the output embedding of the next token.
\textbf{Phase~1} is focused on building up a better feature representation for the current token from its input embedding,
by dealing with tokenization issues (\eg, integrating preceding tokens belonging to the same word), integrating words into larger semantic units, \etc
This phase is not yet directly concerned with predicting the next token, with latents remaining largely orthogonal to output token space (low token energy), leading to small dot products between latents and output token embeddings, and thus to high entropy.

In \textbf{Phase 2,} latents live in an abstract ``concept space'', which, unlike in Phase~1, is no more orthogonal to the output token space.
Rather, latent ``concept embeddings'' are closer to those output token embeddings that can express the respective concept (across languages, synonyms, \etc), leading to low entropy.
Among the concept\hyp relevant tokens, English variants lie closer to the concept embedding than non-English variants (due to the model's overwhelming exposure to English during training), leading to higher probabilities for English than Chinese tokens.
Despite the correlation between concept and token embeddings, concept embeddings also carry much information that goes beyond output tokens (including input\hyp specific contextual information and information about the target language), leading to a still-low token energy.

In \textbf{Phase 3,} the model maps abstract concepts to concrete words\slash tokens in the target language. Information that is irrelevant for next\hyp token prediction is discarded, leading to a spike in token energy.

\xhdr{Sketch}
This model is illustrated---with a strongly simplified toy-like sketch---in \Figref{fig:sphere}(d).
In this picture, the model operates in 3D (rather than the actual 8192D) space.
All embeddings (output tokens and latents) lie on a sphere around the origin.
Token embeddings lie on the equator and are mostly spread out along the $x$-axis (left\slash right), which captures language (English left, Chinese right).
The $y$-axis (front\slash back) captures concepts, in this toy picture along a 1D ``sweetness'' scale.
The $z$-axis (bottom\slash top) provides an extra degree of freedom that can be used to store information about context, language, \etc
A transformer forward pass moves along the surface of the sphere.
In Phase~1, the latent starts out at the north pole, orthogonal to both output token and concept embeddings.
Phase~2 rotates the latent into concept space; English tokens are more likely because their embeddings have a stronger concept component~$y$.
Finally, Phase~3 rotates the latent along the equator into the target language's hemisphere, onto the output token that best captures the active concept in that language.

%%%%%%%%%%%%%%%%%%%%%%%%%%%%%%%%%%%%%%%
\section{Discussion}
\label{sec:discussion}
%%%%%%%%%%%%%%%%%%%%%%%%%%%%%%%%%%%%%%%

In our attempt to answer whether \llama models internally use English as a pivot language, we found that latent embeddings indeed lie further from the correct next token in the input language than from its English analog,
leading to overwhelmingly English internal representations as seen through the logit lens.
It might thus be tempting to conclude that, yes, \llama uses English as an implicit pivot, similar to researchers' prior use of English as an explicit pivot
\cite{shi2022language, ahuja2023mega, huang2023languages}.
But our answer must be more nuanced, as much of the latents' ``energy'' points in directions that are largely orthogonal to output token embeddings and thus do not matter for next\hyp token prediction.
The model can use these directions as extra degrees of freedom for building rich feature representations from its raw inputs 
% lecun1998gradient, 
\cite{yosinski2014transferable, yosinski2015understanding, geva2022transformer},
which could be seen as forming an abstract ``concept space''.
In this interpretation, the model's internal lingua franca is not English but concepts---concepts that are biased toward English.
Hence, English could still be seen as a pivot language, but in a semantic, rather than a purely lexical, sense.

Our experiments involve three text completion tasks.
The translation and cloze tasks operate at a semantic level,
% and yield qualitatively identical results,
whereas the word repetition task is purely syntactic.
% \footnote{- mention procrustes results: for copying, rotating would suffice}
Yet, in most languages (\Figref{fig:copy}) the pattern is similar to that for the two other tasks, with tokens first going through an ``English phase''---possibly because recognizing that the task is to simply copy a token requires semantic understanding, which is achieved only in concept space, which in turn is closer to English token embeddings.

This said, note that the English\hyp first pattern is less pronounced on the repetition task (\Figref{fig:copy}), where the input language rises earlier than on the other tasks or, for Chinese (\Figref{fig:copy}(e)) even simultaneously with, or faster than, English.
This might be due to tokenization: for Chinese we explicitly chose 100\% single\hyp token words, as opposed to only 13\% for Russian, 43\% for German, and 55\% for French (\Tabref{tab:translation-aggr-sizes}).
% With this in mind, \Figref{fig:copy} shows that languages with higher rates of single\hyp token words stray further from the English\hyp first pattern.
% In other words,
Where language\hyp specific tokens are available, the detour through English seems less pronounced.
This supports prior concerns about the importance of tokenization, which not only burdens minority languages with more tokens per word \cite{Artetxe_2020},
% translating to higher monet costs and more difficult seq pred probs.
but, as we show, also
forces latents through an English\hyp biased semantic space.
% - above obs also suggest that, when a lang doesn't have a token for capturing a cpt (as is the case for most langs and most words w/ \llama's tokenizer), this may lead to model taking detour

Future work should investigate in what ways an English bias in latent space could be problematic,
% \eg, via a ``Sapir--Whorf effect'' whereby the available concepts 
\eg, by biasing downstream model behavior.
We see promise in designing experiments building on work from psycholinguistics, which has shown that concepts may carry different emotional values in different languages \cite{Boroditsky2003SexSyntaxSemantics} and that using one word for two concepts (colexification) may affect cognition \cite{di2021colexification}.
Future work should also study how English bias changes when decreasing the dominance of English during training,
% and possibly develop a ``latent EN bias score'' based on our metrics
\eg, by applying our method to \llama derivatives with a different language mix \cite{goddard2023polyglot,pluster2023leolm,huang2023vigogne,kim2023kor},
or by using less Anglocentric tokenizers.

% multilingual LLMs \cite{goddard2023polyglot, openbuddy2023} as well as LLMs for individual languages \cite{pluster2023leolm,clibrain2023es,huang2023vigogne,kim2023kor}.
% including German~\cite{pluster2023leolm}, Spanish~\cite{clibrain2023es}, French~\cite{huang2023vigogne}, and Korean~\cite{kim2023kor}.

Such work will give important clues for decreasing English bias and enabling more equitable AI.

\section*{Limitations}
% - we only use llama 
% - our tasks are simple. this is a feature in our context, and essential as a first step. nonetheless, future work should extend to more complex tasks or even general text.
% - although we speak of "concept space" in our interpretation, we have a clear sense yet of the structure of this space. mapping it out is a future project of its own.

% The translation, repetition, and cloze tasks

In this paper, we focus on the \llama family of language models, which limits the claims we can make about other English-dominated models (but see Appendix~\ref{app:mistral} for initial evidence that Mistral-7B behaves identically). Moreover, since the proposed method relies on model parameters, little can be said about the more widely used closed\hyp source models. Nonetheless, the methods outlined in this paper can be straightforwardly applied to other autoregressive transformers and generalized to non-autoregressive ones (given their parameters are available), a direction that warrants future exploration. 

Additionally, the tasks outlined in the paper are simple and provide a highly controlled, yet toy-like, context for studying the internal language of LLMs. This is essential as a first step to illustrate existence, but future work should extend to a wider range of tasks; these may include more culturally sensitive problems, popular use-cases (\cf \Secref{sec:discussion}), and technical analyses that go beyond single tokens. 

While we find evidence of a ``concept space'' in our interpretation (\Secref{sec:theory}), we have limited understanding of the structure of this space in its original high-dimensional form.
% In turn, the three-dimensional intuition we form may not hold in the original space.
We believe that better understanding and mapping out this concept space is an important future direction and will result in a stronger basis for the presented conceptual model. 

% Finally, while The logit lens also employs the language modeling head from the last layer without additional training 
% to the latents at the intermediate layers. Thus, the decoded probabilities are at best an approximation of the true beliefs of the model after each layer, the further away from the 
% final layer the more so (which is also visible from Fig.~\ref{fig:translation-llama2-7b-ex-single}).  

Finally, while the logit lens grants us approximate access to the internal beliefs about what should be the output at a given sequence position, everything else contained 
in the intermediate representations (e.g., information to construct keys, queries, values, or to perform intermediate calculations that do not directly contribute to the output 
beliefs) remains hidden and only enters the logit lens--based part of our analysis as noise.

% As mentioned in \Secref{sec:Related work}, the tuned lens~\cite{belrose2023eliciting} is a recent method that addresses this issue by learning one additional affine mapping per layer, translating the intermediate representation into one more suitable for decoding the \textit{true} next token. However, because we are interested precisely in the intermediate decodings, in particular, whether their decoded language differs from 
% the language of the \textit{true} next token, and not in skipping layers, we stick to the original logit lens in this work.

\section*{Acknowledgements}
\small{
We thank Nina \citet{rimsky2023decoding} for sharing her \llama{} wrapper and logit lens 
implementation;\footnote{\url{https://github.com/nrimsky/LM-exp/blob/main/intermediate_decoding/intermediate_decoding.ipynb}}
% which was the starting point for our analysis. 
Lucia Quirke for inputs on mechanistic interpretability, on our experimental setup, and for a fruitful discussion; 
Saibo Geng for helping us with the Chinese dataset;
Nicola Cancedda,
David Garcia,
Eric Horvitz,
Manoel Horta Ribeiro,
Maxime Peyrard,
Saibo Geng,
Tim Davidson,
Valentin Hartmann,
and Zachary Horvitz
for insightful discussions and feedback;
% over lunch and the key-input on one of our earlier experiments that the majority of the tokens in the vocabulary of \llama{}'s tokenizer are dedicated for English, which needs to be taken into account when computing language probabilities. 
and Meta for open\hyp sourcing \llama{} and thereby helping democratize LLM research.
Finally, we thank our anonymous peer reviewers for their productive input, which has led, among others, to Appendices~\ref{sec:estonian} and \ref{app:mistral}.
West's lab is partly supported by grants from
Swiss National Science Foundation (200021\_185043, TMSGI2\_211379),
Swiss Data Science Center (P22\_08),
H2020 (952215),
and by generous gifts from Meta, Google, and Microsoft.
}

\bibliography{00MainPaper}

\appendix

\section{Additional methodological details} 
\label{app:info}

\subsection{Word translation}
\label{app:Word translation}

A detail that we omitted in the main paper for brevity is how we translate the English words resulting from the procedure outlined in \Secref{sec:data} to French, German, and Russian. During these translations we translated both the individual words alongside their cloze sentences using DeepL.\footnote{\url{https://www.deepl.com/translator}}
For each word translation, we include the context of the cloze task to disambiguate homonyms. 
%These in-context translations can fail, which is why we do not have the same number of examples for each language and each language pair. 
We then filter the translations to remove words that have the same prefix token across English and the target language. For example, the French translation of the word ``photograph'', ``photographier'', shares the ``photo'' prefix token. Additionally, we parse through the translations and filter any cloze translations where the target word doesn't align with the expected word from the individual word translation, which was due to failures in the DeepL translation. These filterings result in a different number of final words across the different languages. 

We provide the numbers for the aggregated translation task~(\Tabref{tab:translation-aggr-sizes}), repetition task~(\Tabref{tab:copy-sizes}), cloze-task~(\Tabref{tab:cloze-sizes}), and individual translation tasks~(\Tabref{tab:translation_stats}). 

\begin{table}[ht]
\centering
\scriptsize
\begin{tabular}{@{}lcc@{}}
\toprule
                 & Total & Single Token \\ \midrule
 de & 287   & 126           \\
 fr & 162   &  88          \\
 ru & 324   &  45          \\
 zh & 353   &  353         \\\bottomrule
\end{tabular}
\caption{Aggregated translation task dataset sizes.}
\label{tab:translation-aggr-sizes}
\end{table}

\begin{table}[ht]
\centering
\scriptsize
\begin{tabular}{@{}lcc@{}}
\toprule
        & Total & Single Token \\ \midrule
 de          & 104   & 45           \\
 en          & 132   & 132          \\
 fr          & 56    & 31           \\
 ru          & 115   & 15           \\
 zh          & 139   & 139          \\\bottomrule
\end{tabular}
\caption{Repetition task dataset sizes.}
\label{tab:copy-sizes}
\end{table}

\begin{table}[ht]
\centering
\scriptsize
\begin{tabular}{@{}lcc@{}}
\toprule
        & Total & Single Token \\ \midrule
\ \ \ de          & 104   & 45           \\
\ \ \ en          & 132   & 132          \\
\ \ \ fr          & 56    & 31           \\
\ \ \ ru          & 115   & 15           \\
\ \ \ zh          & 139   & 139          \\\bottomrule
\end{tabular}
\caption{Cloze task dataset sizes.}
\label{tab:cloze-sizes}
\end{table}

\begin{table}[ht]
    \centering
    \scriptsize
    \begin{tabular}{lccccc}
\toprule
     & {de} & {en} & {fr} & {ru} & {zh} \\
    \midrule
    {de} & -- & 120 (120) & 56 (31) & 105 (15) & 120 (120) \\
    {en} & 104 (45) & -- & 57 (31) & 114 (15) & 132 (132) \\
    {fr} & 93 (40) & 118 (118) & -- & 104 (15) & 118 (118) \\
    {ru} & 90 (41) & 114 (114) & 49 (26) & -- & 115 (115) \\
    {zh} & 104 (45) & 132 (132) & 57 (31) & 115 (15) & -- \\
    \bottomrule
    \end{tabular}
    \caption{Translation statistics between languages, including total numbers and single\hyp token translations (in brackets).}
    \label{tab:translation_stats}
\end{table}

\subsection{Computing language probabilities}
\label{app:Computing language probabilities}

In order to compute language probabilities, we search \llama{}'s vocabulary for all tokens that could be the first token of the correct word in the respective language. In particular, we search \llama{}'s vocabulary for all prefixes of the word without and with leading space.\footnote{Represented by ``\_''.}
For Chinese and Russian we also consider tokenizations based on the UTF-8 encodings of their unicode characters. For a language $\ell$ and its corresponding target word $w$, we define 
\begin{equation}
    P(\text{lang} = \ell) := \sum_{t_{\ell} \in \text{Start}(w)} P(x_{n+1} = t_{\ell}),
\end{equation} 
where $\text{Start}(w)$ denotes the set of starting tokens of the word $w$. 

For example, if the correct next Chinese word is ``\zh{花}'' (``flower''), which can be tokenized either using the single token ``\zh{花}'' or via its UTF-8 encoding ``<0xE8>$\cdot$<0x8A>$\cdot$<0xB1>'', we have 
$P(\text{lang} = \lang{ZH}) = P(x_{n+1} = \text{``\zh{花}''}) + P(x_{n+1} = \text{``<0xE8>''})$ and 
$P(\text{lang} = \lang{EN}) =  P(x_{n+1} = \text{``f''}) + P(x_{n+1} = \text{``fl''}) + P(x_{n+1} = \text{``flow''}) + P(x_{n+1} = \text{``\_f''}) + P(x_{n+1} = \text{``\_fl''}) + P(x_{n+1} = \text{``\_flo''}) + P(x_{n+1} = \text{``\_flow''}) + P(x_{n+1} = \text{``\_flower''})$ (all the token-level prefixes of ``flower'' and ``\_flower'').

\section{Additional results}
\label{app:results}

Here we provide the results for all languages: Chinese, English, French, German, and Russian.

\begin{figure*}[ht!]
     \centering
     \begin{subfigure}[b]{0.3\textwidth}
         \centering
         \includegraphics[width=\textwidth]{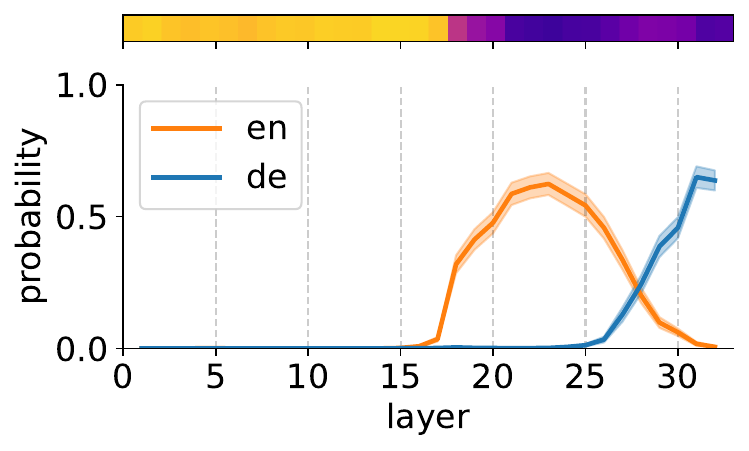}
     \end{subfigure}
     \hfill
     \begin{subfigure}[b]{0.28\textwidth}
         \centering
         \caption{Translation (-> DE)}
         \includegraphics[width=\textwidth]{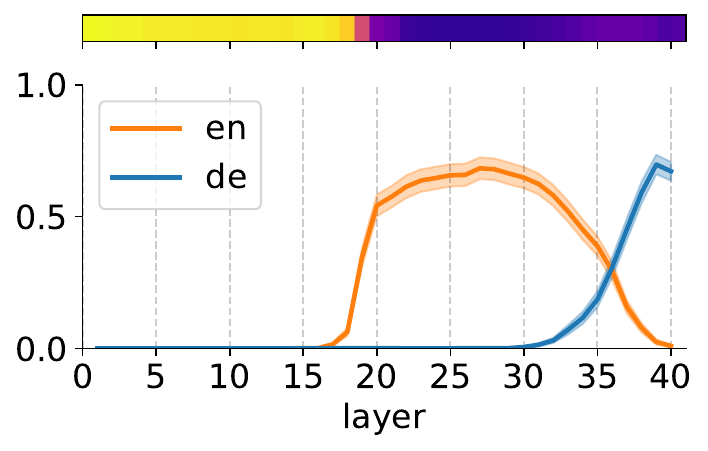}
     \end{subfigure}
     \hfill
     \begin{subfigure}[b]{0.343\textwidth}
         \centering
         \includegraphics[width=\textwidth]{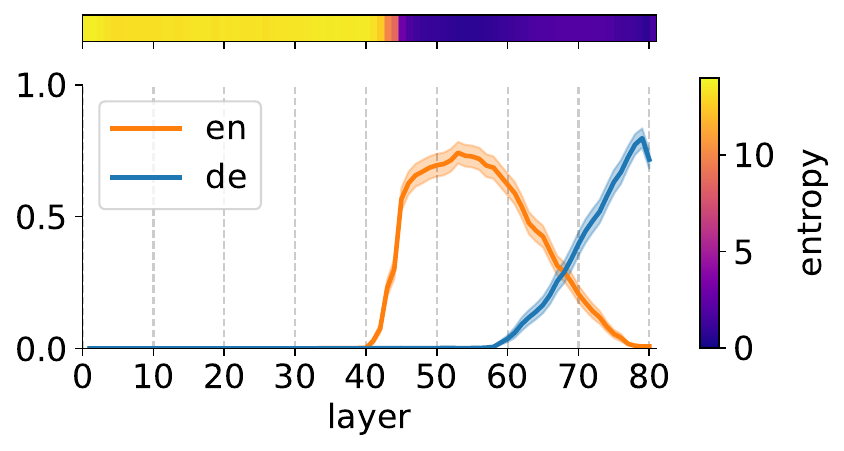}
     \end{subfigure}

     \begin{subfigure}[b]{0.3\textwidth}
         \centering
         \includegraphics[width=\textwidth]{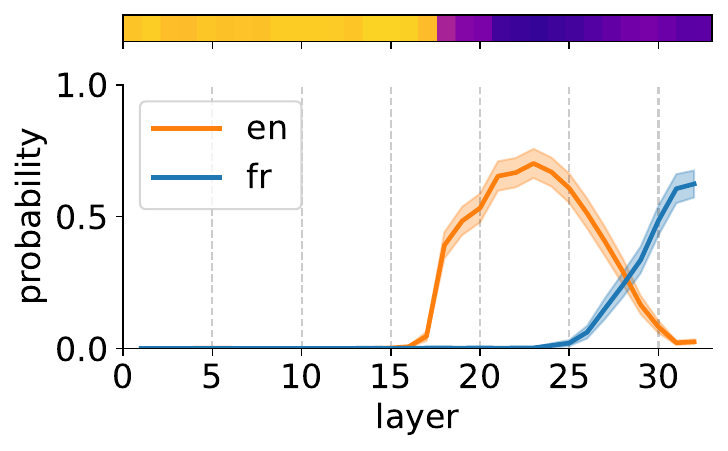}
     \end{subfigure}
     \hfill
     \begin{subfigure}[b]{0.28\textwidth}
         \centering
         \caption{Translation (-> FR)}
         \includegraphics[width=\textwidth]{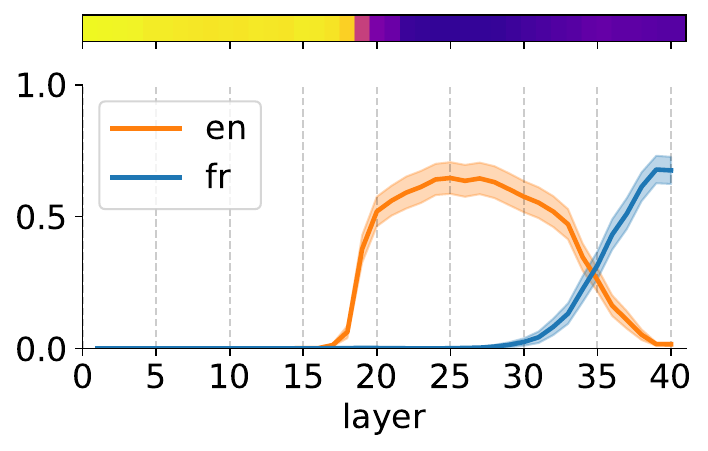}
     \end{subfigure}
     \hfill
     \begin{subfigure}[b]{0.343\textwidth}
         \centering
         \includegraphics[width=\textwidth]{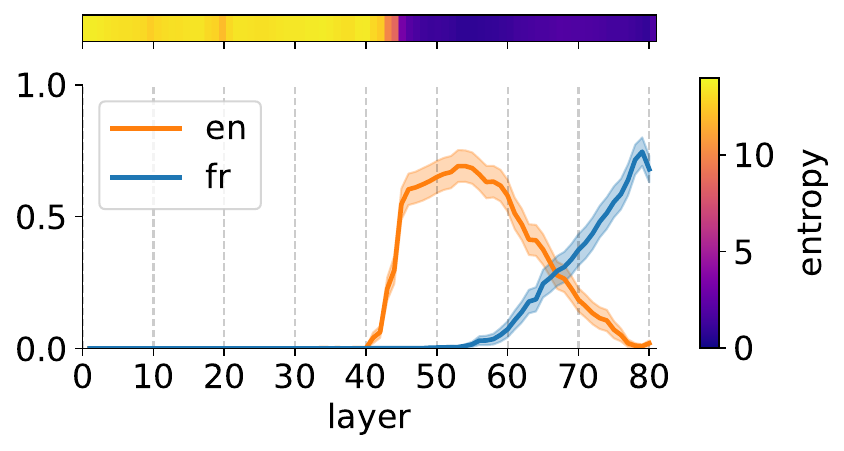}
     \end{subfigure}

     \begin{subfigure}[b]{0.3\textwidth}
         \centering
         \includegraphics[width=\textwidth]{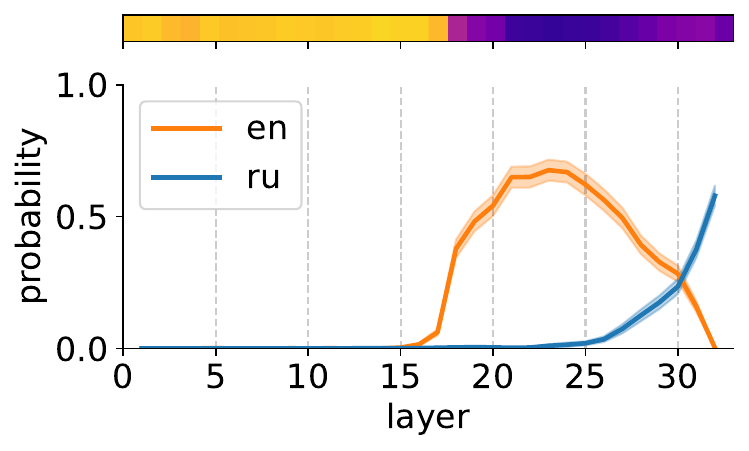}
     \end{subfigure}
     \hfill
     \begin{subfigure}[b]{0.28\textwidth}
         \centering
         \caption{Translation (-> RU)}
         \includegraphics[width=\textwidth]{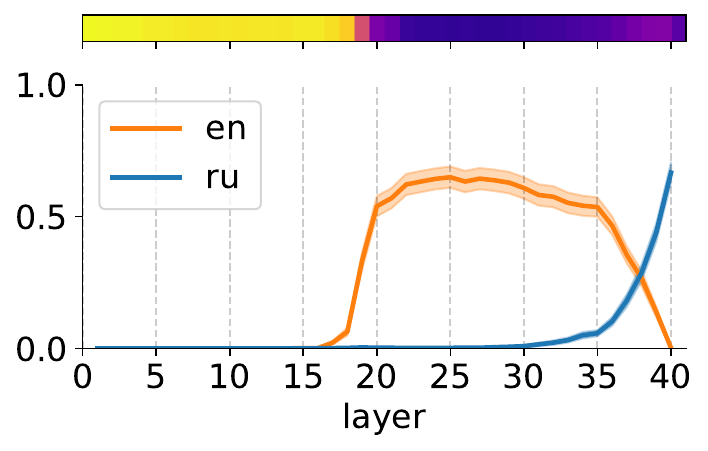}
     \end{subfigure}
     \hfill
     \begin{subfigure}[b]{0.343\textwidth}
         \centering
         \includegraphics[width=\textwidth]{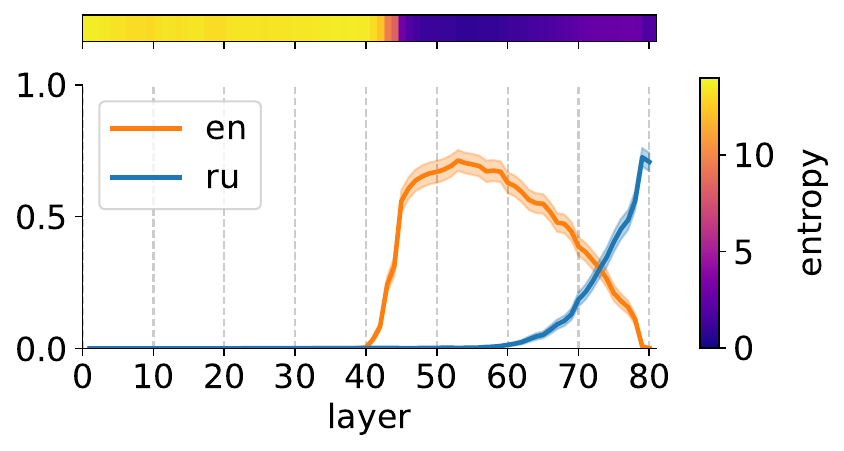}
     \end{subfigure}

     \begin{subfigure}[b]{0.3\textwidth}
         \centering
         \includegraphics[width=\textwidth]{figures/translation_aggr/7b_zh_probas_ent_aggr.pdf}
     \end{subfigure}
     \hfill
     \begin{subfigure}[b]{0.28\textwidth}
         \centering
         \caption{Translation (-> ZH)}
         \includegraphics[width=\textwidth]{figures/translation_aggr/13b_zh_probas_ent_aggr.pdf}
     \end{subfigure}
     \hfill
     \begin{subfigure}[b]{0.343\textwidth}
         \centering
         \includegraphics[width=\textwidth]{figures/translation_aggr/70b_zh_probas_ent_aggr.pdf}
     \end{subfigure}
\caption{Figures illustrate the translation task where \llama{} 7B, 13B, and 70B are tasked with translating a word from all non-English input languages to output language. There is one column per model size. The x-axis shows the layer number of the model, and the y-axis the total probability mass falling on the correct token across languages. The orange line illustrates the probability of the correct target word in English and the blue line shows it for the non-English output language. We do not include the probability the input language since it is zero throughout. Means and 95\% Gaussian confidence intervals have been computed over the input examples, numbers in \Appref{app:info}.}
     \label{fig:translation}
 \end{figure*}

\begin{figure*}[ht!]
     \centering
     \begin{subfigure}[b]{0.3\textwidth}
         \centering
         \includegraphics[width=\textwidth]{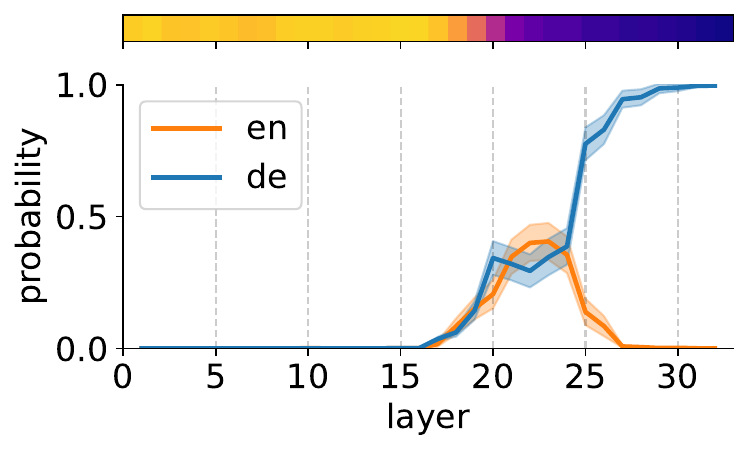}
     \end{subfigure}
     \hfill
     \begin{subfigure}[b]{0.28\textwidth}
         \centering
         \caption{Repetition (DE)}
         \includegraphics[width=\textwidth]{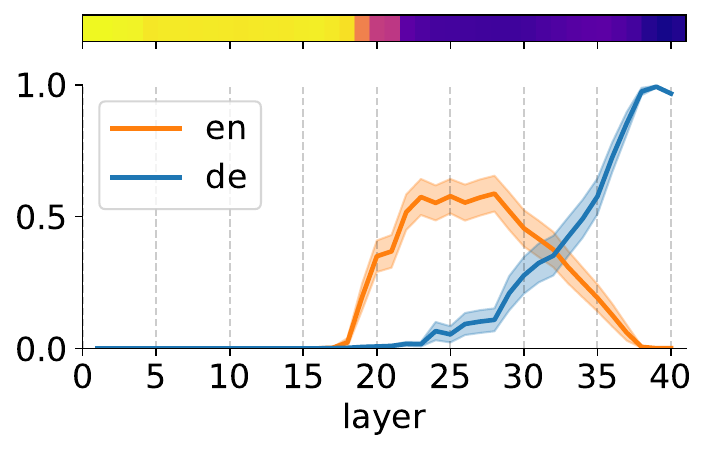}
     \end{subfigure}
     \hfill
     \begin{subfigure}[b]{0.343\textwidth}
         \centering
         \includegraphics[width=\textwidth]{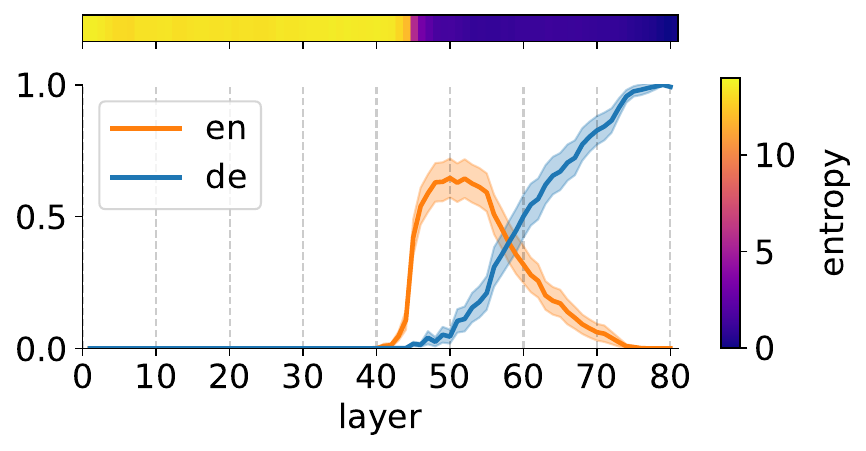}
     \end{subfigure}

     \centering
     \begin{subfigure}[b]{0.3\textwidth}
         \centering
         \includegraphics[width=\textwidth]{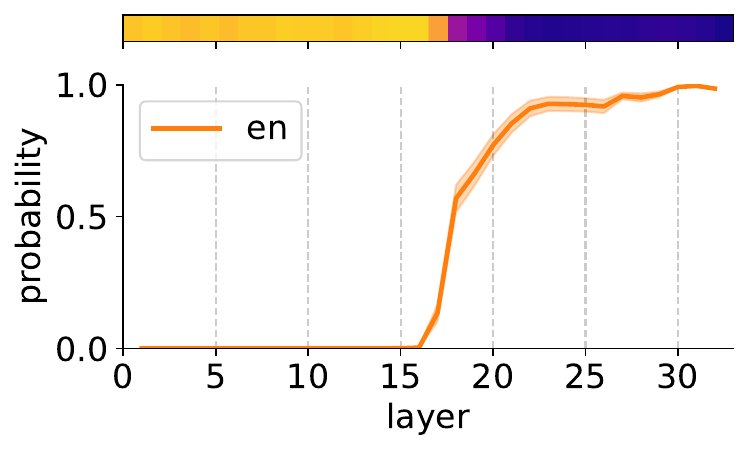}
     \end{subfigure}
     \hfill
     \begin{subfigure}[b]{0.28\textwidth}
         \centering
         \caption{Repetition (EN)}
         \includegraphics[width=\textwidth]{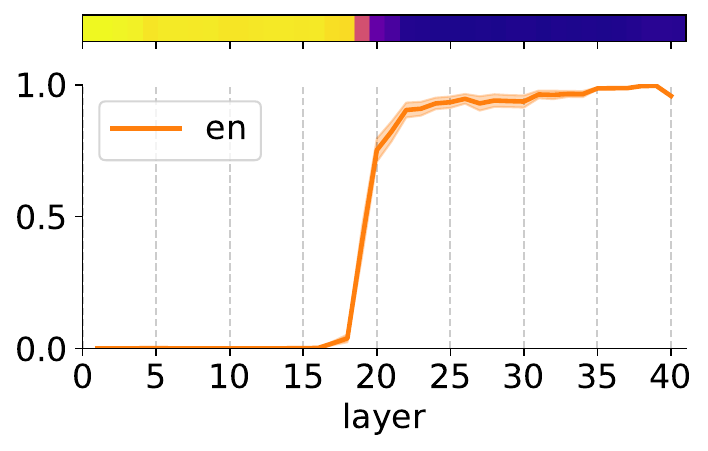}
     \end{subfigure}
     \hfill
     \begin{subfigure}[b]{0.343\textwidth}
         \centering
         \includegraphics[width=\textwidth]{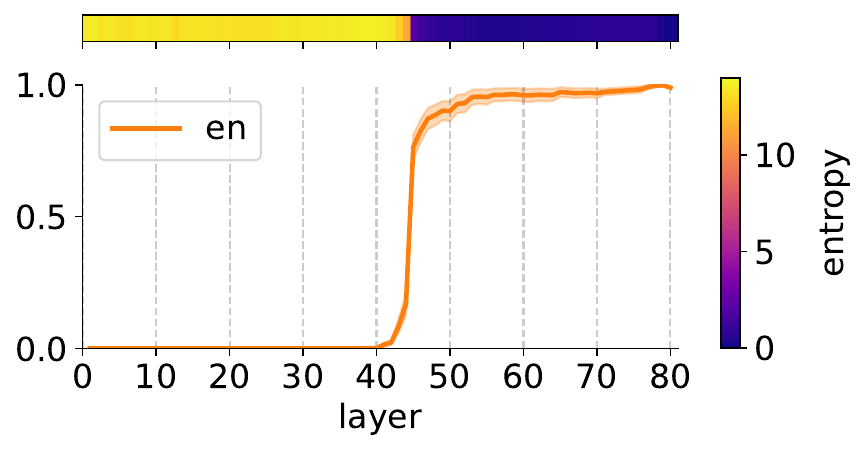}
     \end{subfigure}

     \begin{subfigure}[b]{0.3\textwidth}
         \centering
         \includegraphics[width=\textwidth]{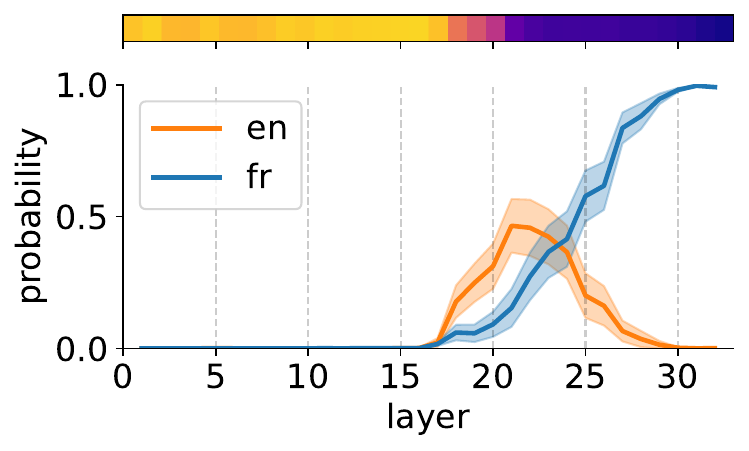}
     \end{subfigure}
     \hfill
     \begin{subfigure}[b]{0.28\textwidth}
         \centering
         \caption{Repetition (FR)}
         \includegraphics[width=\textwidth]{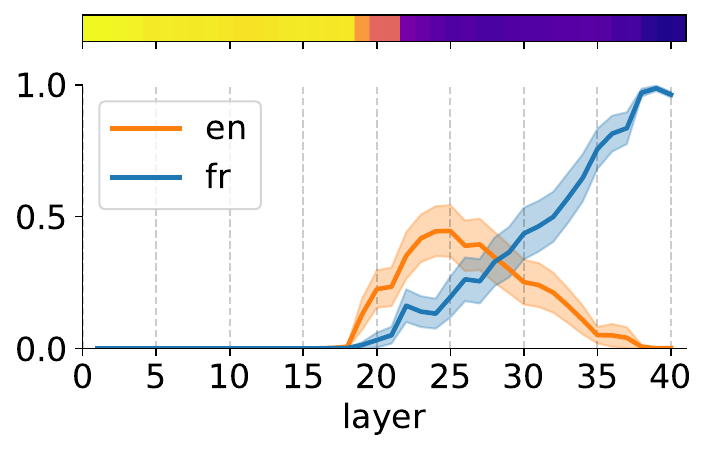}
     \end{subfigure}
     \hfill
     \begin{subfigure}[b]{0.343\textwidth}
         \centering
         \includegraphics[width=\textwidth]{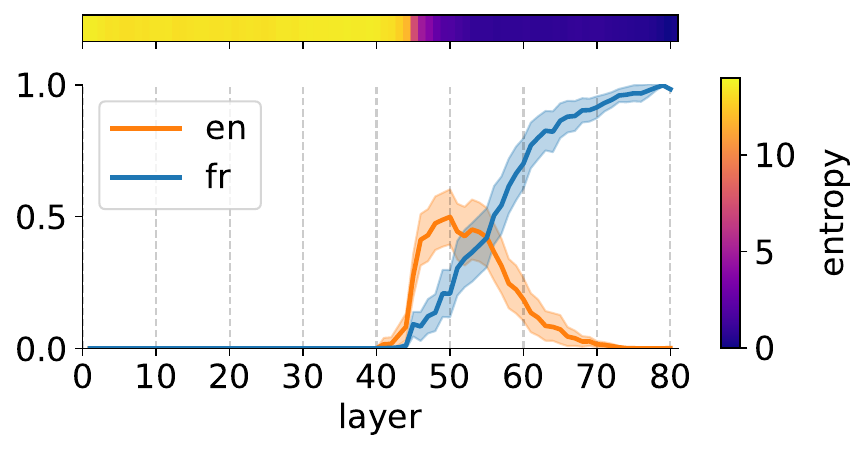}
     \end{subfigure}

     \begin{subfigure}[b]{0.3\textwidth}
         \centering
         \includegraphics[width=\textwidth]{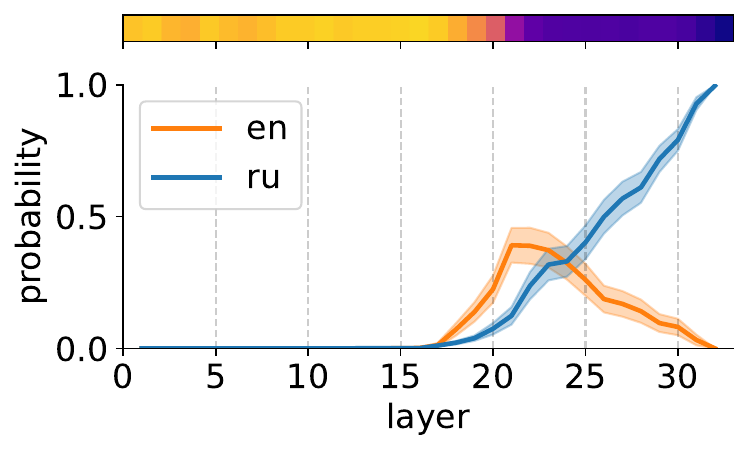}
     \end{subfigure}
     \hfill
     \begin{subfigure}[b]{0.28\textwidth}
         \centering
         \caption{Repetition (RU)}
         \includegraphics[width=\textwidth]{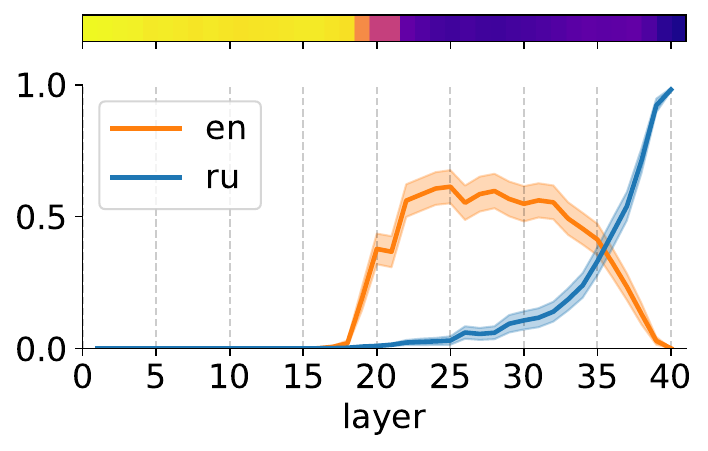}
     \end{subfigure}
     \hfill
     \begin{subfigure}[b]{0.343\textwidth}
         \centering
         \includegraphics[width=\textwidth]{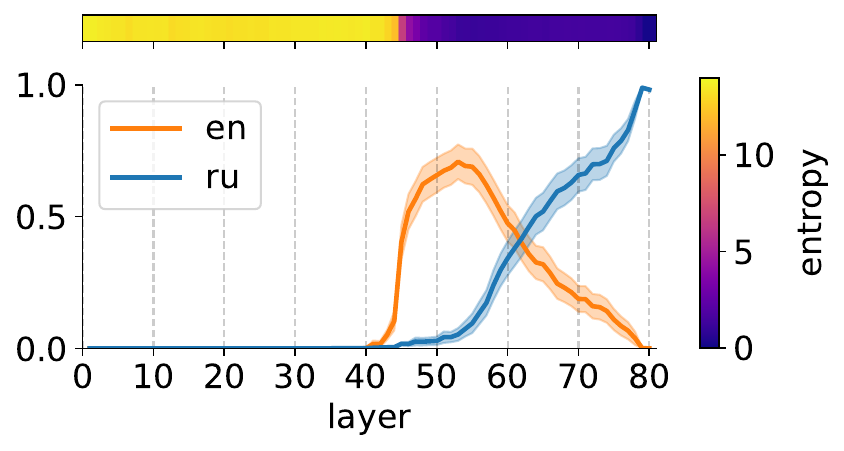}
     \end{subfigure}

     \begin{subfigure}[b]{0.3\textwidth}
         \centering
         \includegraphics[width=\textwidth]{figures/same/7b_zh_zh_probas_ent.pdf}
     \end{subfigure}
     \hfill
     \begin{subfigure}[b]{0.28\textwidth}
         \centering
         \caption{Repetition (ZH)}
         \includegraphics[width=\textwidth]{figures/same/13b_zh_zh_probas_ent.pdf}
     \end{subfigure}
     \hfill
     \begin{subfigure}[b]{0.343\textwidth}
         \centering
         \includegraphics[width=\textwidth]{figures/same/70b_zh_zh_probas_ent.pdf}
     \end{subfigure}

     \caption{Figures illustrate the repetition task where \llama{} 7B, 13B, and 70B are tasked with copying a non-English word. There is one column per model size. The x-axis shows the layer number of the model, and the y-axis the total probability mass falling on the correct token across languages. The orange line illustrates the probability of the correct target word in English and the blue line shows it for the non-English output language. Means and 95\% Gaussian confidence intervals have been computed over the input examples, numbers in \Appref{app:info}.}
     \label{fig:copy}
 \end{figure*}

\begin{figure*}[ht!]
     \centering
     
     \begin{subfigure}[b]{0.3\textwidth}
         \centering
         \includegraphics[width=\textwidth]{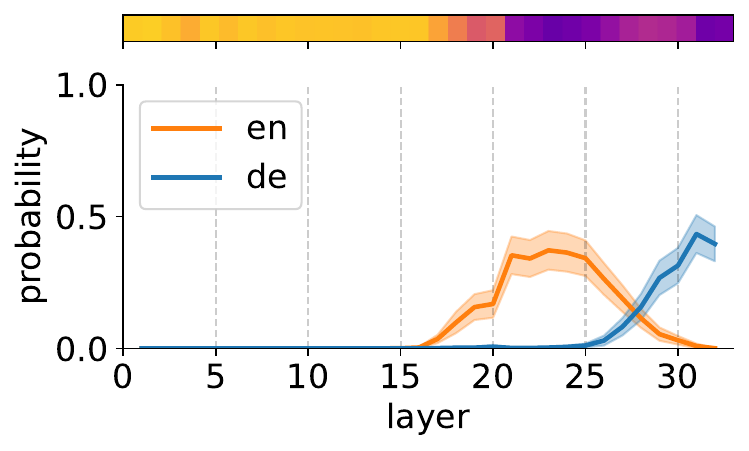}
     \end{subfigure}
     \hfill
     \begin{subfigure}[b]{0.28\textwidth}
         \centering
         \caption{Cloze task (DE)}
         \includegraphics[width=\textwidth]{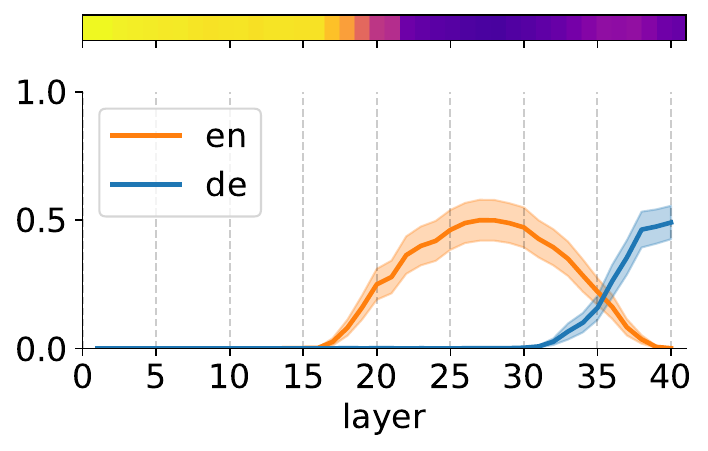}
     \end{subfigure}
     \hfill
     \begin{subfigure}[b]{0.343\textwidth}
         \centering
         \includegraphics[width=\textwidth]{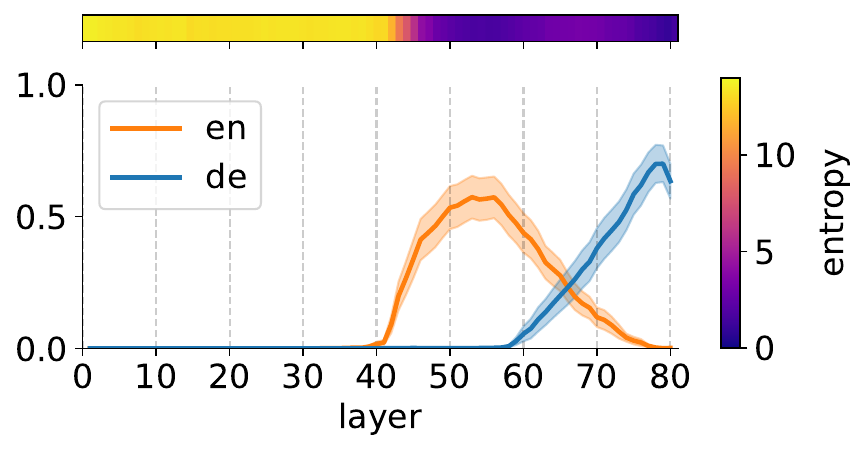}
     \end{subfigure}

      \begin{subfigure}[b]{0.3\textwidth}
         \centering
         \includegraphics[width=\textwidth]{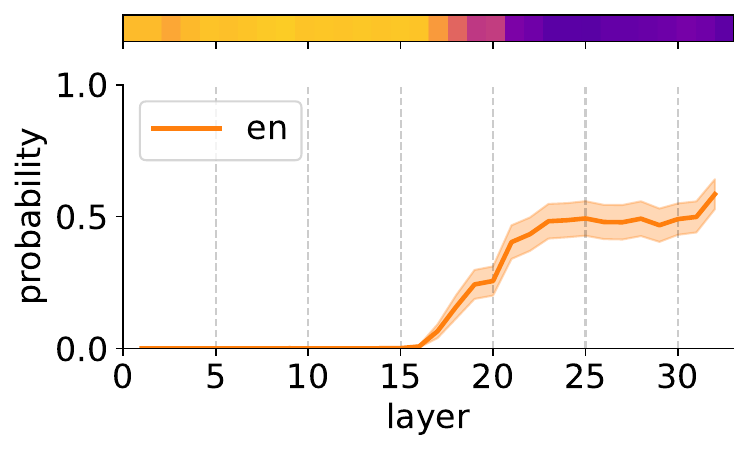}
     \end{subfigure}
     \hfill
     \begin{subfigure}[b]{0.28\textwidth}
         \centering
         \caption{Cloze task (EN)}
         \includegraphics[width=\textwidth]{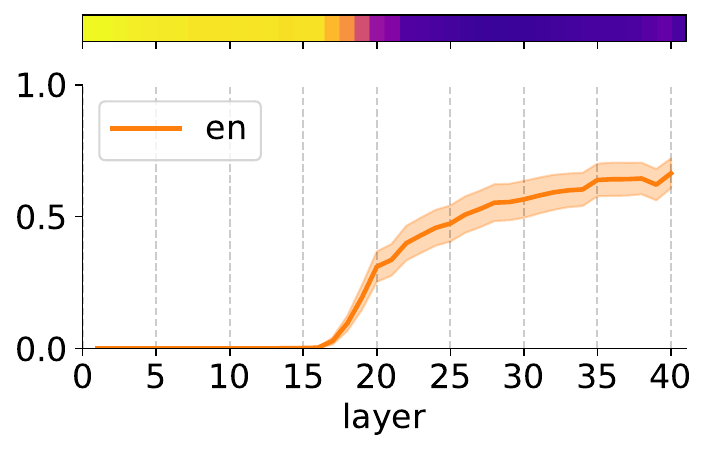}
     \end{subfigure}
     \hfill
     \begin{subfigure}[b]{0.343\textwidth}
         \centering
         \includegraphics[width=\textwidth]{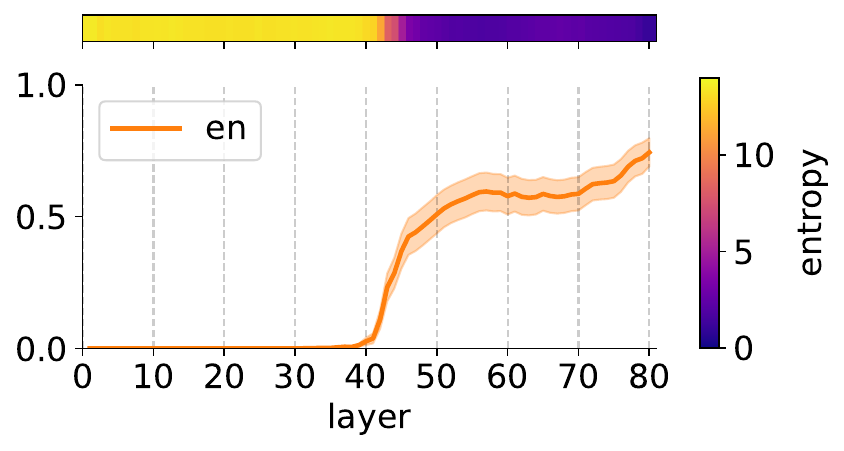}
     \end{subfigure}

     \begin{subfigure}[b]{0.3\textwidth}
         \centering
         \includegraphics[width=\textwidth]{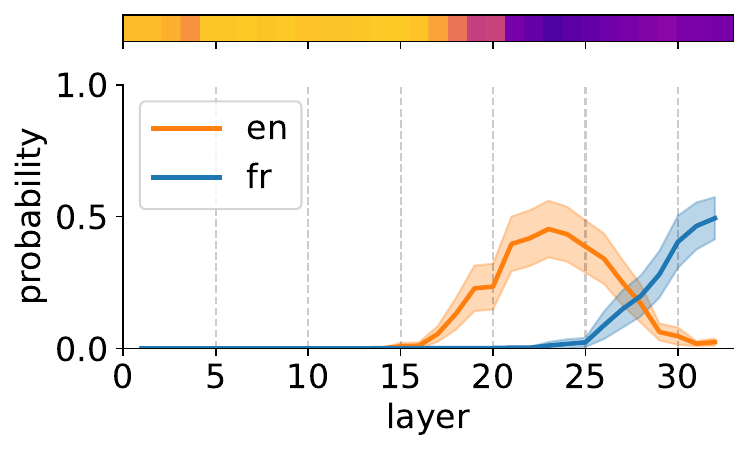}
     \end{subfigure}
     \hfill
     \begin{subfigure}[b]{0.28\textwidth}
         \centering
         \caption{Cloze task (FR)}
         \includegraphics[width=\textwidth]{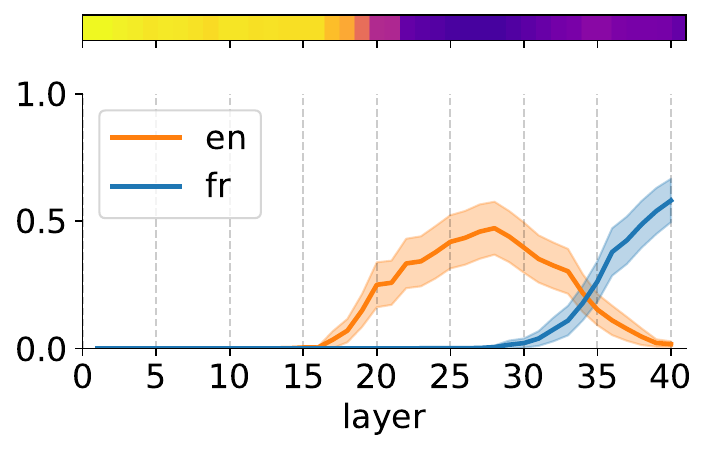}
     \end{subfigure}
     \hfill
     \begin{subfigure}[b]{0.343\textwidth}
         \centering
         \includegraphics[width=\textwidth]{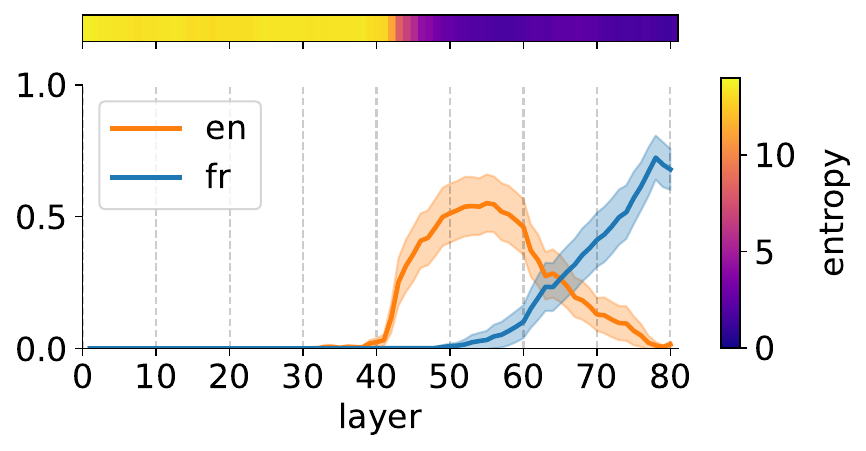}
     \end{subfigure}

     \begin{subfigure}[b]{0.3\textwidth}
         \centering
         \includegraphics[width=\textwidth]{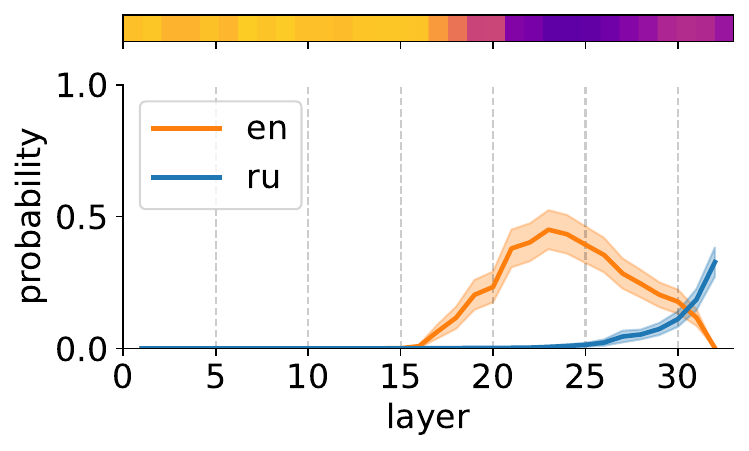}
     \end{subfigure}
     \hfill
     \begin{subfigure}[b]{0.28\textwidth}
         \centering
         \caption{Cloze task (RU)}
         \includegraphics[width=\textwidth]{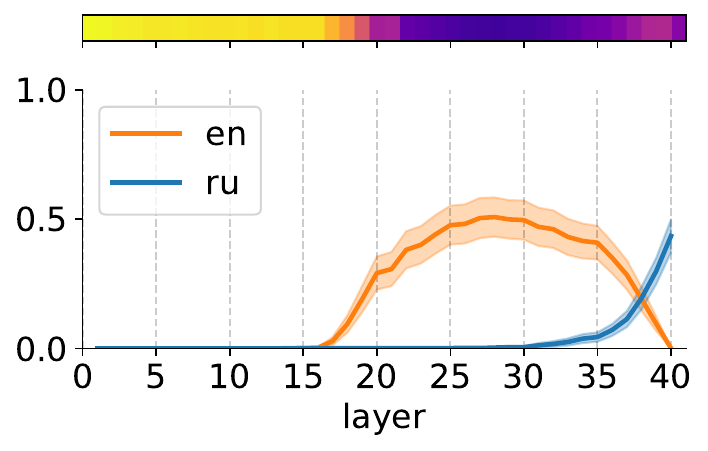}
     \end{subfigure}
     \hfill
     \begin{subfigure}[b]{0.343\textwidth}
         \centering
         \includegraphics[width=\textwidth]{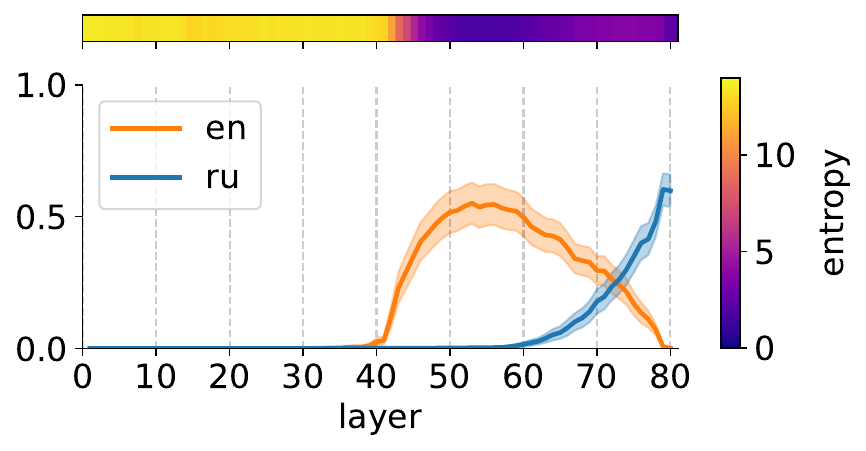}
     \end{subfigure}

     \begin{subfigure}[b]{0.3\textwidth}
         \centering
         \includegraphics[width=\textwidth]{figures/cloze/7b_zh_probas_ent.pdf}
         \caption*{7B}
     \end{subfigure}
     \hfill
     \begin{subfigure}[b]{0.28\textwidth}
         \centering
         \caption{Cloze task (ZH)}
         \includegraphics[width=\textwidth]{figures/cloze/13b_zh_probas_ent.pdf}
         \caption*{13B}
     \end{subfigure}
     \hfill
     \begin{subfigure}[b]{0.343\textwidth}
         \centering
         \includegraphics[width=\textwidth]{figures/cloze/70b_zh_probas_ent.pdf}
         \caption*{70B}
     \end{subfigure}
     \caption{Figures show the same plots only for the cloze task where the correct token is defined in a fill-in-the-blank setting. In the plots, we illustrate the results for German. Means and 95\% Gaussian confidence intervals have been computed over the input examples, numbers in \Appref{app:info}.}
     \label{fig:cloze}
 \end{figure*}

\begin{figure*}[ht!]
     \centering
     \begin{subfigure}[b]{0.3\textwidth}
         \centering
         \includegraphics[width=\textwidth]{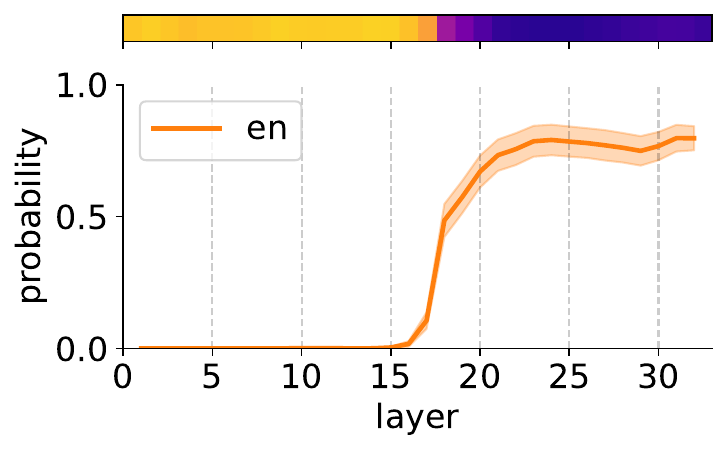}
     \end{subfigure}
     \hfill
     \begin{subfigure}[b]{0.28\textwidth}
         \centering
         \caption{Translation (DE->EN)}
         \includegraphics[width=\textwidth]{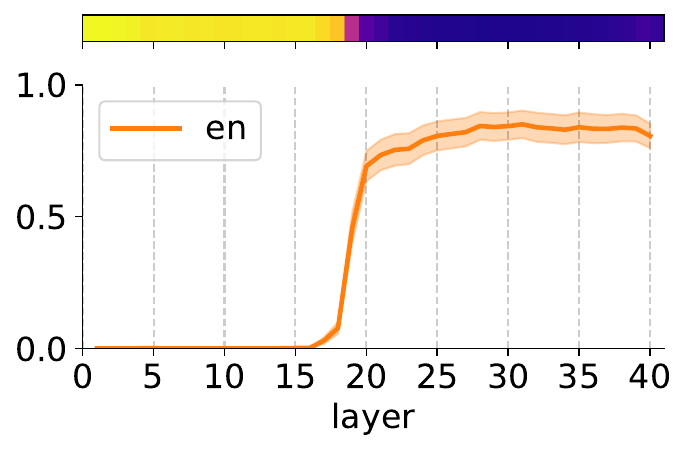}
     \end{subfigure}
     \hfill
     \begin{subfigure}[b]{0.343\textwidth}
         \centering
         \includegraphics[width=\textwidth]{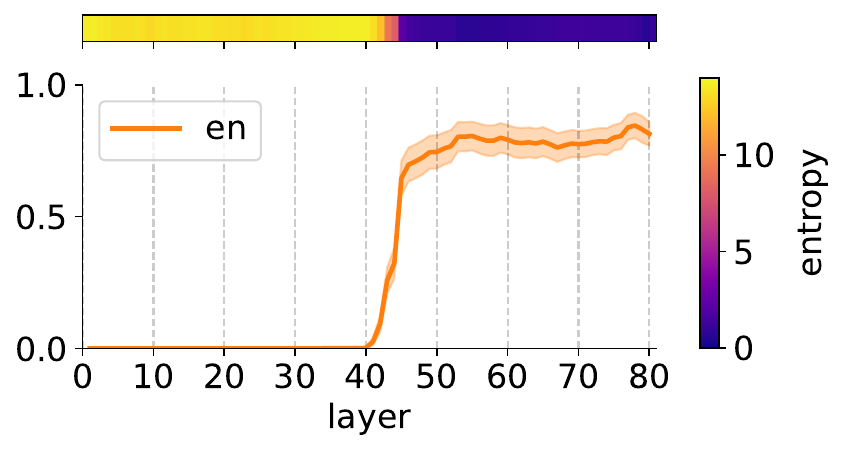}
     \end{subfigure}
     
     \begin{subfigure}[b]{0.3\textwidth}
         \centering
         \includegraphics[width=\textwidth]{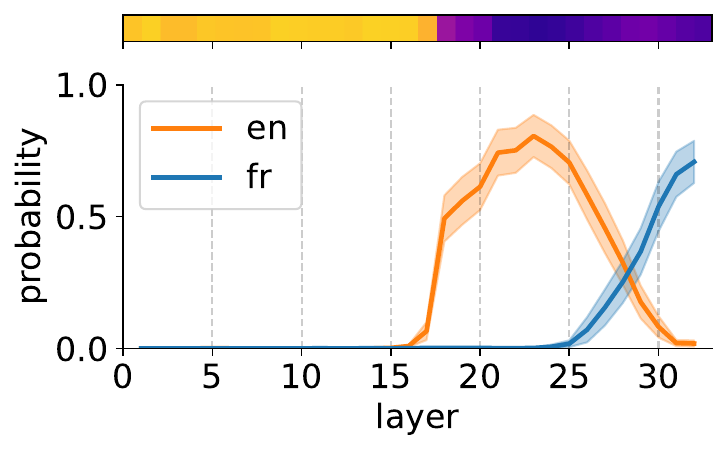}
     \end{subfigure}
     \hfill
     \begin{subfigure}[b]{0.28\textwidth}
         \centering
         \caption{Translation (DE->FR)}
         \includegraphics[width=\textwidth]{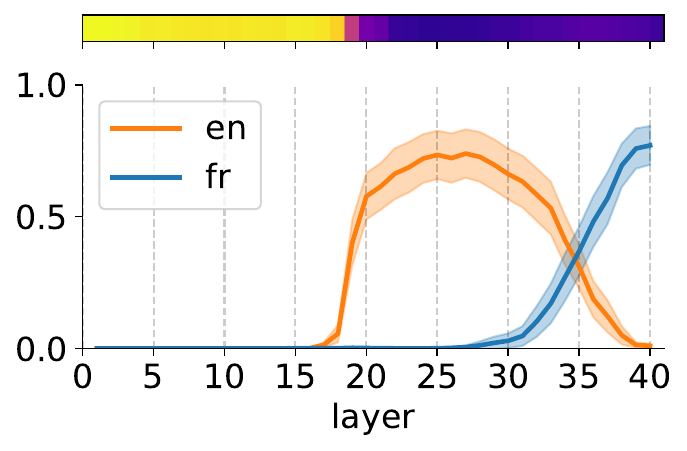}
     \end{subfigure}
     \hfill
     \begin{subfigure}[b]{0.343\textwidth}
         \centering
         \includegraphics[width=\textwidth]{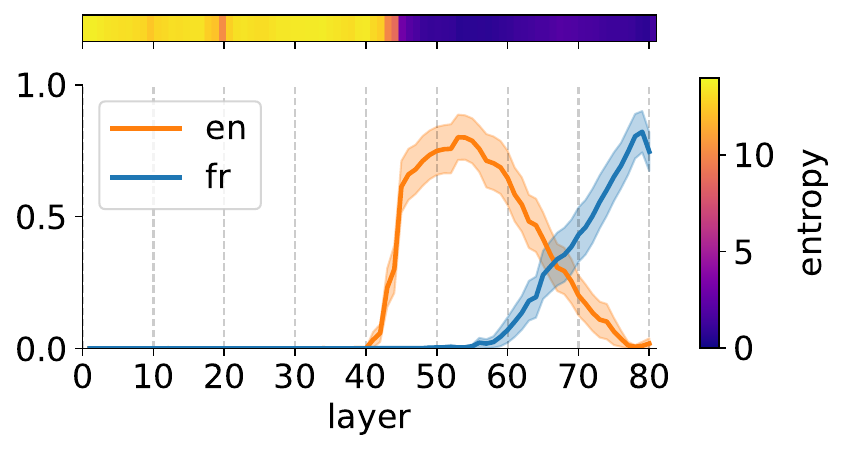}
     \end{subfigure}

     \begin{subfigure}[b]{0.3\textwidth}
         \centering
         \includegraphics[width=\textwidth]{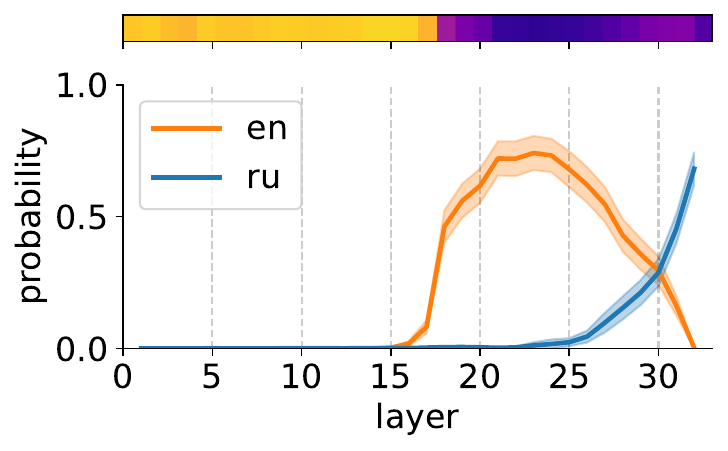}
     \end{subfigure}
     \hfill
     \begin{subfigure}[b]{0.28\textwidth}
         \centering
         \caption{Translation (DE->RU)}
         \includegraphics[width=\textwidth]{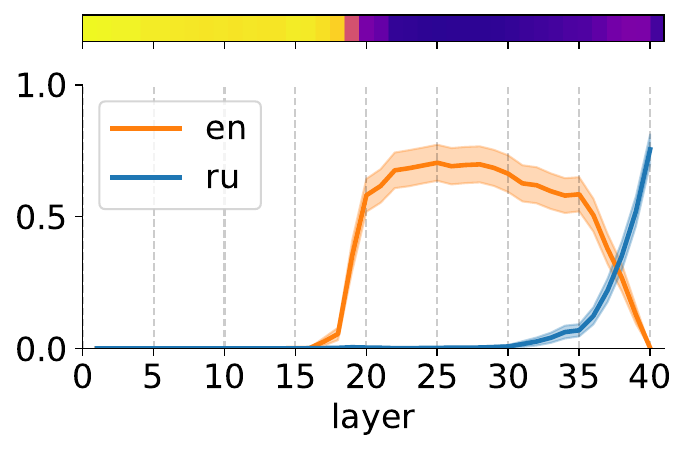}
     \end{subfigure}
     \hfill
     \begin{subfigure}[b]{0.343\textwidth}
         \centering
         \includegraphics[width=\textwidth]{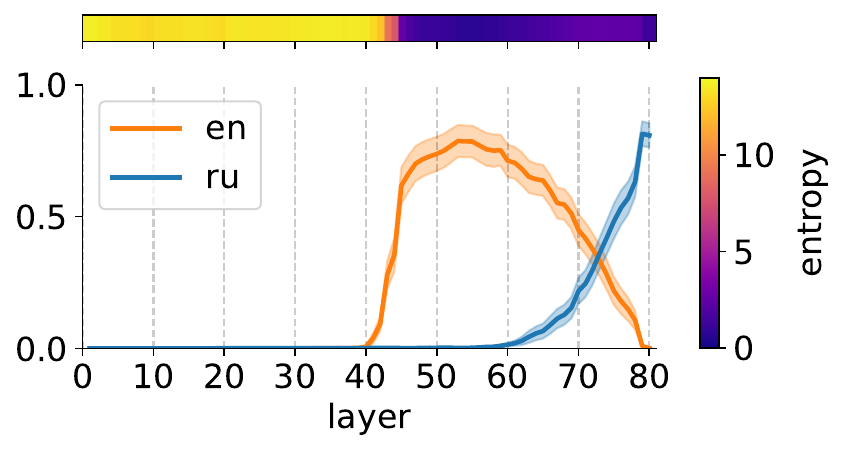}
     \end{subfigure}

     \begin{subfigure}[b]{0.3\textwidth}
         \centering
         \includegraphics[width=\textwidth]{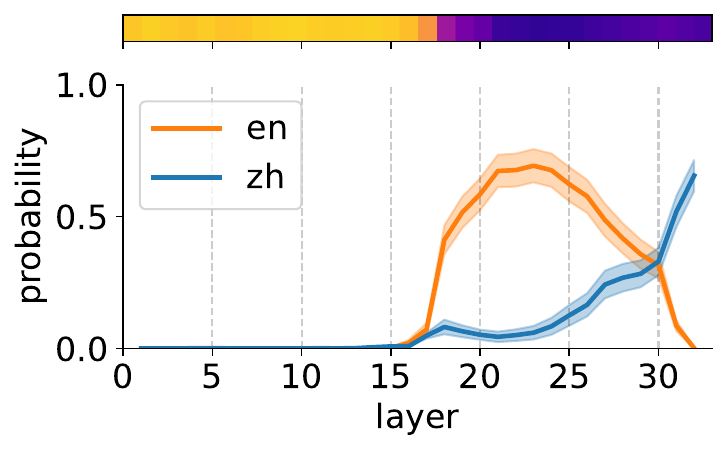}
     \end{subfigure}
     \hfill
     \begin{subfigure}[b]{0.28\textwidth}
         \centering
         \caption{Translation (DE->ZH)}
         \includegraphics[width=\textwidth]{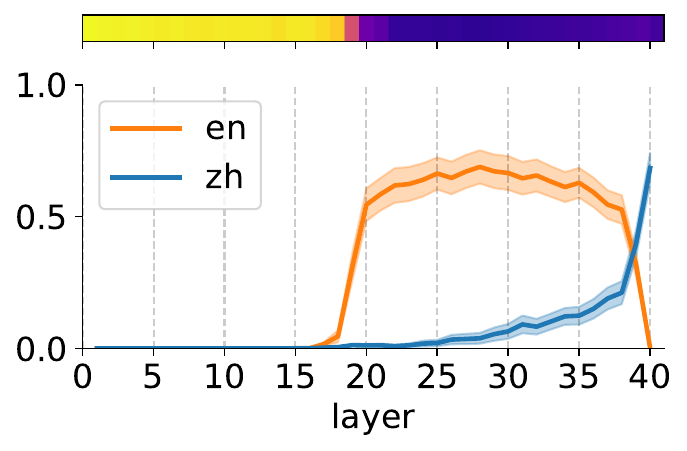}
     \end{subfigure}
     \hfill
     \begin{subfigure}[b]{0.343\textwidth}
         \centering
         \includegraphics[width=\textwidth]{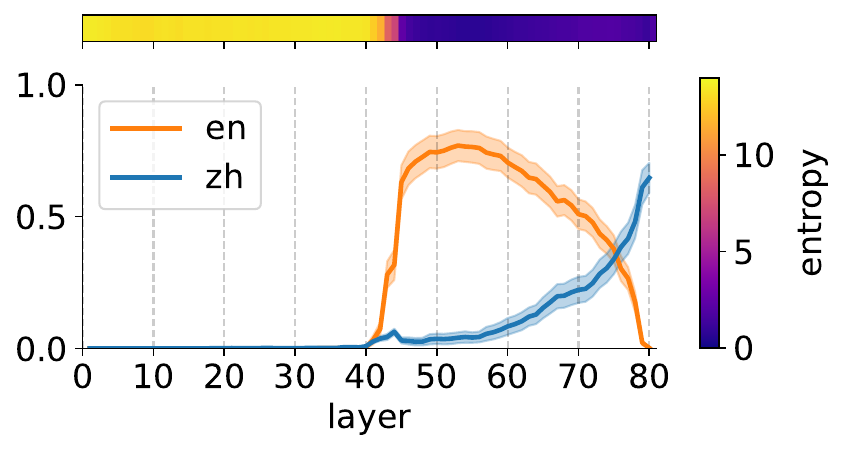}
     \end{subfigure}
     
     \caption{Figures illustrate the translation task where \llama{} 7B, 13B, and 70B are tasked with translating a word from non-English input language to output language. There is one column per model size. The x-axis shows the layer number of the model, and the y-axis the total probability mass falling on the correct token across languages. The orange line illustrates the probability of the correct target word in English and the blue line shows it for the non-English output language. We do not include the probability the input language since it is zero throughout. Means and 95\% Gaussian confidence intervals have been computed over the input examples, numbers in \Appref{app:info}.}
     \label{fig:translation-de}
 \end{figure*}

\begin{figure*}[ht!]
     \centering
     \begin{subfigure}[b]{0.3\textwidth}
         \centering
         \includegraphics[width=\textwidth]{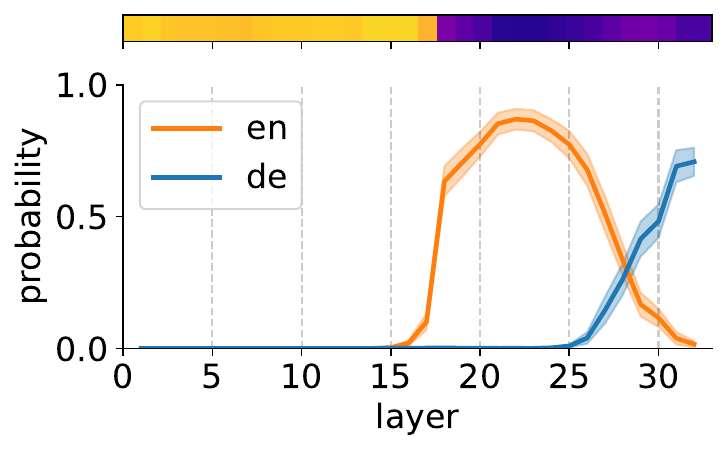}
     \end{subfigure}
     \hfill
     \begin{subfigure}[b]{0.28\textwidth}
         \centering
         \caption{Translation (EN->DE)}
         \includegraphics[width=\textwidth]{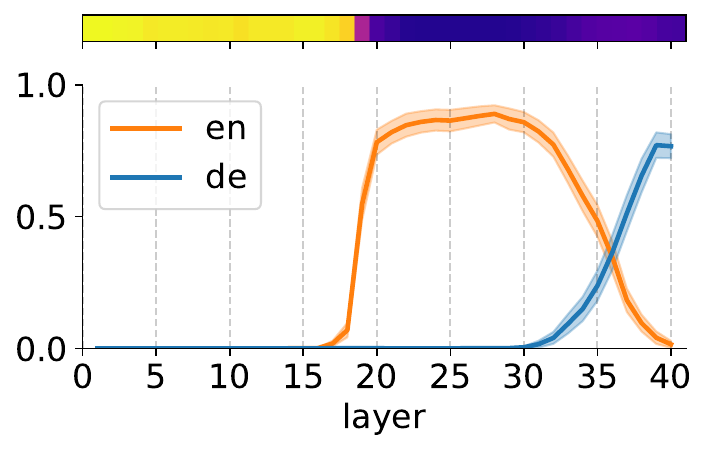}
     \end{subfigure}
     \hfill
     \begin{subfigure}[b]{0.343\textwidth}
         \centering
         \includegraphics[width=\textwidth]{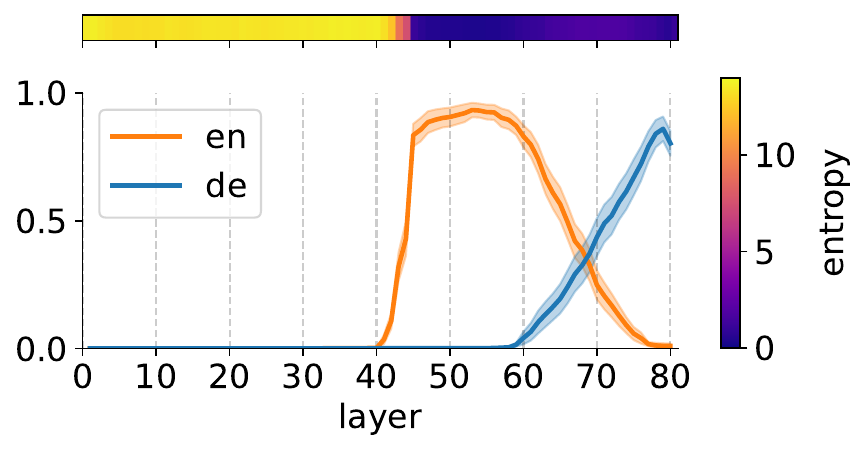}
     \end{subfigure}
     
     \begin{subfigure}[b]{0.3\textwidth}
         \centering
         \includegraphics[width=\textwidth]{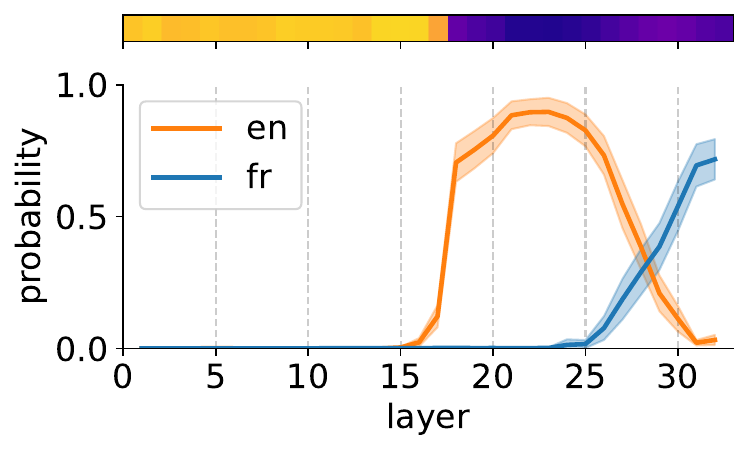}
     \end{subfigure}
     \hfill
     \begin{subfigure}[b]{0.28\textwidth}
         \centering
         \caption{Translation (EN->FR)}
         \includegraphics[width=\textwidth]{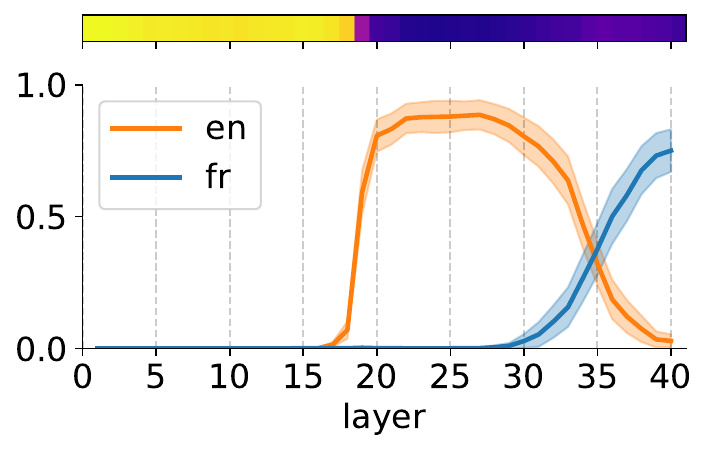}
     \end{subfigure}
     \hfill
     \begin{subfigure}[b]{0.343\textwidth}
         \centering
         \includegraphics[width=\textwidth]{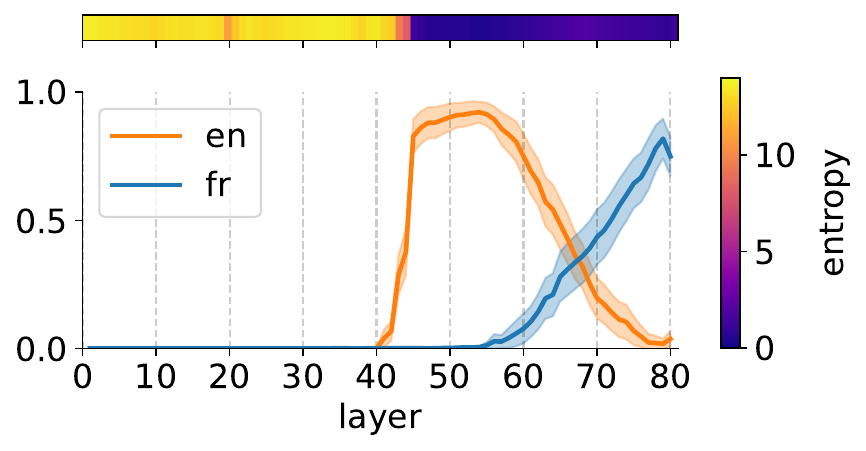}
     \end{subfigure}

     \begin{subfigure}[b]{0.3\textwidth}
         \centering
         \includegraphics[width=\textwidth]{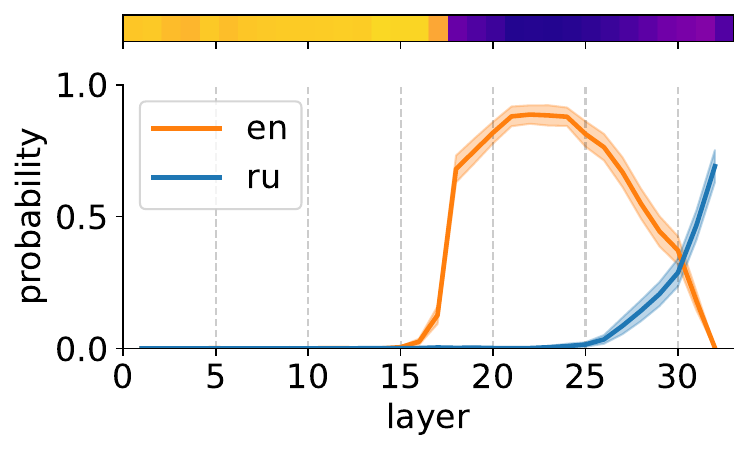}
     \end{subfigure}
     \hfill
     \begin{subfigure}[b]{0.28\textwidth}
         \centering
         \caption{Translation (EN->RU)}
         \includegraphics[width=\textwidth]{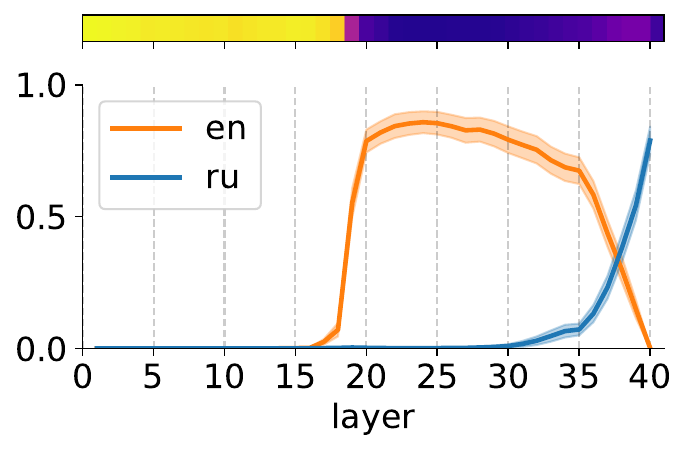}
     \end{subfigure}
     \hfill
     \begin{subfigure}[b]{0.343\textwidth}
         \centering
         \includegraphics[width=\textwidth]{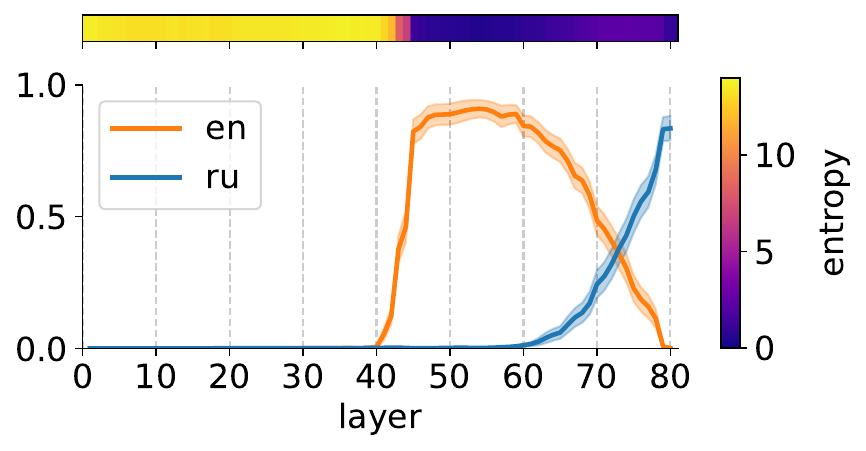}
     \end{subfigure}

     \begin{subfigure}[b]{0.3\textwidth}
         \centering
         \includegraphics[width=\textwidth]{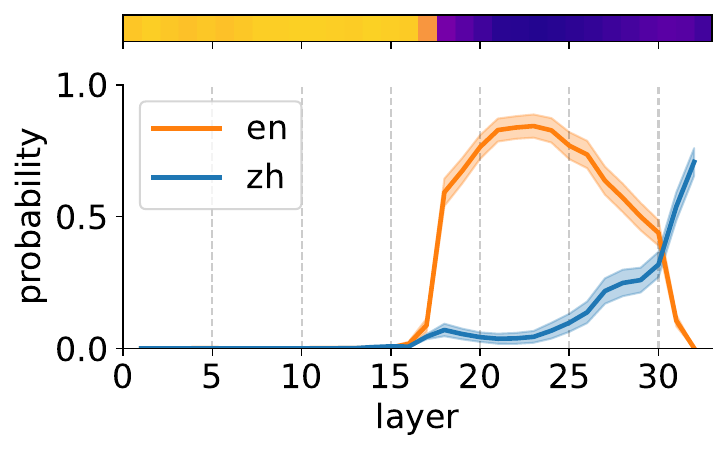}
     \end{subfigure}
     \hfill
     \begin{subfigure}[b]{0.28\textwidth}
         \centering
         \caption{Translation (EN->ZH)}
         \includegraphics[width=\textwidth]{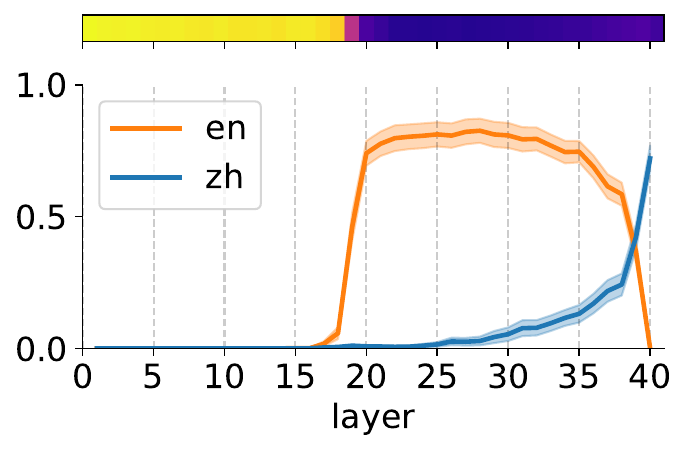}
     \end{subfigure}
     \hfill
     \begin{subfigure}[b]{0.343\textwidth}
         \centering
         \includegraphics[width=\textwidth]{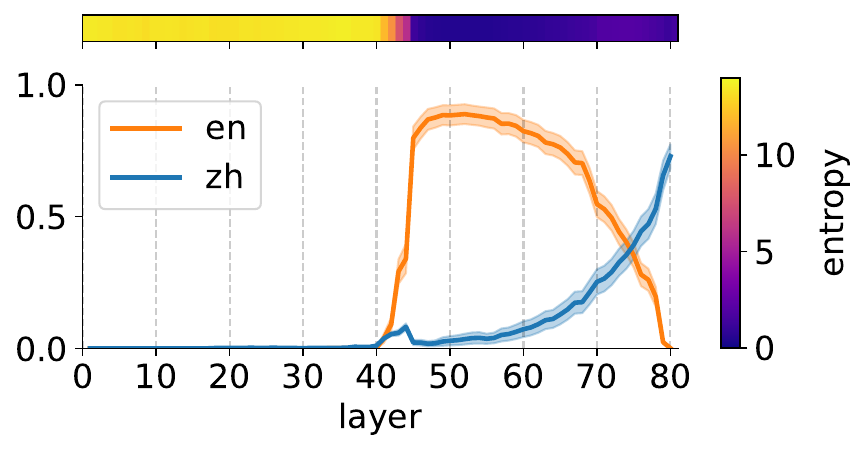}
     \end{subfigure}
     
     \caption{Figures illustrate the translation task where \llama{} 7B, 13B, and 70B are tasked with translating a word from English input language to output language. There is one column per model size. The x-axis shows the layer number of the model, and the y-axis the total probability mass falling on the correct token across languages. The orange line illustrates the probability of the correct target word in English and the blue line shows it for the non-English output language. We do not include the probability the input language since it is zero throughout. Means and 95\% Gaussian confidence intervals have been computed over the input examples, numbers in \Appref{app:info}.}
     \label{fig:translation-en}
 \end{figure*}

\begin{figure*}[ht!]
     \centering
     \begin{subfigure}[b]{0.3\textwidth}
         \centering
         \includegraphics[width=\textwidth]{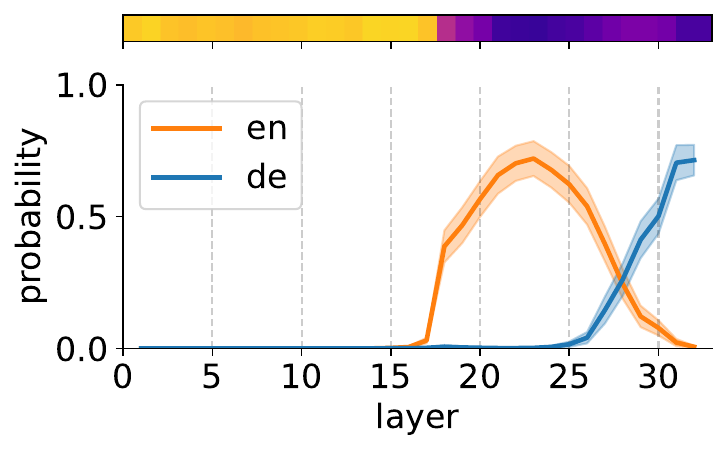}
     \end{subfigure}
     \hfill
     \begin{subfigure}[b]{0.28\textwidth}
         \centering
         \caption{Translation (FR->DE)}
         \includegraphics[width=\textwidth]{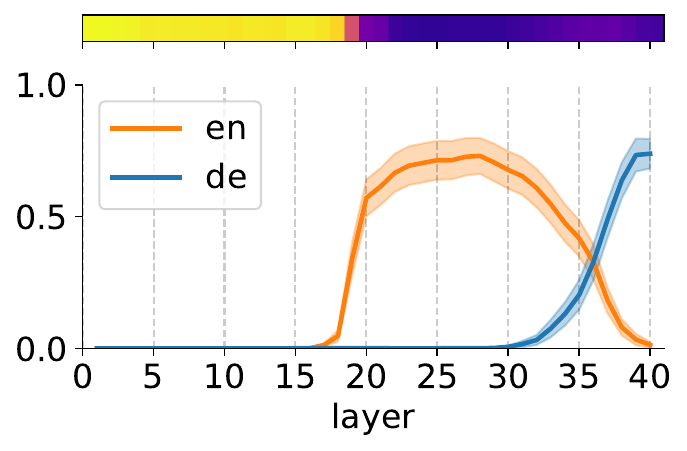}
     \end{subfigure}
     \hfill
     \begin{subfigure}[b]{0.343\textwidth}
         \centering
         \includegraphics[width=\textwidth]{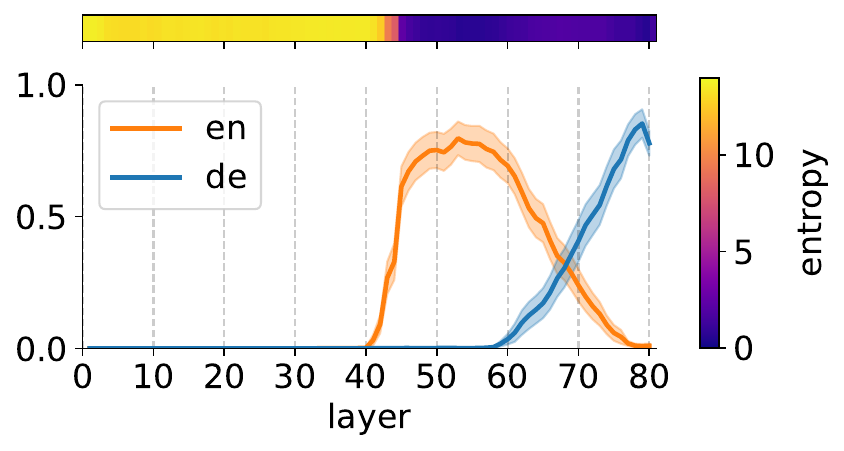}
     \end{subfigure}
     
     \begin{subfigure}[b]{0.3\textwidth}
         \centering
         \includegraphics[width=\textwidth]{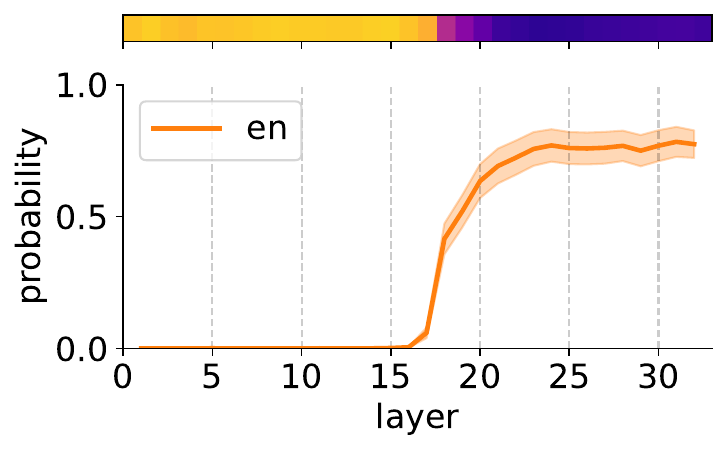}
     \end{subfigure}
     \hfill
     \begin{subfigure}[b]{0.28\textwidth}
         \centering
         \caption{Translation (FR->EN)}
         \includegraphics[width=\textwidth]{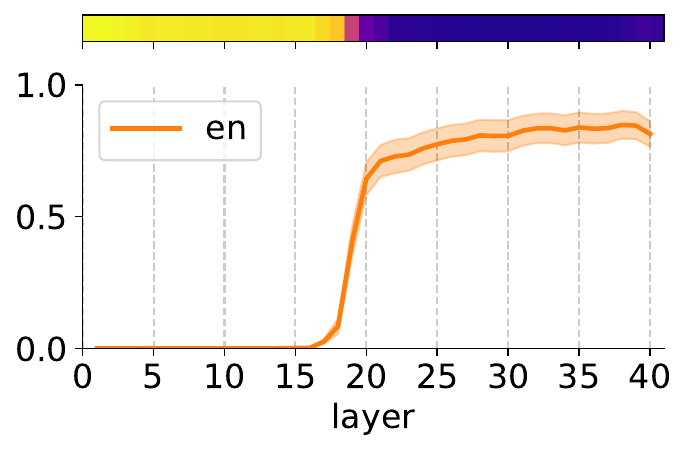}
     \end{subfigure}
     \hfill
     \begin{subfigure}[b]{0.343\textwidth}
         \centering
         \includegraphics[width=\textwidth]{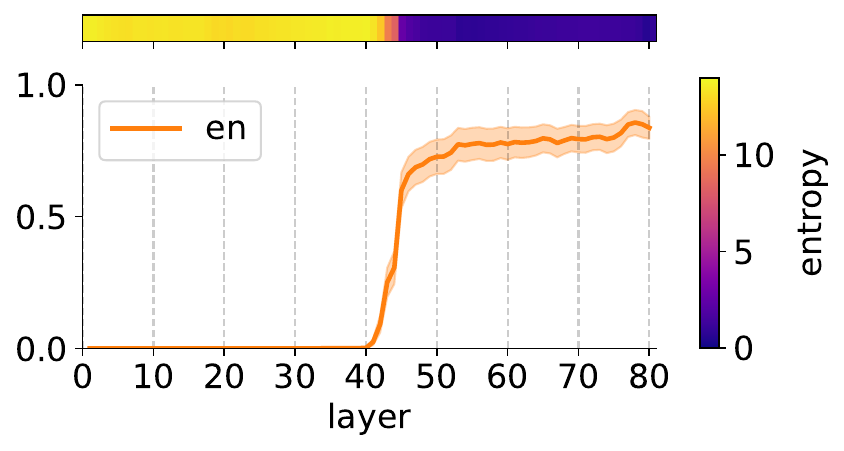}
     \end{subfigure}

     \begin{subfigure}[b]{0.3\textwidth}
         \centering
         \includegraphics[width=\textwidth]{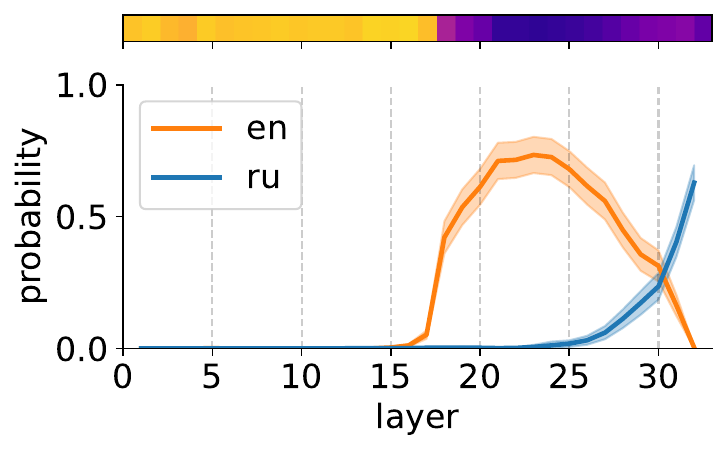}
     \end{subfigure}
     \hfill
     \begin{subfigure}[b]{0.28\textwidth}
         \centering
         \caption{Translation (FR->RU)}
         \includegraphics[width=\textwidth]{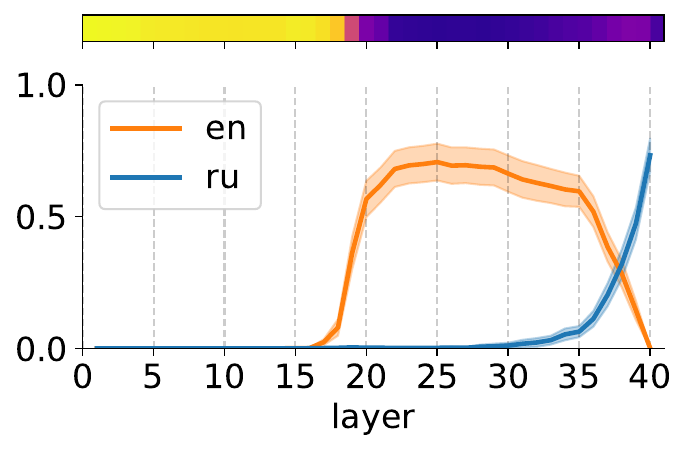}
     \end{subfigure}
     \hfill
     \begin{subfigure}[b]{0.343\textwidth}
         \centering
         \includegraphics[width=\textwidth]{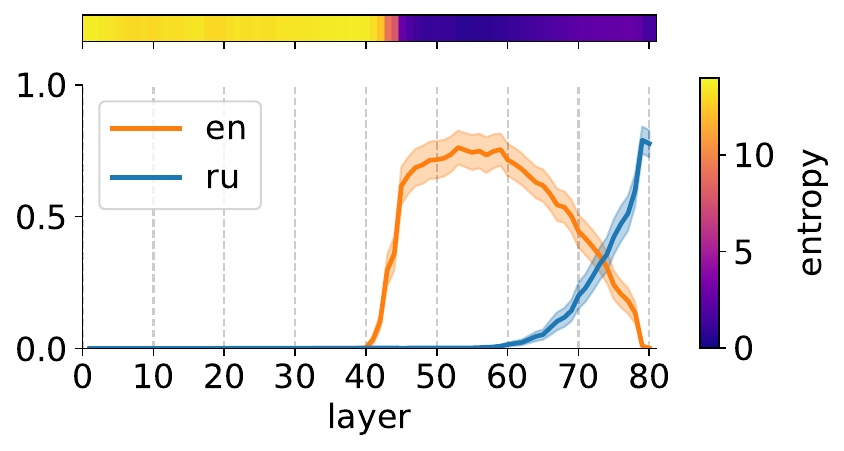}
     \end{subfigure}

     \begin{subfigure}[b]{0.3\textwidth}
         \centering
         \includegraphics[width=\textwidth]{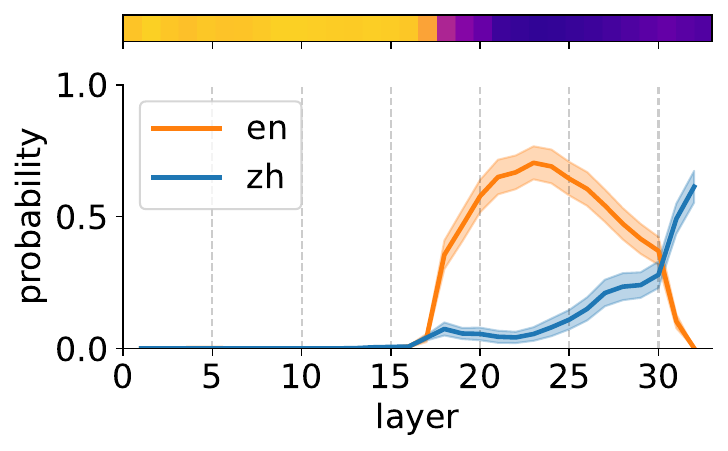}
     \end{subfigure}
     \hfill
     \begin{subfigure}[b]{0.28\textwidth}
         \centering
         \caption{Translation (FR->ZH)}
         \includegraphics[width=\textwidth]{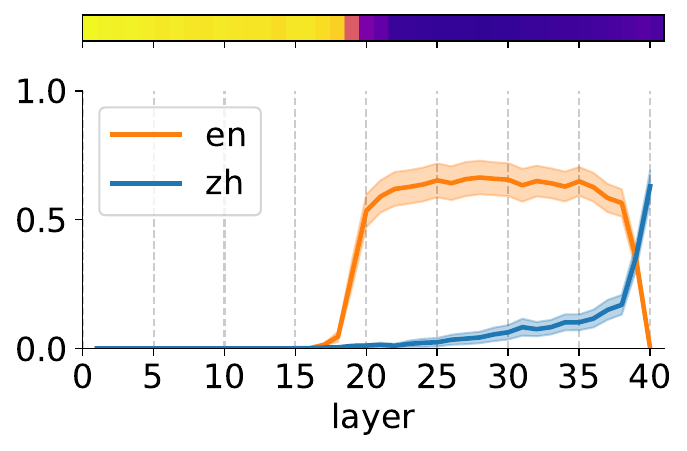}
     \end{subfigure}
     \hfill
     \begin{subfigure}[b]{0.343\textwidth}
         \centering
         \includegraphics[width=\textwidth]{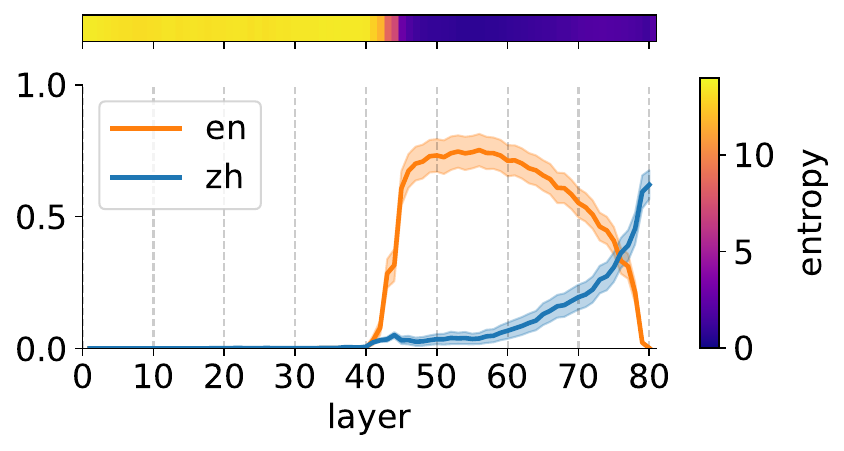}
     \end{subfigure}
     
     \caption{Figures illustrate the translation task where \llama{} 7B, 13B, and 70B are tasked with translating a word from non-English input language to output language. There is one column per model size. The x-axis shows the layer number of the model, and the y-axis the total probability mass falling on the correct token across languages. The orange line illustrates the probability of the correct target word in English and the blue line shows it for the non-English output language. We do not include the probability the input language since it is zero throughout. Means and 95\% Gaussian confidence intervals have been computed over the input examples, numbers in \Appref{app:info}.}
     \label{fig:translation-fr}
 \end{figure*}

\begin{figure*}[ht!]
     \centering
     \begin{subfigure}[b]{0.3\textwidth}
         \centering
         \includegraphics[width=\textwidth]{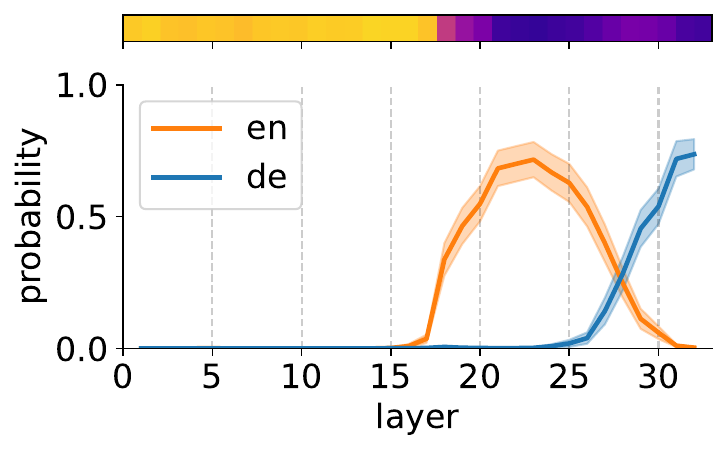}
     \end{subfigure}
     \hfill
     \begin{subfigure}[b]{0.28\textwidth}
         \centering
         \caption{Translation (RU->DE)}
         \includegraphics[width=\textwidth]{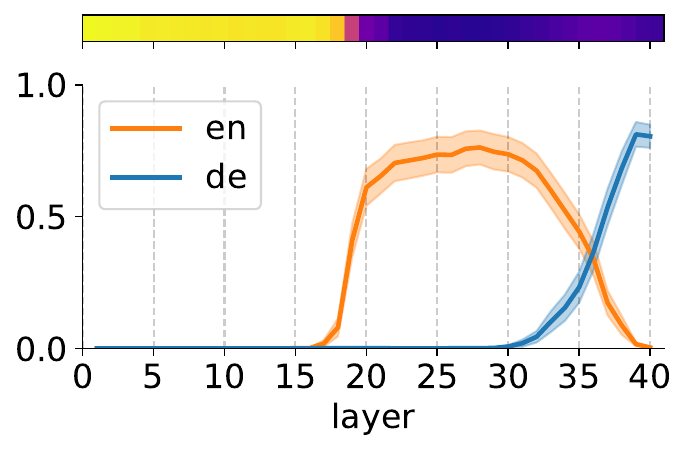}
     \end{subfigure}
     \hfill
     \begin{subfigure}[b]{0.343\textwidth}
         \centering
         \includegraphics[width=\textwidth]{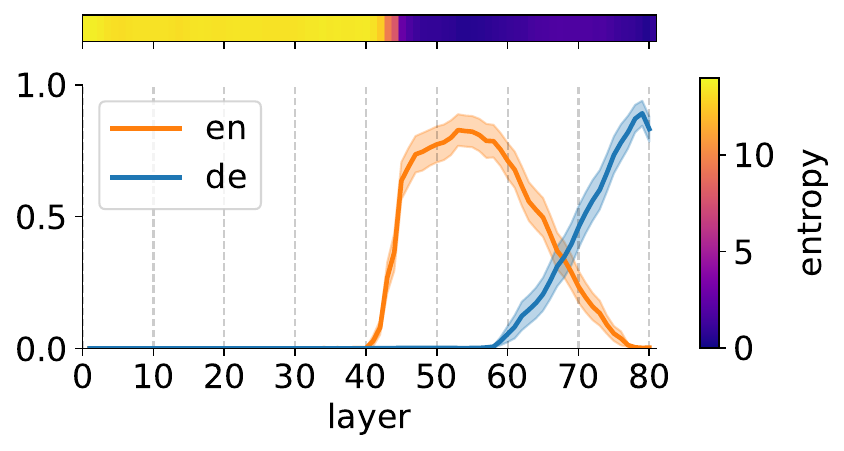}
     \end{subfigure}
     
     \begin{subfigure}[b]{0.3\textwidth}
         \centering
         \includegraphics[width=\textwidth]{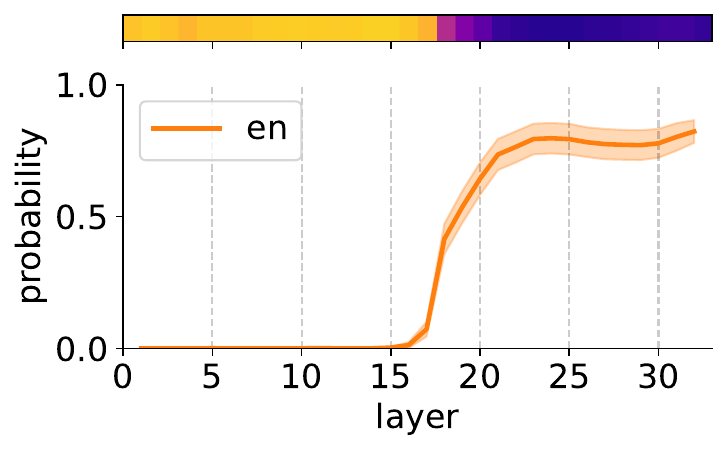}
     \end{subfigure}
     \hfill
     \begin{subfigure}[b]{0.28\textwidth}
         \centering
         \caption{Translation (RU->EN)}
         \includegraphics[width=\textwidth]{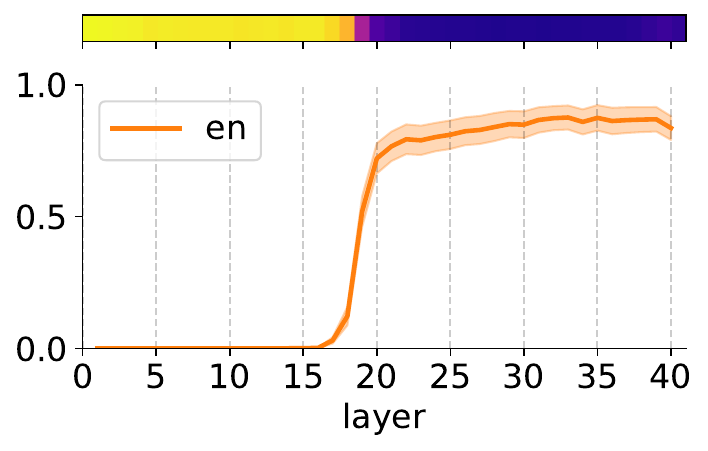}
     \end{subfigure}
     \hfill
     \begin{subfigure}[b]{0.343\textwidth}
         \centering
         \includegraphics[width=\textwidth]{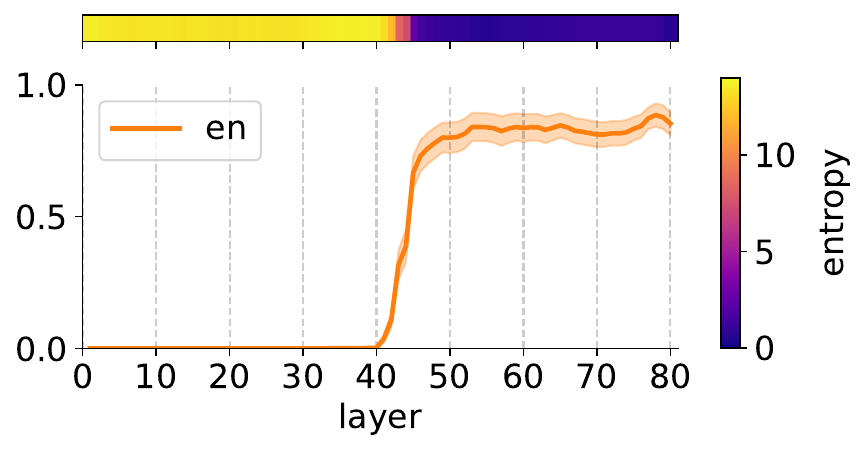}
     \end{subfigure}

     \begin{subfigure}[b]{0.3\textwidth}
         \centering
         \includegraphics[width=\textwidth]{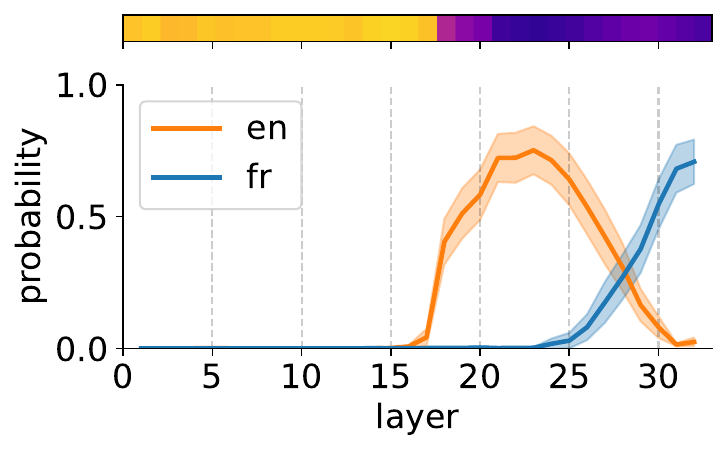}
     \end{subfigure}
     \hfill
     \begin{subfigure}[b]{0.28\textwidth}
         \centering
         \caption{Translation (RU->FR)}
         \includegraphics[width=\textwidth]{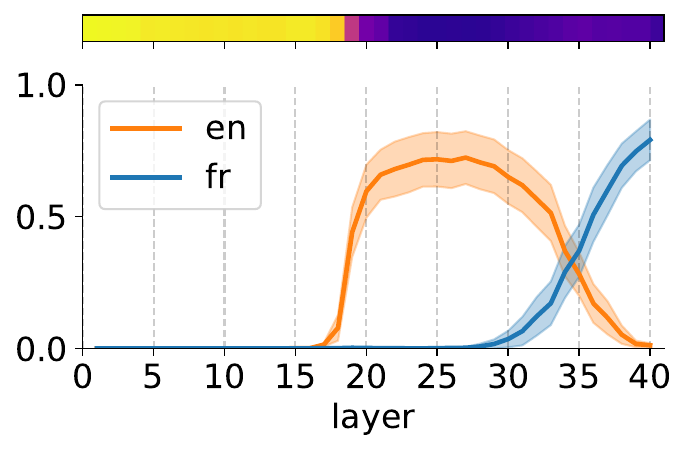}
     \end{subfigure}
     \hfill
     \begin{subfigure}[b]{0.343\textwidth}
         \centering
         \includegraphics[width=\textwidth]{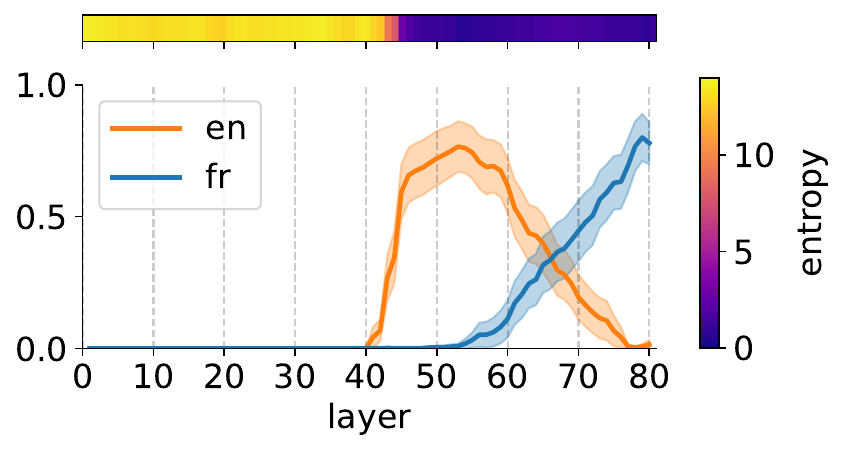}
     \end{subfigure}

     \begin{subfigure}[b]{0.3\textwidth}
         \centering
         \includegraphics[width=\textwidth]{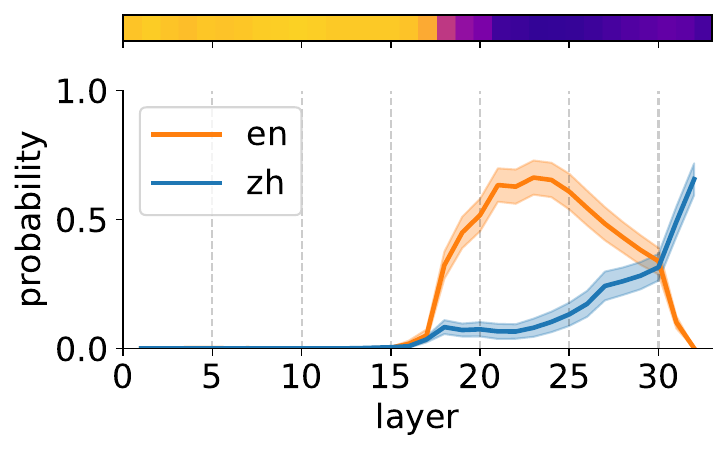}
     \end{subfigure}
     \hfill
     \begin{subfigure}[b]{0.28\textwidth}
         \centering
         \caption{Translation (RU->ZH)}
         \includegraphics[width=\textwidth]{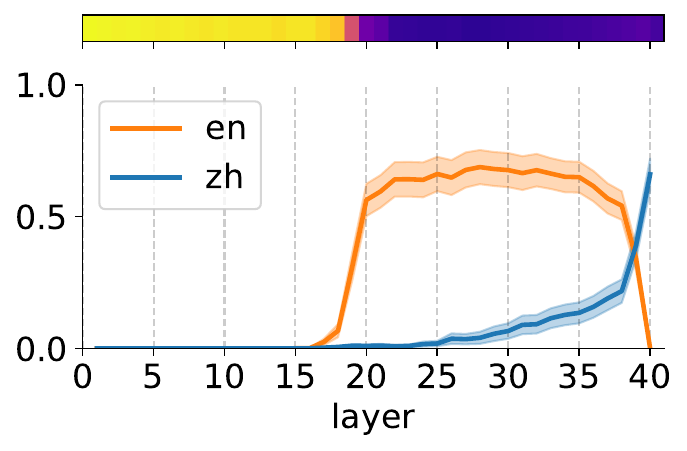}
     \end{subfigure}
     \hfill
     \begin{subfigure}[b]{0.343\textwidth}
         \centering
         \includegraphics[width=\textwidth]{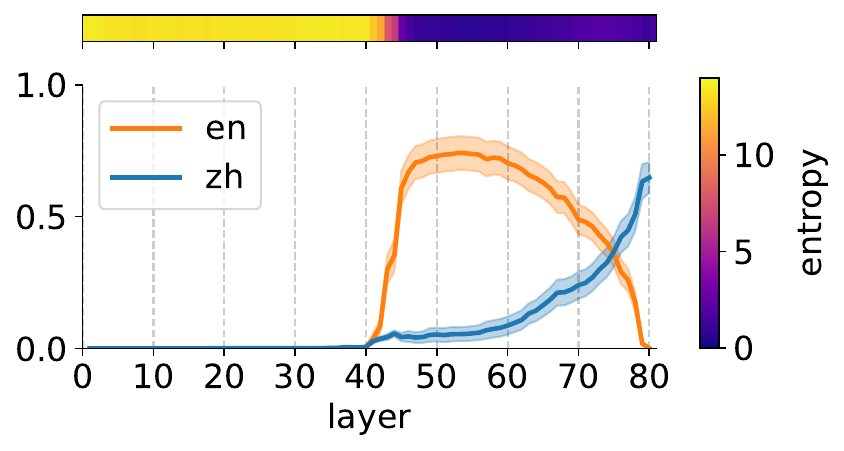}
     \end{subfigure}
     
     \caption{Figures illustrate the translation task where \llama{} 7B, 13B, and 70B are tasked with translating a word from non-English input language to output language. There is one column per model size. The x-axis shows the layer number of the model, and the y-axis the total probability mass falling on the correct token across languages. The orange line illustrates the probability of the correct target word in English and the blue line shows it for the non-English output language. We do not include the probability the input language since it is zero throughout. Means and 95\% Gaussian confidence intervals have been computed over the input examples, numbers in \Appref{app:info}.}
     \label{fig:translation-ru}
 \end{figure*}

\begin{figure*}[ht!]
     \centering
     \begin{subfigure}[b]{0.3\textwidth}
         \centering
         \includegraphics[width=\textwidth]{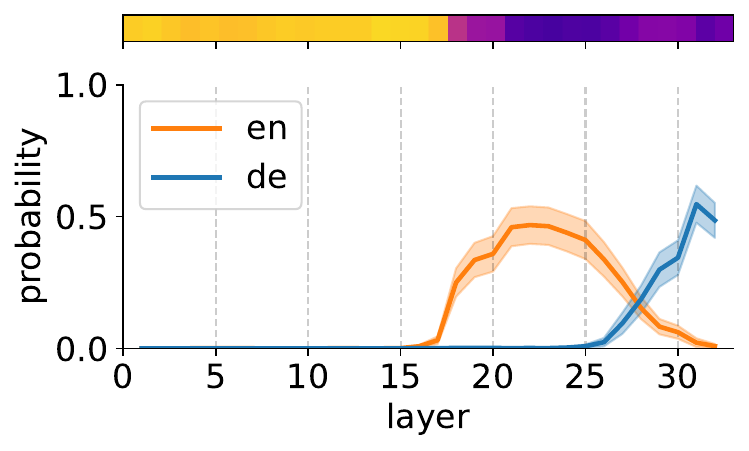}
     \end{subfigure}
     \hfill
     \begin{subfigure}[b]{0.28\textwidth}
         \centering
         \caption{Translation (ZH->DE)}
         \includegraphics[width=\textwidth]{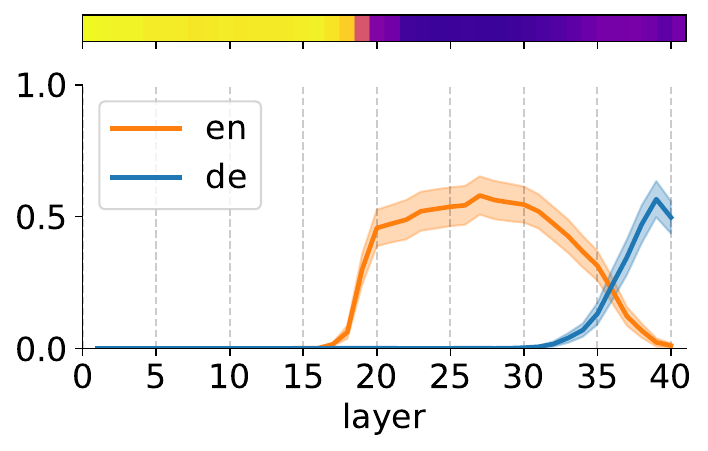}
     \end{subfigure}
     \hfill
     \begin{subfigure}[b]{0.343\textwidth}
         \centering
         \includegraphics[width=\textwidth]{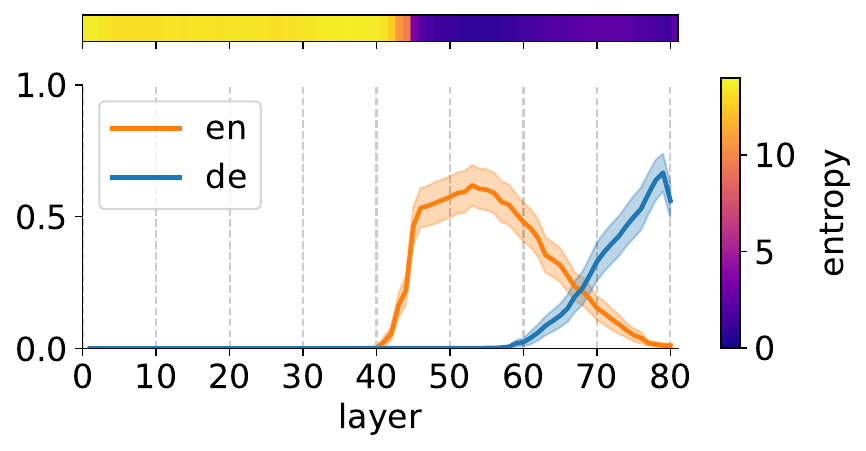}
     \end{subfigure}
     
     \begin{subfigure}[b]{0.3\textwidth}
         \centering
         \includegraphics[width=\textwidth]{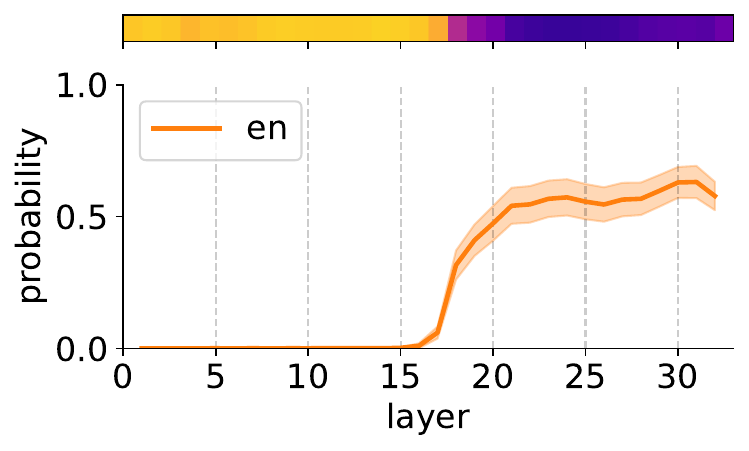}
     \end{subfigure}
     \hfill
     \begin{subfigure}[b]{0.28\textwidth}
         \centering
         \caption{Translation (ZH->EN)}
         \includegraphics[width=\textwidth]{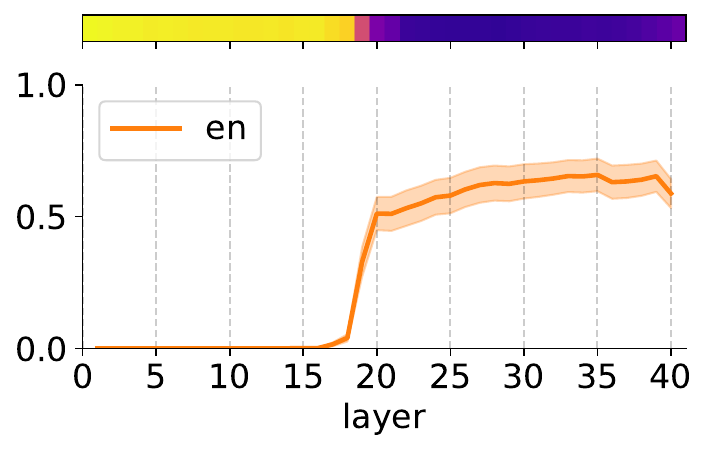}
     \end{subfigure}
     \hfill
     \begin{subfigure}[b]{0.343\textwidth}
         \centering
         \includegraphics[width=\textwidth]{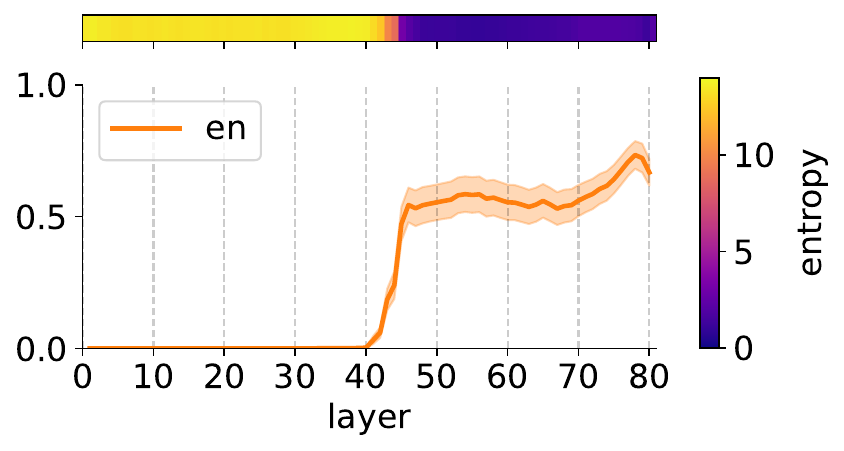}
     \end{subfigure}

     \begin{subfigure}[b]{0.3\textwidth}
         \centering
         \includegraphics[width=\textwidth]{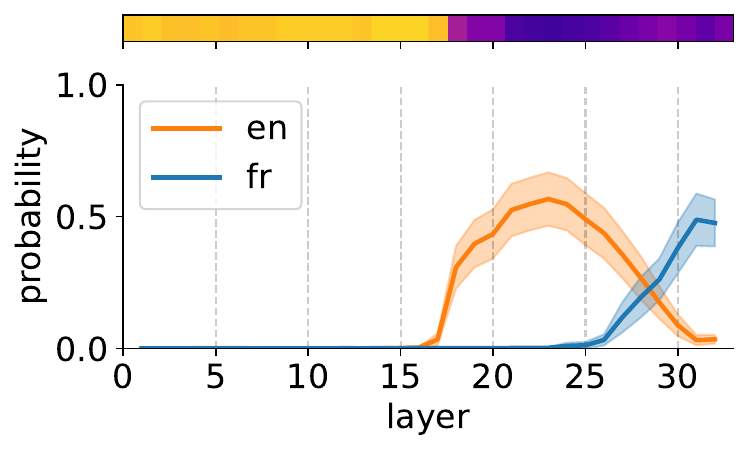}
     \end{subfigure}
     \hfill
     \begin{subfigure}[b]{0.28\textwidth}
         \centering
         \caption{Translation (ZH->FR)}
         \includegraphics[width=\textwidth]{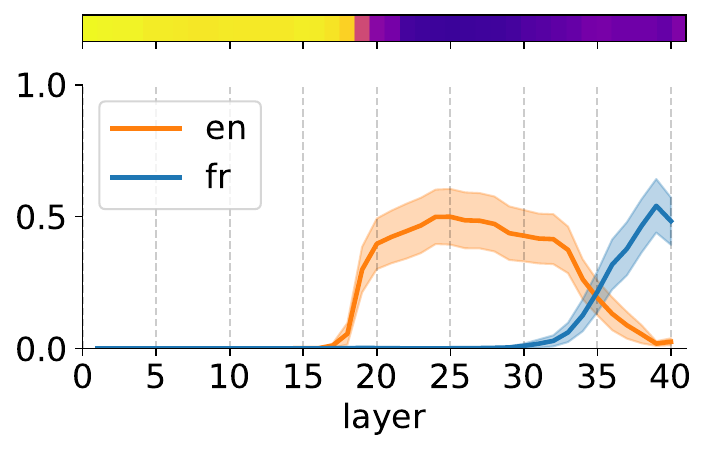}
     \end{subfigure}
     \hfill
     \begin{subfigure}[b]{0.343\textwidth}
         \centering
         \includegraphics[width=\textwidth]{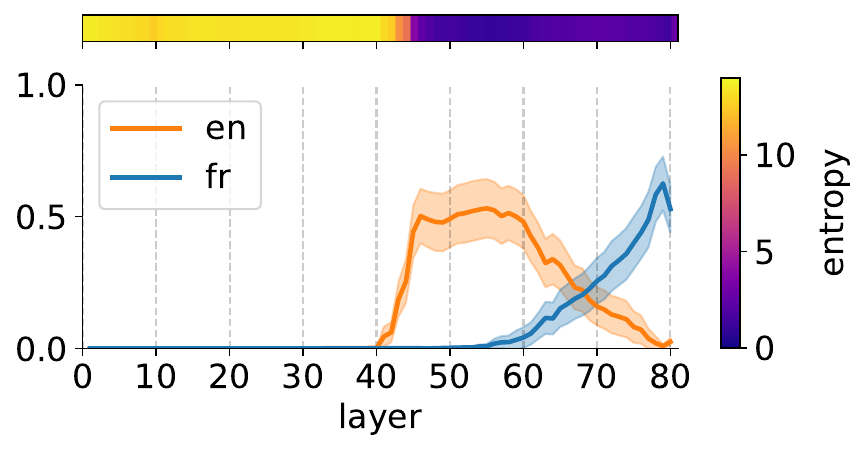}
     \end{subfigure}

     \begin{subfigure}[b]{0.3\textwidth}
         \centering
         \includegraphics[width=\textwidth]{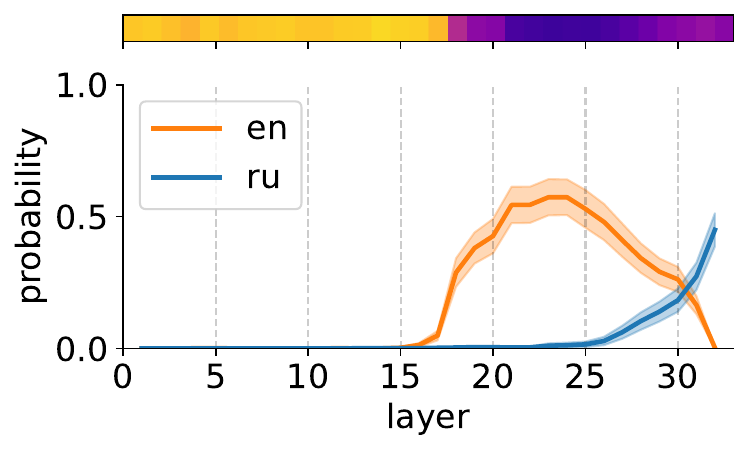}
     \end{subfigure}
     \hfill
     \begin{subfigure}[b]{0.28\textwidth}
         \centering
         \caption{Translation (ZH->RU)}
         \includegraphics[width=\textwidth]{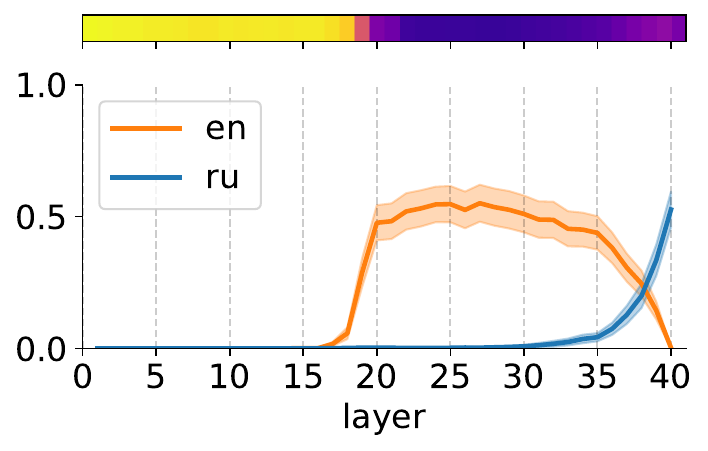}
     \end{subfigure}
     \hfill
     \begin{subfigure}[b]{0.343\textwidth}
         \centering
         \includegraphics[width=\textwidth]{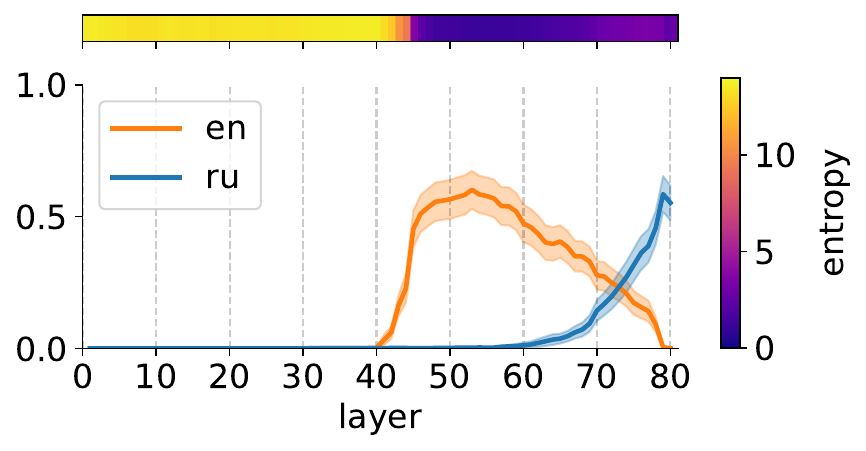}
     \end{subfigure}
     
     \caption{Figures illustrate the translation task where \llama{} 7B, 13B, and 70B are tasked with translating a word from non-English input language to output language. There is one column per model size. The x-axis shows the layer number of the model, and the y-axis the total probability mass falling on the correct token across languages. The orange line illustrates the probability of the correct target word in English and the blue line shows it for the non-English output language. We do not include the probability the input language since it is zero throughout. Means and 95\% Gaussian confidence intervals have been computed over the input examples, numbers in \Appref{app:info}.}
     \label{fig:translation-zh}
 \end{figure*}

\xhdr{Language probability} Language probability plots (with entropy heatmaps) for the aggregated translation task are in Fig.~\ref{fig:translation}, for the repetition task in Fig.~\ref{fig:copy}, and, for the cloze task in Fig.~\ref{fig:cloze}. Additionally, we provide the translation task results for individual language pairs in Fig.~\ref{fig:translation-de}, Fig.~\ref{fig:translation-en}, Fig.~\ref{fig:translation-fr}, Fig.~\ref{fig:translation-ru}, Fig.~\ref{fig:translation-zh}.

We observe the same pattern---noise in the early layers, English in the middle, target language in the end---across almost all languages and model sizes. The only exception is the Chinese repetition task.

\begin{figure*}[ht!]
     \centering
     \begin{subfigure}[b]{0.3\textwidth}
         \centering
         \includegraphics[width=\textwidth]{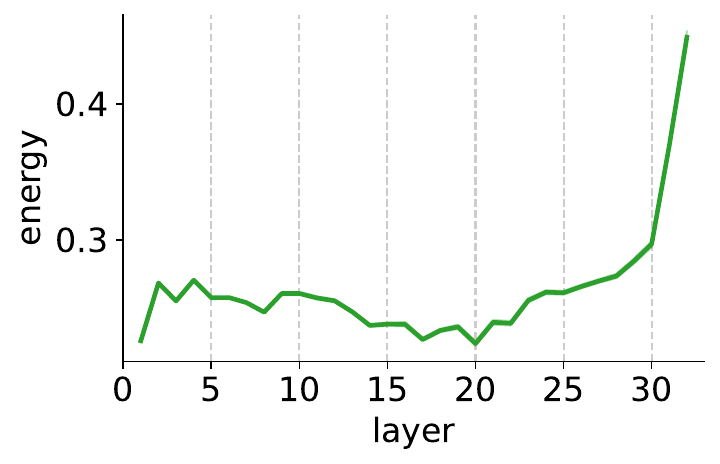}
     \end{subfigure}
     \hfill
     \begin{subfigure}[b]{0.28\textwidth}
         \centering
         \caption{Translation (-> DE)}
         \includegraphics[width=\textwidth]{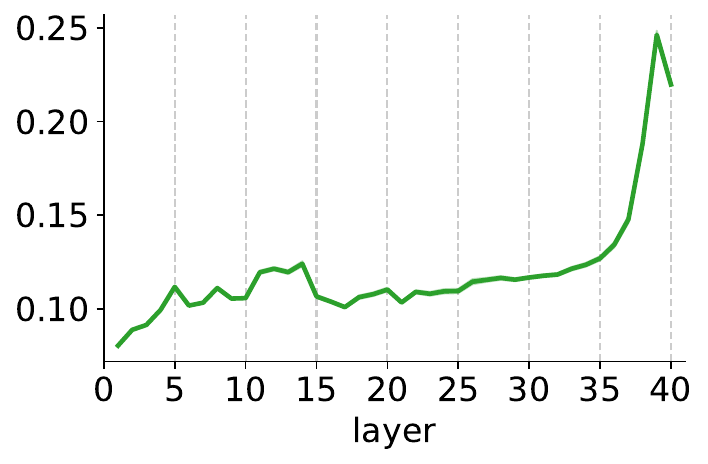}
     \end{subfigure}
     \hfill
     \begin{subfigure}[b]{0.28\textwidth}
         \centering
         \includegraphics[width=\textwidth]{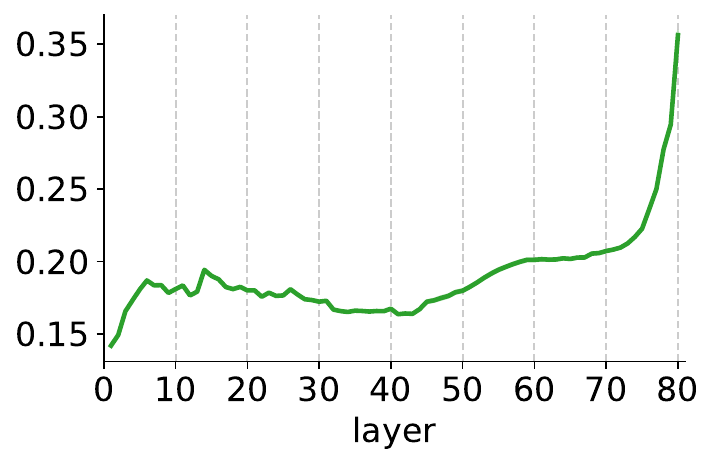}
     \end{subfigure}

     \begin{subfigure}[b]{0.3\textwidth}
         \centering
         \includegraphics[width=\textwidth]{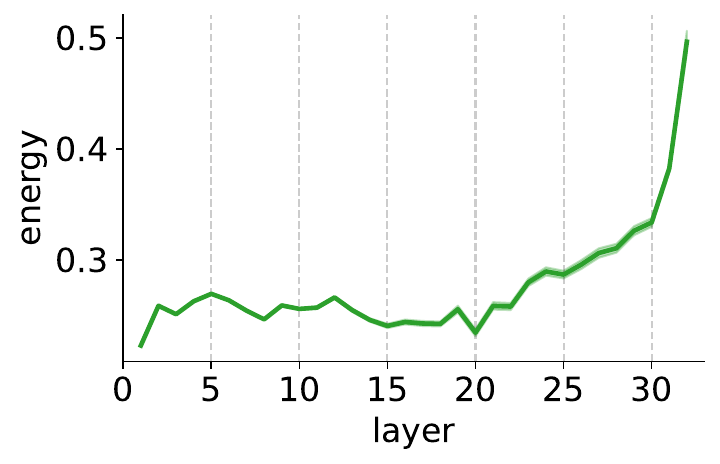}
     \end{subfigure}
     \hfill
     \begin{subfigure}[b]{0.28\textwidth}
         \centering
         \caption{Translation (-> FR)}
         \includegraphics[width=\textwidth]{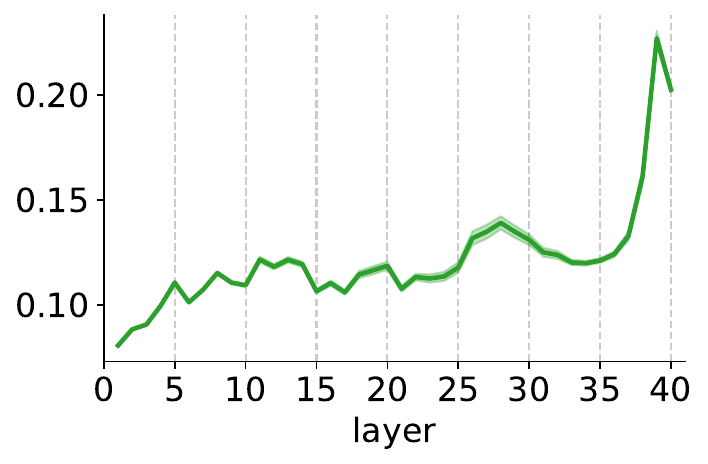}
     \end{subfigure}
     \hfill
     \begin{subfigure}[b]{0.28\textwidth}
         \centering
         \includegraphics[width=\textwidth]{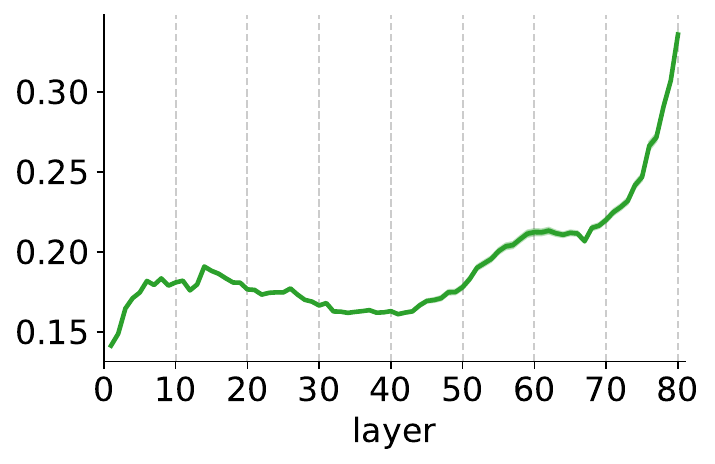}
     \end{subfigure}

     \begin{subfigure}[b]{0.3\textwidth}
         \centering
         \includegraphics[width=\textwidth]{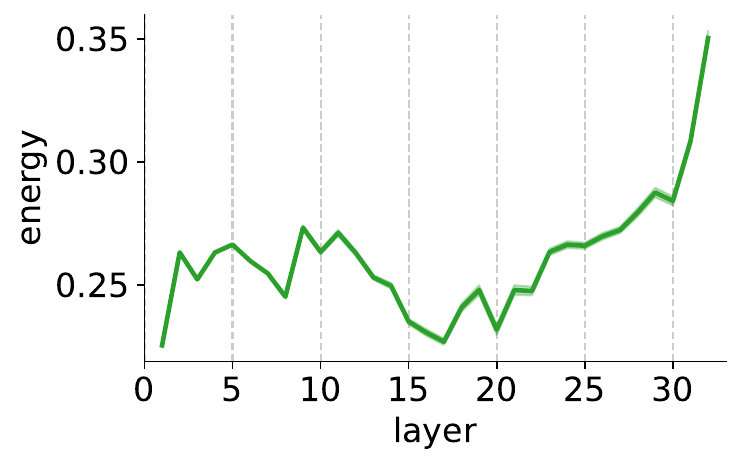}
     \end{subfigure}
     \hfill
     \begin{subfigure}[b]{0.28\textwidth}
         \centering
         \caption{Translation (-> RU)}
         \includegraphics[width=\textwidth]{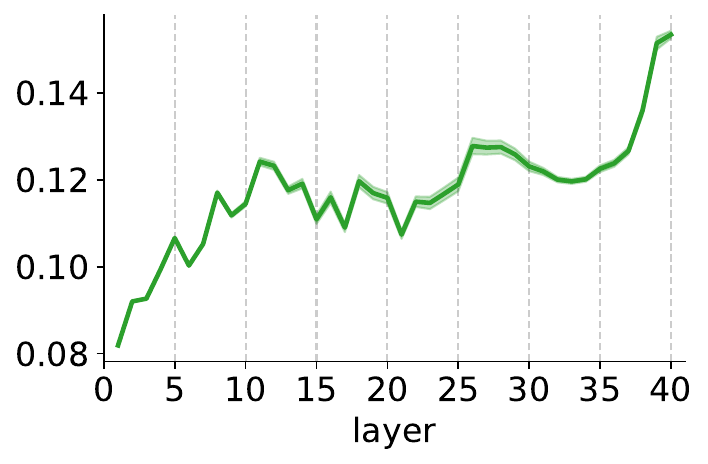}
     \end{subfigure}
     \hfill
     \begin{subfigure}[b]{0.28\textwidth}
         \centering
         \includegraphics[width=\textwidth]{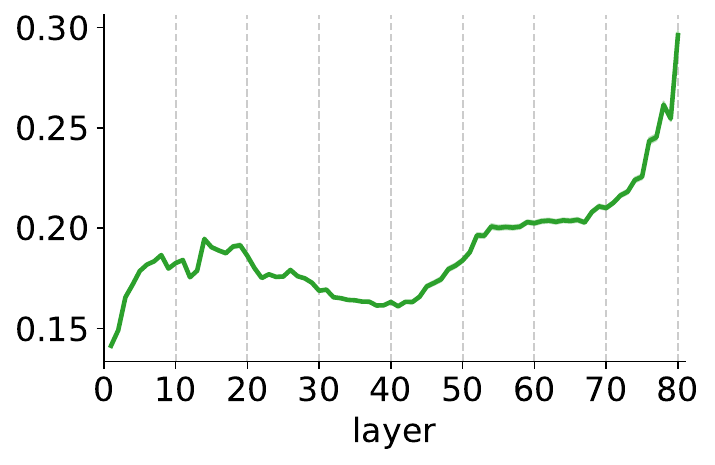}
     \end{subfigure}

     \begin{subfigure}[b]{0.3\textwidth}
         \centering
         \includegraphics[width=\textwidth]{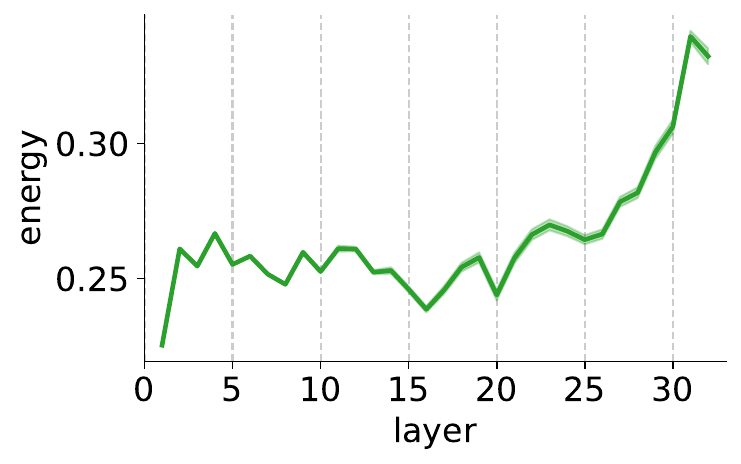}
     \end{subfigure}
     \hfill
     \begin{subfigure}[b]{0.28\textwidth}
         \centering
         \caption{Translation (-> ZH)}
         \includegraphics[width=\textwidth]{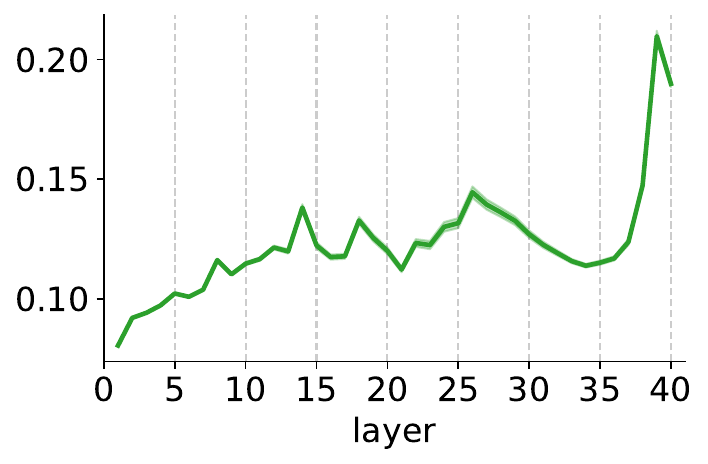}
     \end{subfigure}
     \hfill
     \begin{subfigure}[b]{0.28\textwidth}
         \centering
         \includegraphics[width=\textwidth]{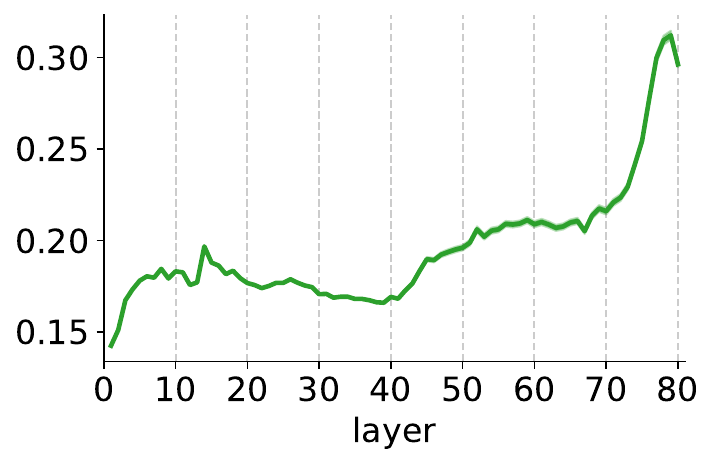}
     \end{subfigure}
\caption{Figures illustrate the translation task where \llama{} 7B, 13B, and 70B are tasked with translating a word from all non-English input languages to output language. There is one column per model size. The x-axis shows the layer number of the model, and the y-axis the energy. Means and 95\% Gaussian confidence intervals have been computed over the input examples, numbers in \Appref{app:info}.}
     \label{fig:translation-energy}
 \end{figure*}

\begin{figure*}[ht!]
     \centering
     \begin{subfigure}[b]{0.3\textwidth}
         \centering
         \includegraphics[width=\textwidth]{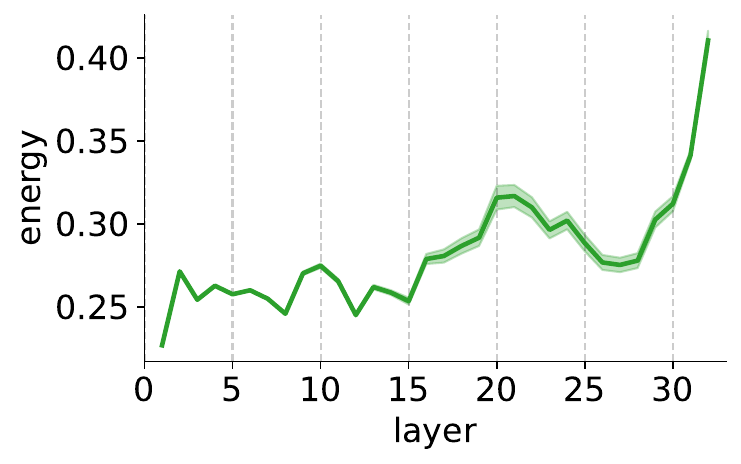}
     \end{subfigure}
     \hfill
     \begin{subfigure}[b]{0.28\textwidth}
         \centering
         \caption{Repetition (DE)}
         \includegraphics[width=\textwidth]{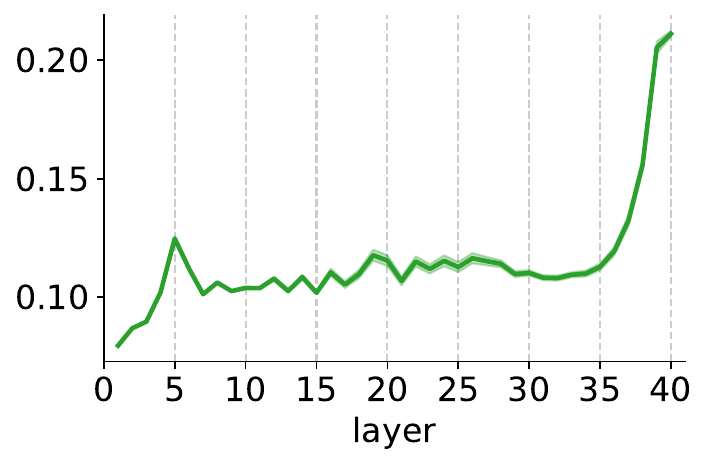}
     \end{subfigure}
     \hfill
     \begin{subfigure}[b]{0.28\textwidth}
         \centering
         \includegraphics[width=\textwidth]{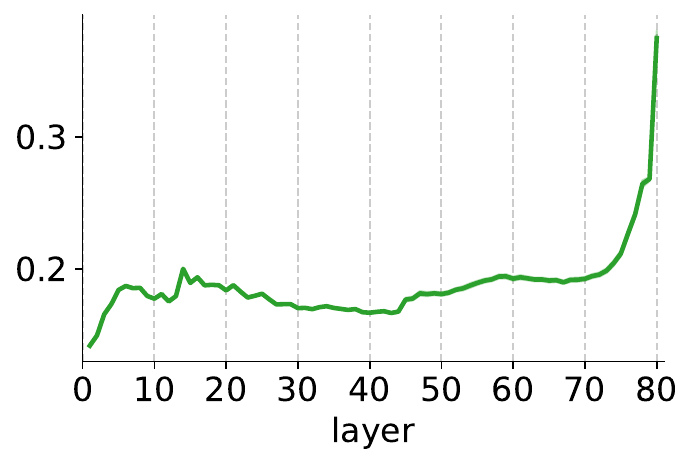}
     \end{subfigure}

     \centering
     \begin{subfigure}[b]{0.3\textwidth}
         \centering
         \includegraphics[width=\textwidth]{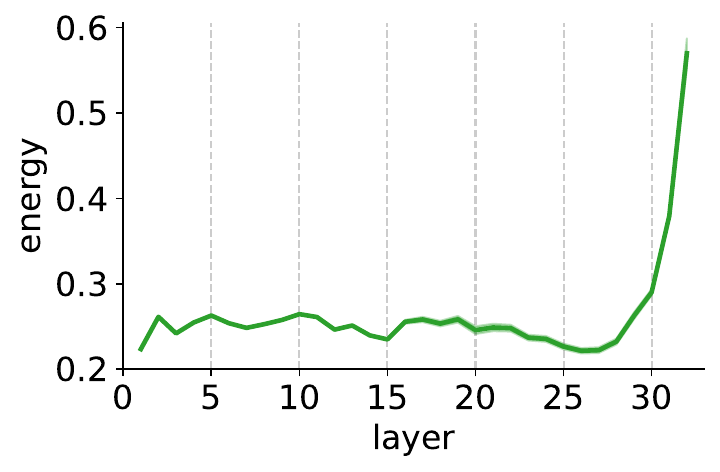}
     \end{subfigure}
     \hfill
     \begin{subfigure}[b]{0.28\textwidth}
         \centering
         \caption{Repetition (EN)}
         \includegraphics[width=\textwidth]{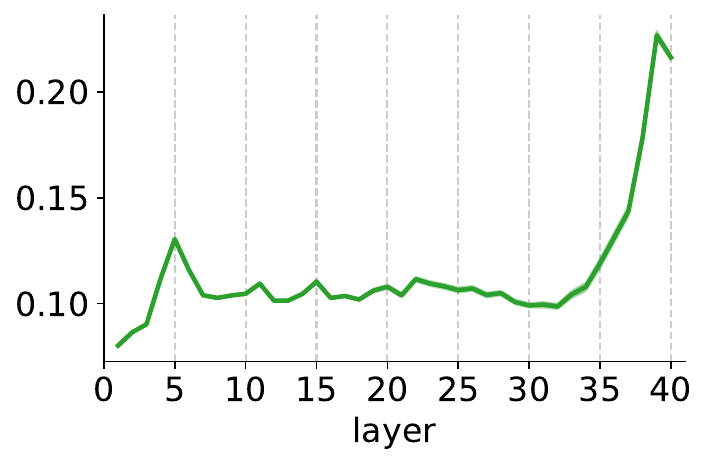}
     \end{subfigure}
     \hfill
     \begin{subfigure}[b]{0.28\textwidth}
         \centering
         \includegraphics[width=\textwidth]{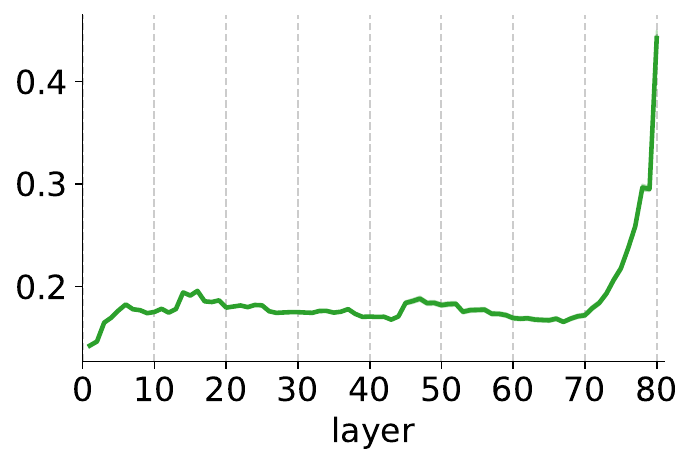}
     \end{subfigure}

     \begin{subfigure}[b]{0.3\textwidth}
         \centering
         \includegraphics[width=\textwidth]{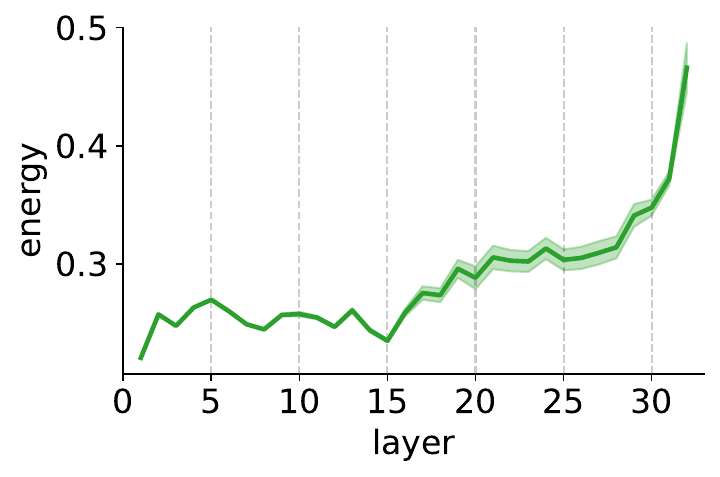}
     \end{subfigure}
     \hfill
     \begin{subfigure}[b]{0.28\textwidth}
         \centering
         \caption{Repetition (FR)}
         \includegraphics[width=\textwidth]{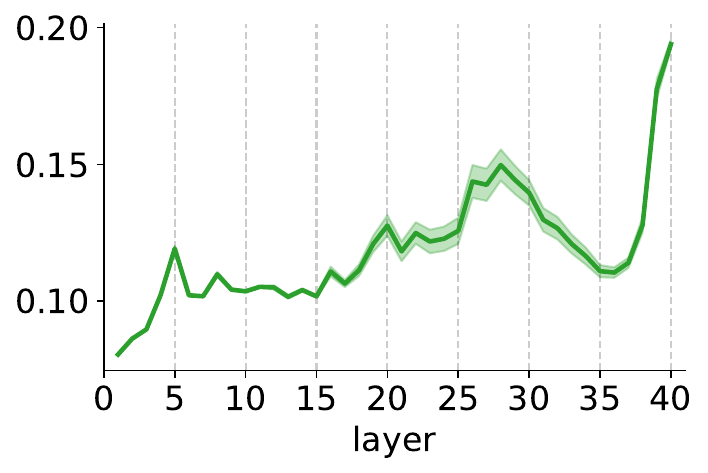}
     \end{subfigure}
     \hfill
     \begin{subfigure}[b]{0.28\textwidth}
         \centering
         \includegraphics[width=\textwidth]{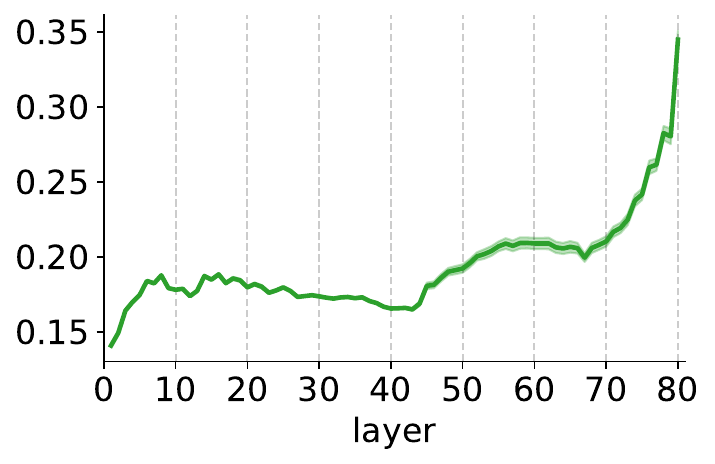}
     \end{subfigure}

     \begin{subfigure}[b]{0.3\textwidth}
         \centering
         \includegraphics[width=\textwidth]{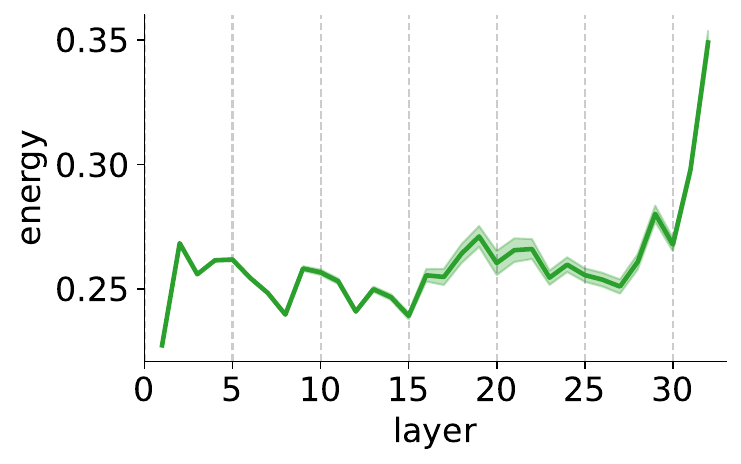}
     \end{subfigure}
     \hfill
     \begin{subfigure}[b]{0.28\textwidth}
         \centering
         \caption{Repetition (RU)}
         \includegraphics[width=\textwidth]{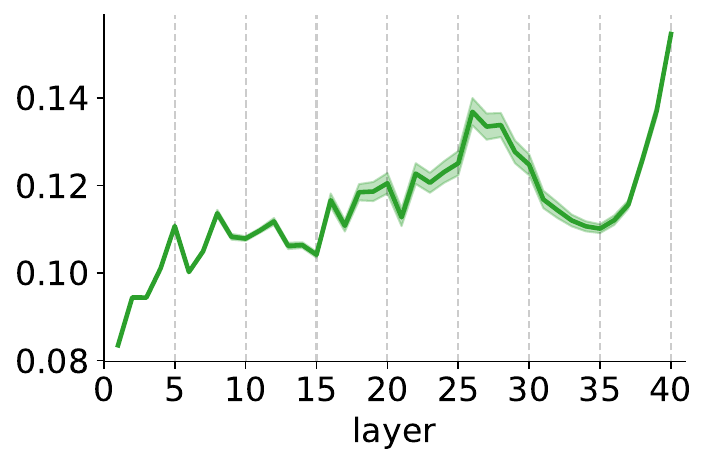}
     \end{subfigure}
     \hfill
     \begin{subfigure}[b]{0.28\textwidth}
         \centering
         \includegraphics[width=\textwidth]{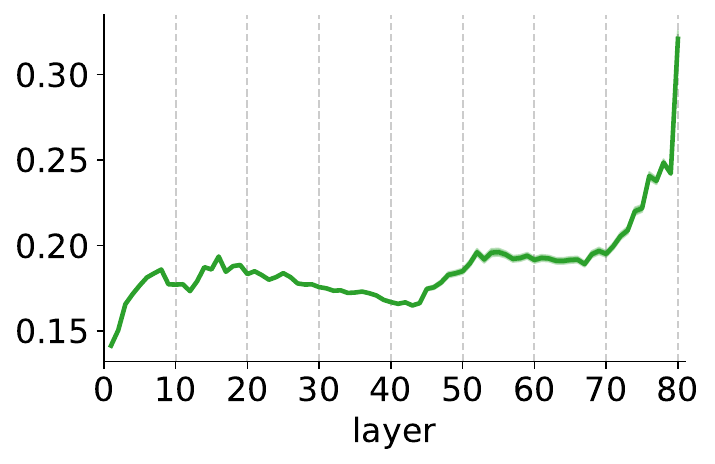}
     \end{subfigure}

     \begin{subfigure}[b]{0.3\textwidth}
         \centering
         \includegraphics[width=\textwidth]{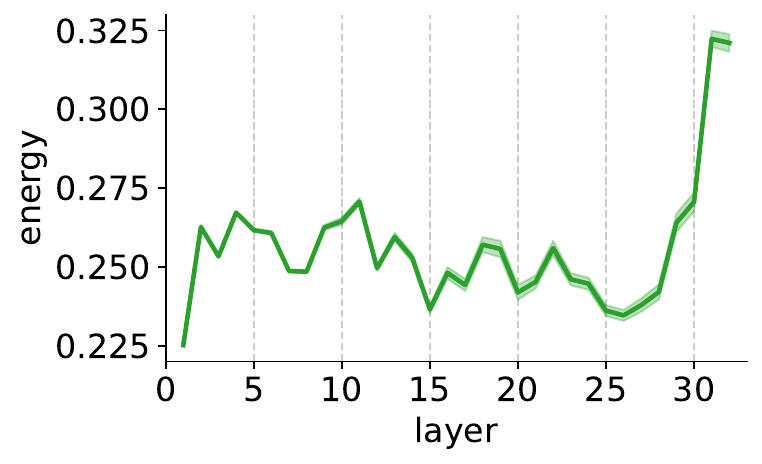}
     \end{subfigure}
     \hfill
     \begin{subfigure}[b]{0.28\textwidth}
         \centering
         \caption{Repetition (ZH)}
         \includegraphics[width=\textwidth]{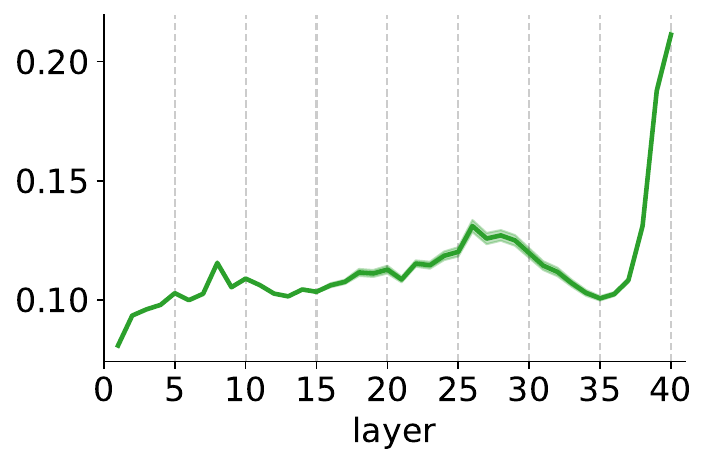}
     \end{subfigure}
     \hfill
     \begin{subfigure}[b]{0.28\textwidth}
         \centering
         \includegraphics[width=\textwidth]{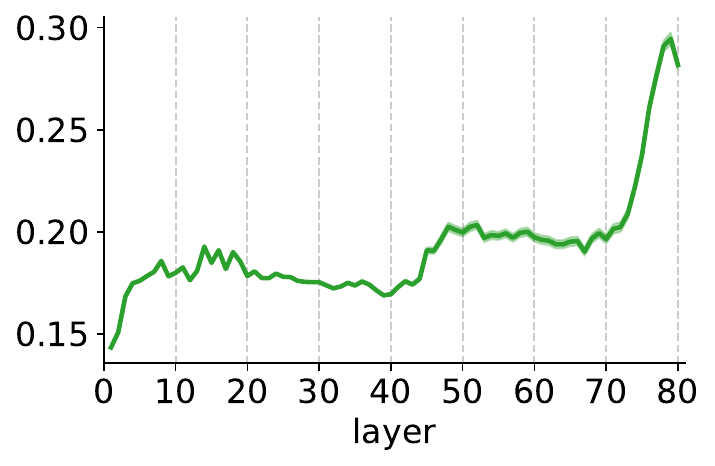}
     \end{subfigure}

     \caption{Figures illustrate the energy plots for the repetition task where \llama{} 7B, 13B, and 70B are tasked with copying a non-English word. There is one column per model size. The x-axis shows the layer number of the model, and the y-axis the energy. Means and 95\% Gaussian confidence intervals have been computed over the input examples, numbers in \Appref{app:info}.}
     \label{fig:copy-energy}
 \end{figure*}

\begin{figure*}[ht!]
     \centering
     
     \begin{subfigure}[b]{0.3\textwidth}
         \centering
         \includegraphics[width=\textwidth]{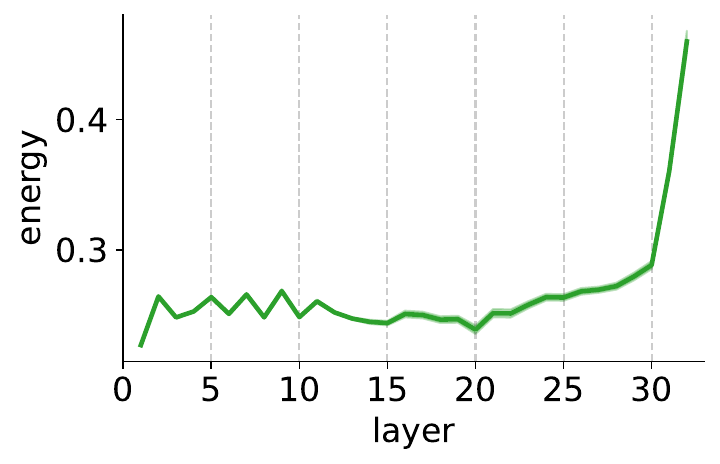}
     \end{subfigure}
     \hfill
     \begin{subfigure}[b]{0.28\textwidth}
         \centering
         \caption{Cloze task (DE)}
         \includegraphics[width=\textwidth]{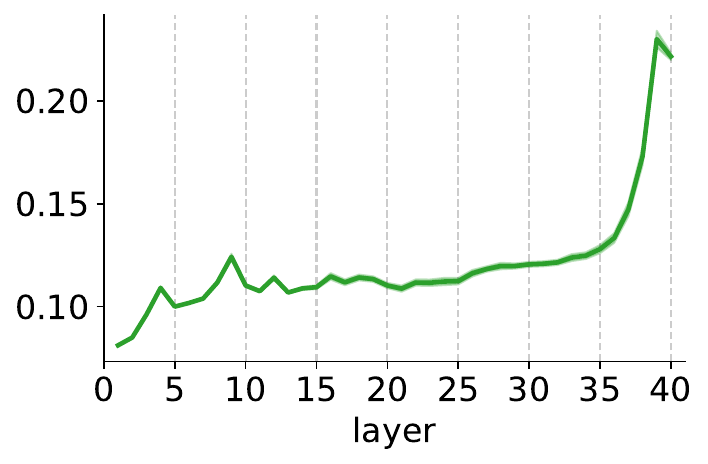}
     \end{subfigure}
     \hfill
     \begin{subfigure}[b]{0.28\textwidth}
         \centering
         \includegraphics[width=\textwidth]{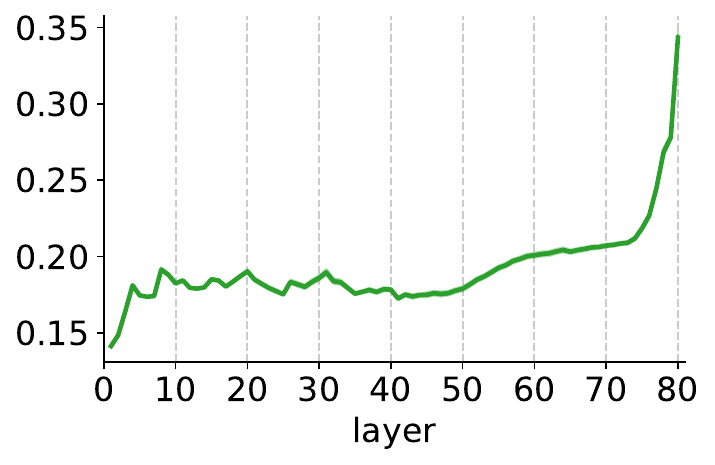}
     \end{subfigure}

      \begin{subfigure}[b]{0.3\textwidth}
         \centering
         \includegraphics[width=\textwidth]{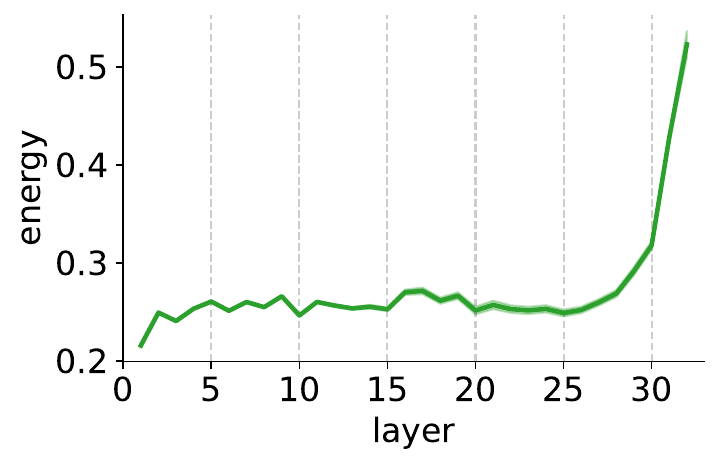}
     \end{subfigure}
     \hfill
     \begin{subfigure}[b]{0.28\textwidth}
         \centering
         \caption{Cloze task (EN)}
         \includegraphics[width=\textwidth]{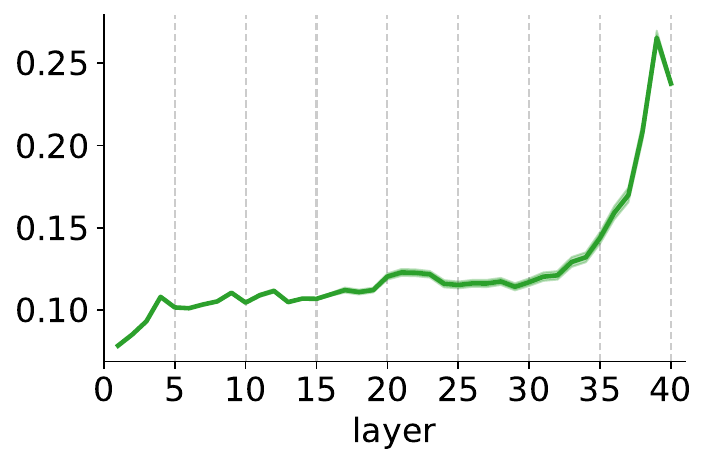}
     \end{subfigure}
     \hfill
     \begin{subfigure}[b]{0.28\textwidth}
         \centering
         \includegraphics[width=\textwidth]{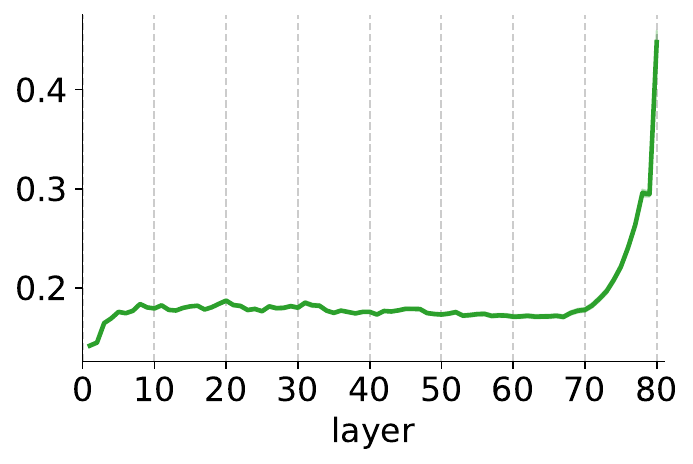}
     \end{subfigure}

     \begin{subfigure}[b]{0.3\textwidth}
         \centering
         \includegraphics[width=\textwidth]{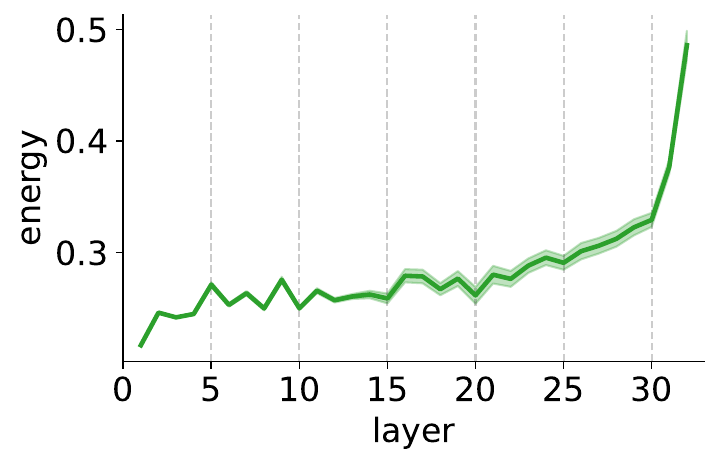}
     \end{subfigure}
     \hfill
     \begin{subfigure}[b]{0.28\textwidth}
         \centering
         \caption{Cloze task (FR)}
         \includegraphics[width=\textwidth]{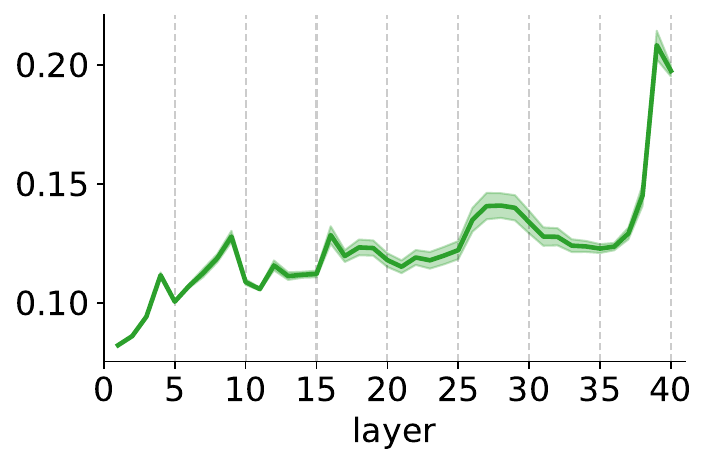}
     \end{subfigure}
     \hfill
     \begin{subfigure}[b]{0.28\textwidth}
         \centering
         \includegraphics[width=\textwidth]{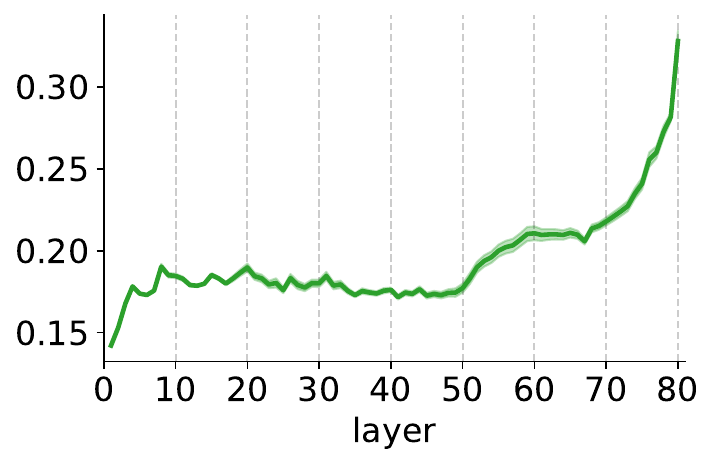}
     \end{subfigure}

     \begin{subfigure}[b]{0.3\textwidth}
         \centering
         \includegraphics[width=\textwidth]{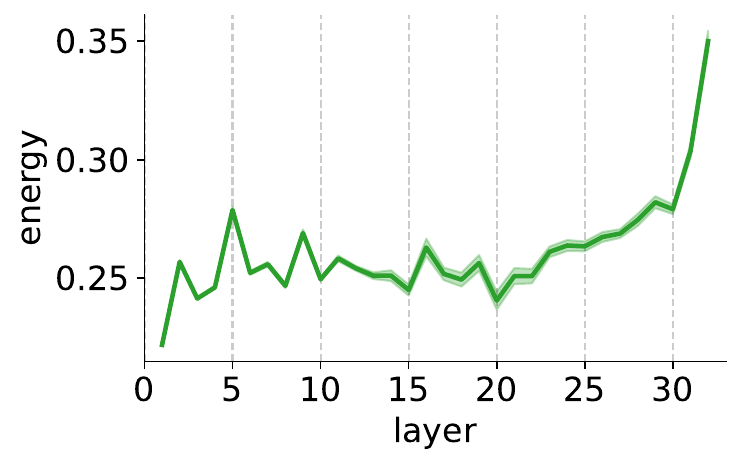}
     \end{subfigure}
     \hfill
     \begin{subfigure}[b]{0.28\textwidth}
         \centering
         \caption{Cloze task (RU)}
         \includegraphics[width=\textwidth]{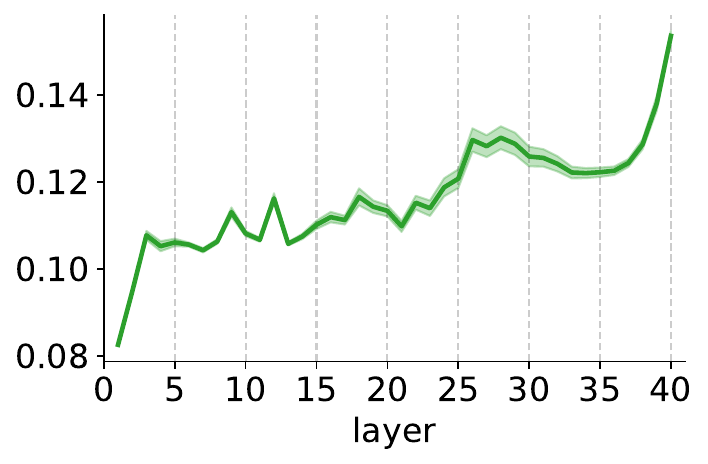}
     \end{subfigure}
     \hfill
     \begin{subfigure}[b]{0.28\textwidth}
         \centering
         \includegraphics[width=\textwidth]{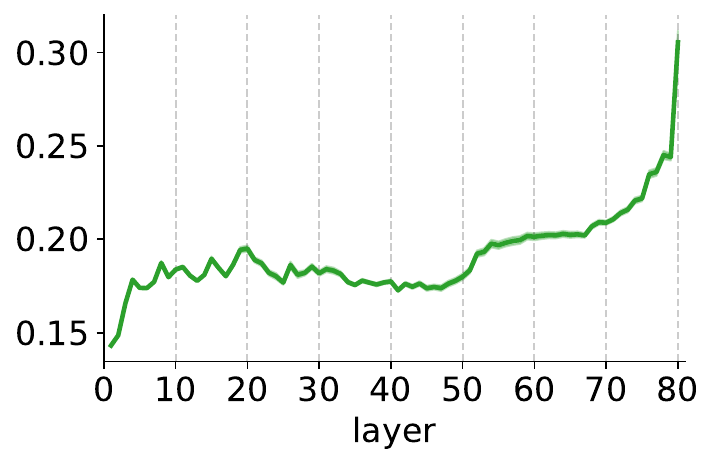}
     \end{subfigure}

     \begin{subfigure}[b]{0.3\textwidth}
         \centering
         \includegraphics[width=\textwidth]{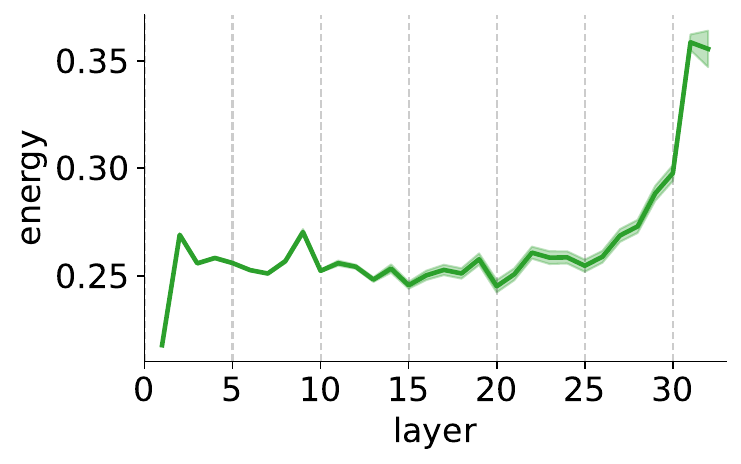}
         \caption*{7B}
     \end{subfigure}
     \hfill
     \begin{subfigure}[b]{0.28\textwidth}
         \centering
         \caption{Cloze task (ZH)}
         \includegraphics[width=\textwidth]{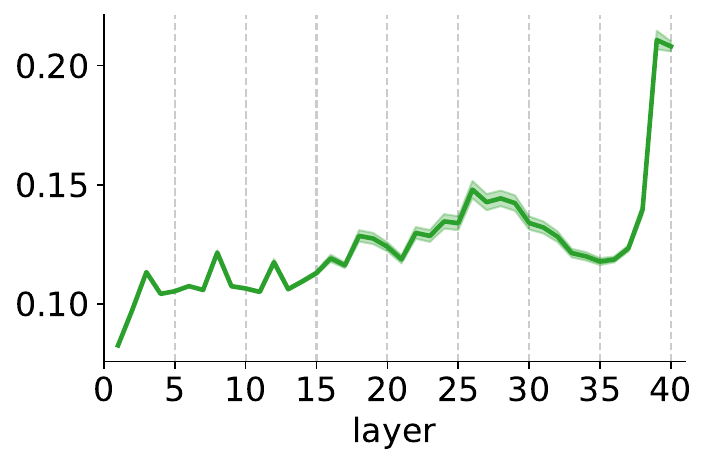}
         \caption*{13B}
     \end{subfigure}
     \hfill
     \begin{subfigure}[b]{0.28\textwidth}
         \centering
         \includegraphics[width=\textwidth]{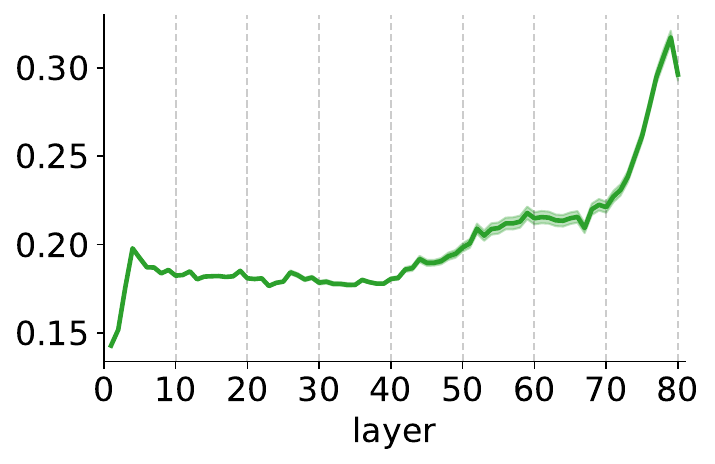}
         \caption*{70B}
     \end{subfigure}
     \caption{Figures show the same plots only for the cloze task where the correct token is defined in a fill-in-the-blank setting. In the plots, we illustrate the results for German. Means and 95\% Gaussian confidence intervals have been computed over the input examples, numbers in \Appref{app:info}.}
     \label{fig:cloze-energy}
 \end{figure*}

\begin{figure*}[ht!]
     \centering
     \begin{subfigure}[b]{0.3\textwidth}
         \centering
         \includegraphics[width=\textwidth]{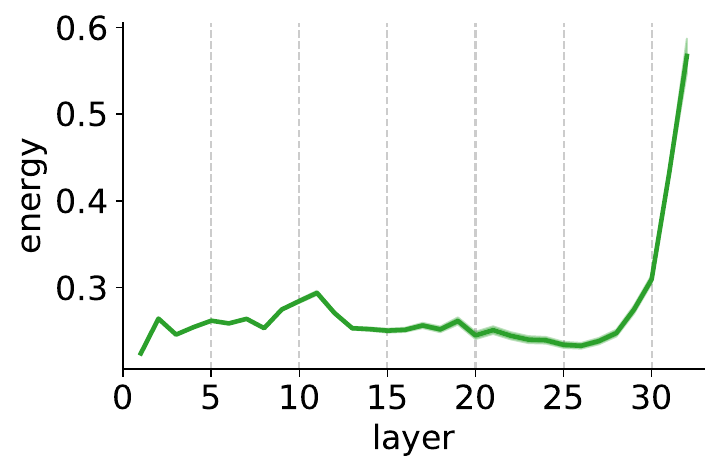}
     \end{subfigure}
     \hfill
     \begin{subfigure}[b]{0.28\textwidth}
         \centering
         \caption{Translation (DE->EN)}
         \includegraphics[width=\textwidth]{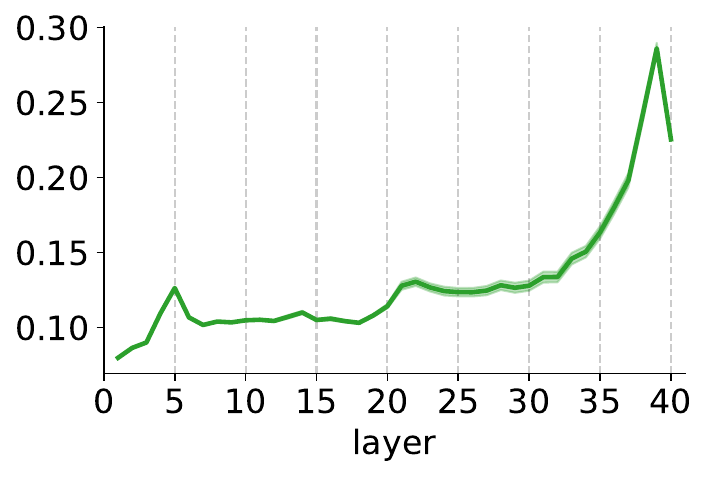}
     \end{subfigure}
     \hfill
     \begin{subfigure}[b]{0.28\textwidth}
         \centering
         \includegraphics[width=\textwidth]{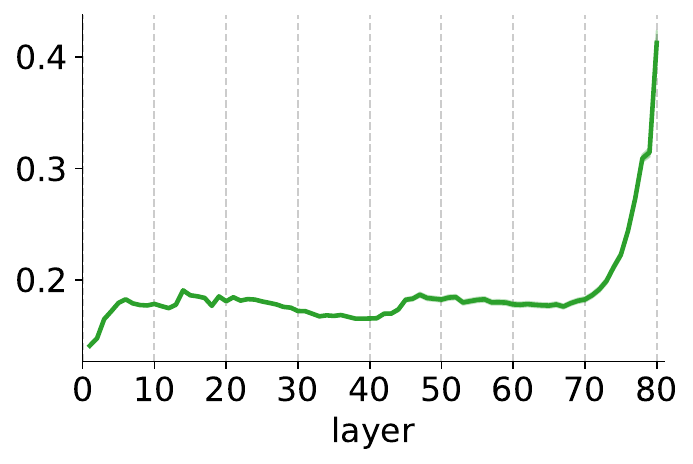}
     \end{subfigure}
     
     \begin{subfigure}[b]{0.3\textwidth}
         \centering
         \includegraphics[width=\textwidth]{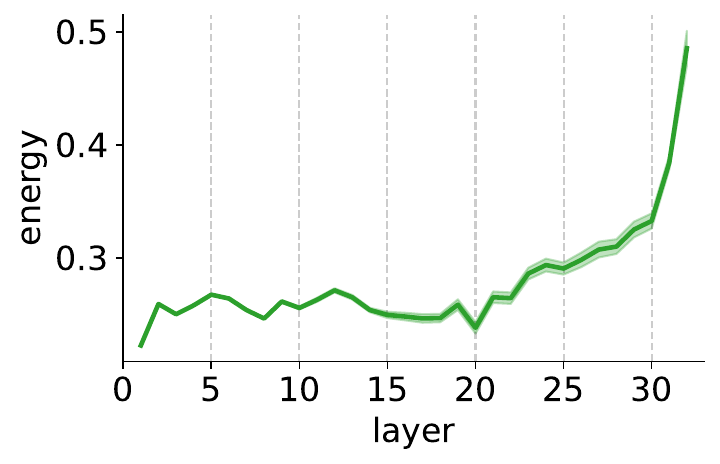}
     \end{subfigure}
     \hfill
     \begin{subfigure}[b]{0.28\textwidth}
         \centering
         \caption{Translation (DE->FR)}
         \includegraphics[width=\textwidth]{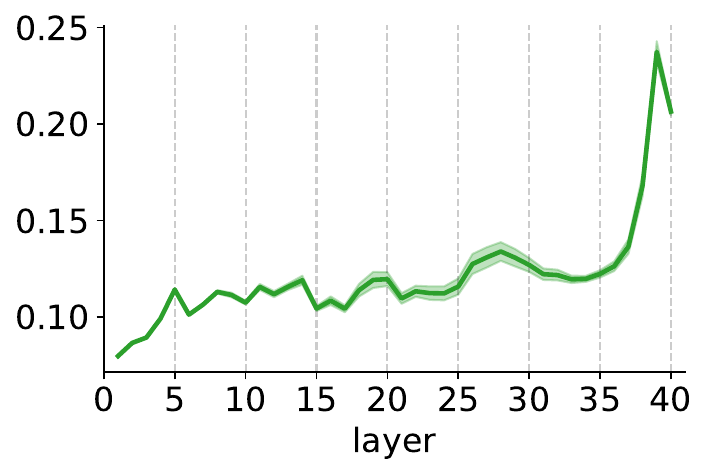}
     \end{subfigure}
     \hfill
     \begin{subfigure}[b]{0.28\textwidth}
         \centering
         \includegraphics[width=\textwidth]{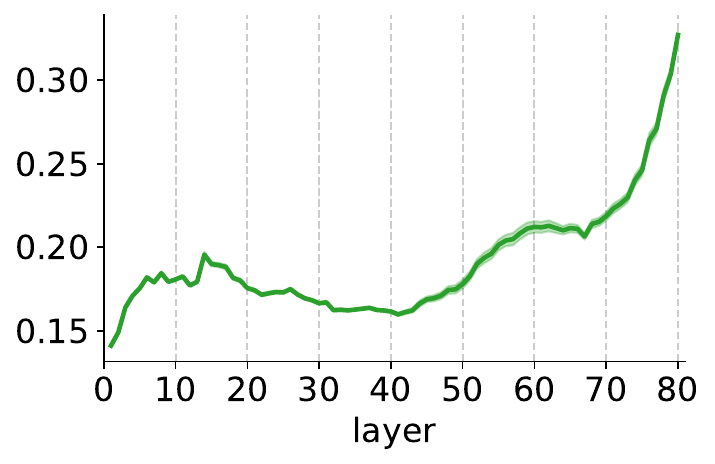}
     \end{subfigure}

     \begin{subfigure}[b]{0.3\textwidth}
         \centering
         \includegraphics[width=\textwidth]{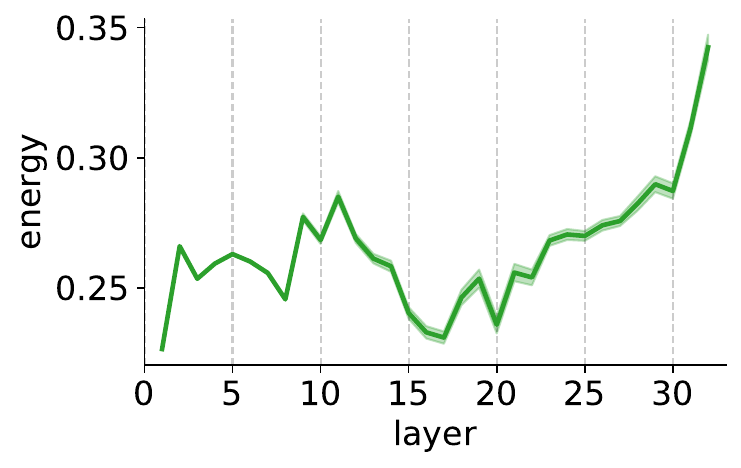}
     \end{subfigure}
     \hfill
     \begin{subfigure}[b]{0.28\textwidth}
         \centering
         \caption{Translation (DE->RU)}
         \includegraphics[width=\textwidth]{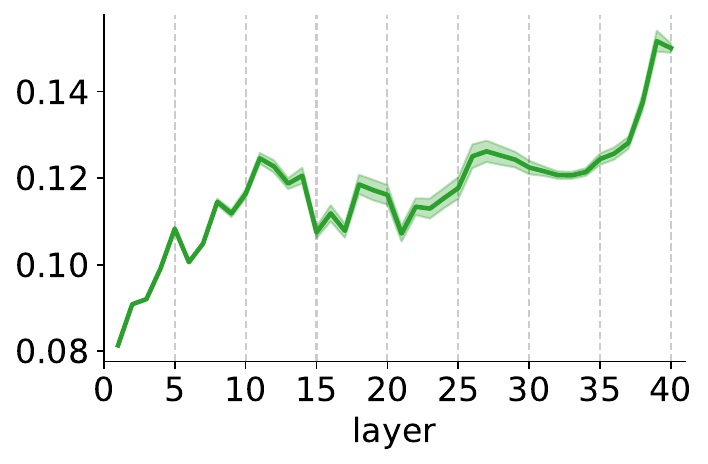}
     \end{subfigure}
     \hfill
     \begin{subfigure}[b]{0.28\textwidth}
         \centering
         \includegraphics[width=\textwidth]{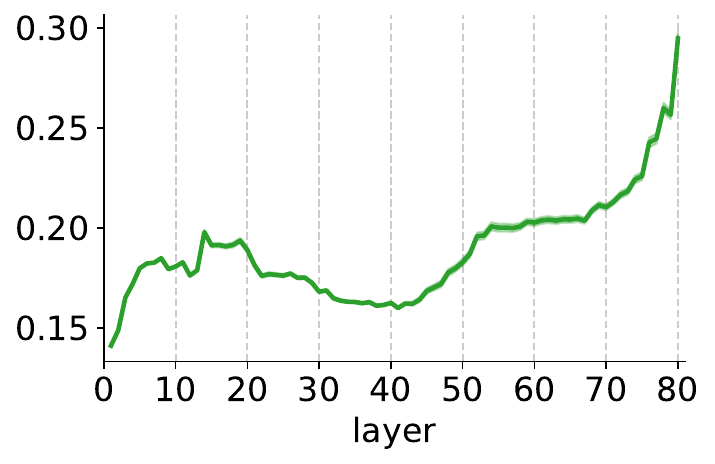}
     \end{subfigure}

     \begin{subfigure}[b]{0.3\textwidth}
         \centering
         \includegraphics[width=\textwidth]{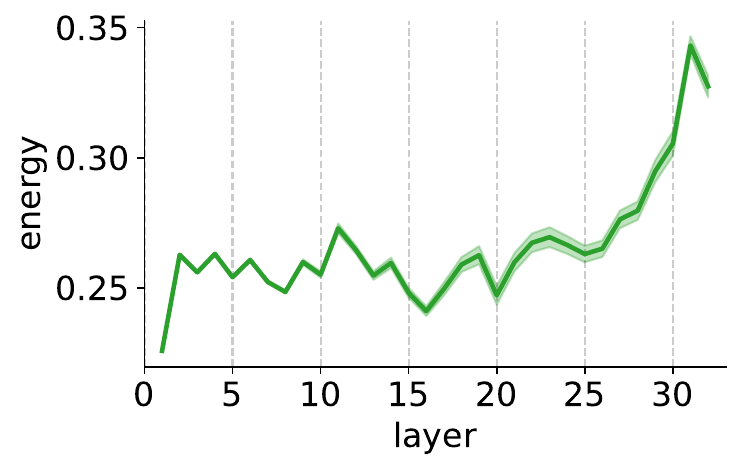}
     \end{subfigure}
     \hfill
     \begin{subfigure}[b]{0.28\textwidth}
         \centering
         \caption{Translation (DE->ZH)}
         \includegraphics[width=\textwidth]{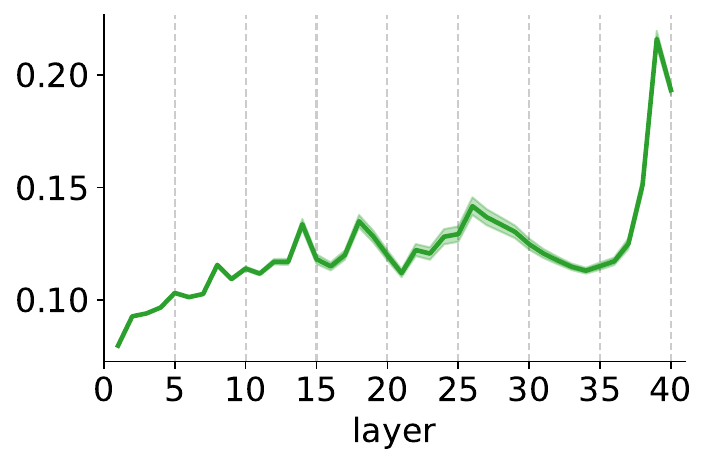}
     \end{subfigure}
     \hfill
     \begin{subfigure}[b]{0.28\textwidth}
         \centering
         \includegraphics[width=\textwidth]{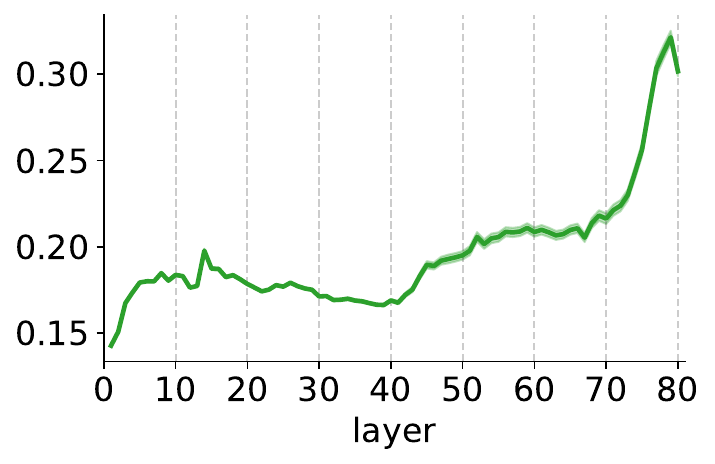}
     \end{subfigure}
     
     \caption{Figures illustrate the translation task where \llama{} 7B, 13B, and 70B are tasked with translating a word from non-English input language to output language. There is one column per model size. The x-axis shows the layer number of the model, and the y-axis the energy. Means and 95\% Gaussian confidence intervals have been computed over the input examples, numbers in \Appref{app:info}.}
     \label{fig:translation-de-energy}
 \end{figure*}

\begin{figure*}[ht!]
     \centering
     \begin{subfigure}[b]{0.3\textwidth}
         \centering
         \includegraphics[width=\textwidth]{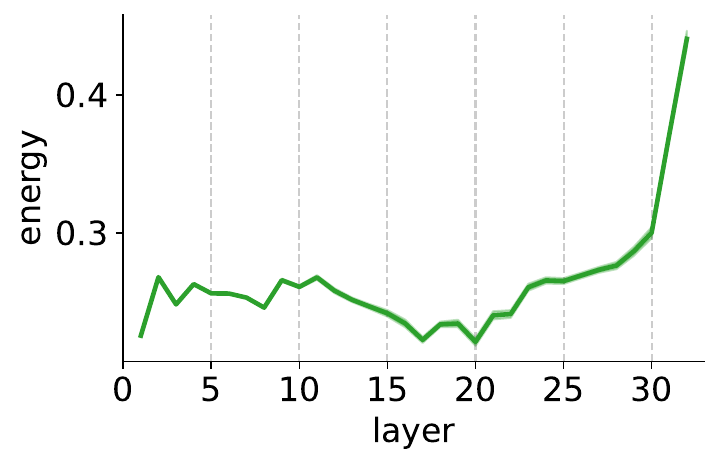}
     \end{subfigure}
     \hfill
     \begin{subfigure}[b]{0.28\textwidth}
         \centering
         \caption{Translation (EN->DE)}
         \includegraphics[width=\textwidth]{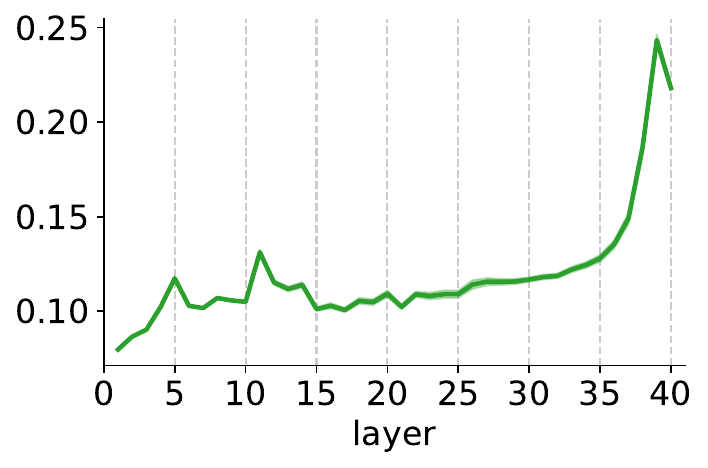}
     \end{subfigure}
     \hfill
     \begin{subfigure}[b]{0.28\textwidth}
         \centering
         \includegraphics[width=\textwidth]{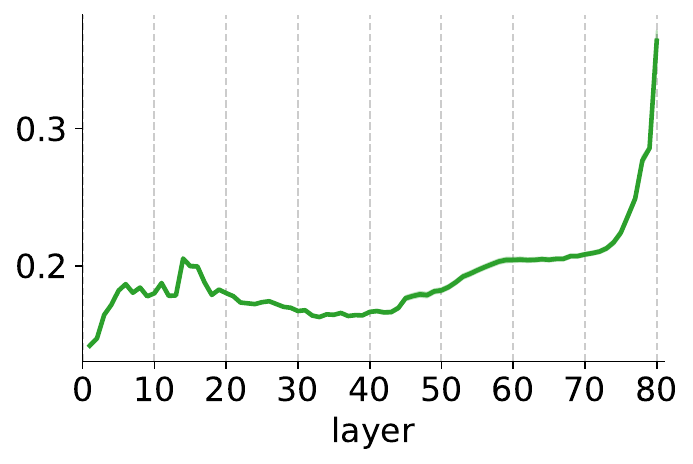}
     \end{subfigure}
     
     \begin{subfigure}[b]{0.3\textwidth}
         \centering
         \includegraphics[width=\textwidth]{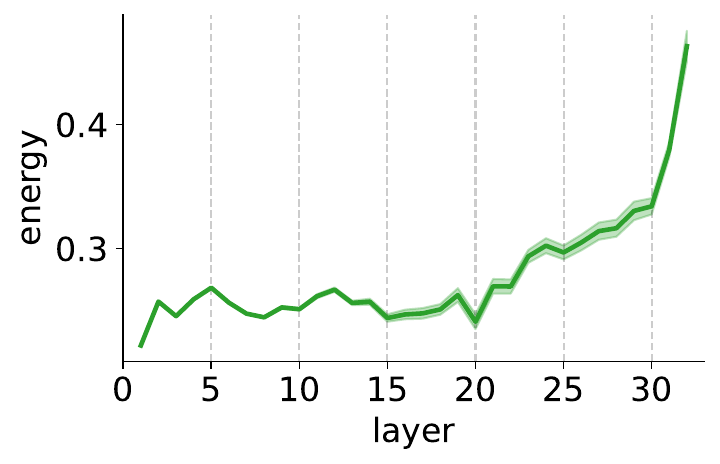}
     \end{subfigure}
     \hfill
     \begin{subfigure}[b]{0.28\textwidth}
         \centering
         \caption{Translation (EN->FR)}
         \includegraphics[width=\textwidth]{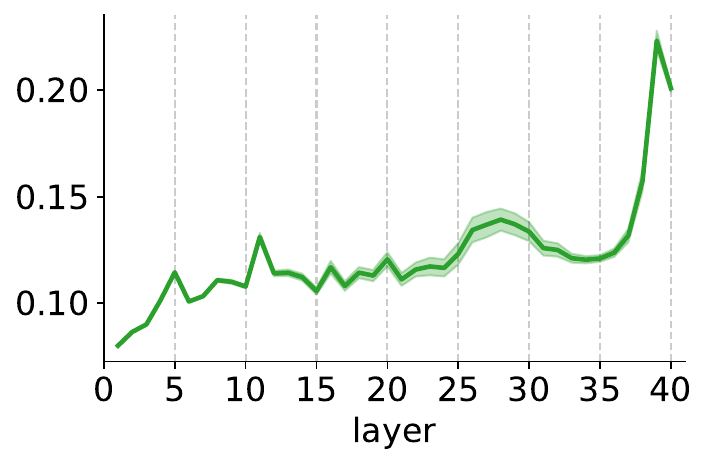}
     \end{subfigure}
     \hfill
     \begin{subfigure}[b]{0.28\textwidth}
         \centering
         \includegraphics[width=\textwidth]{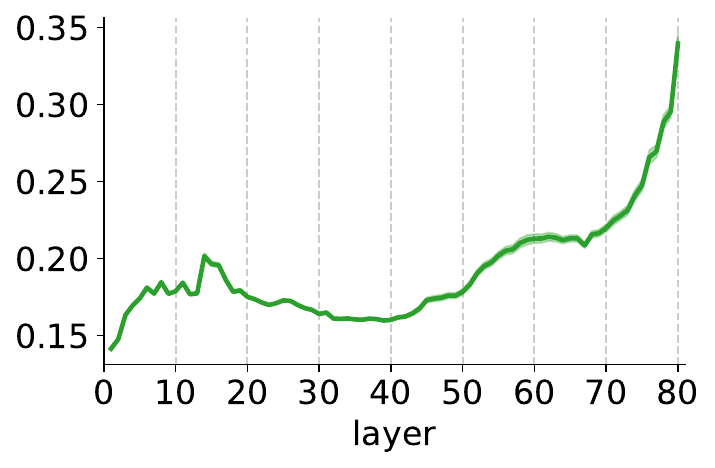}
     \end{subfigure}

     \begin{subfigure}[b]{0.3\textwidth}
         \centering
         \includegraphics[width=\textwidth]{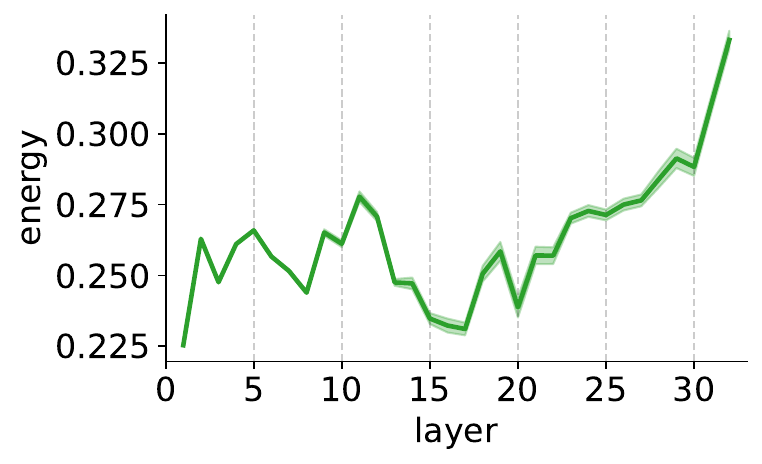}
     \end{subfigure}
     \hfill
     \begin{subfigure}[b]{0.28\textwidth}
         \centering
         \caption{Translation (EN->RU)}
         \includegraphics[width=\textwidth]{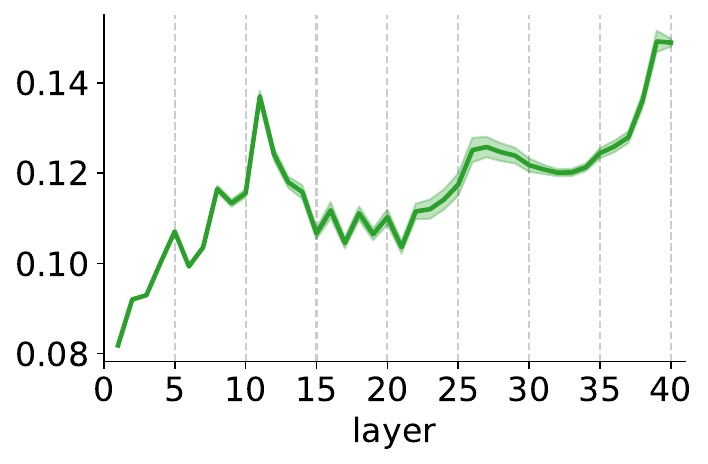}
     \end{subfigure}
     \hfill
     \begin{subfigure}[b]{0.28\textwidth}
         \centering
         \includegraphics[width=\textwidth]{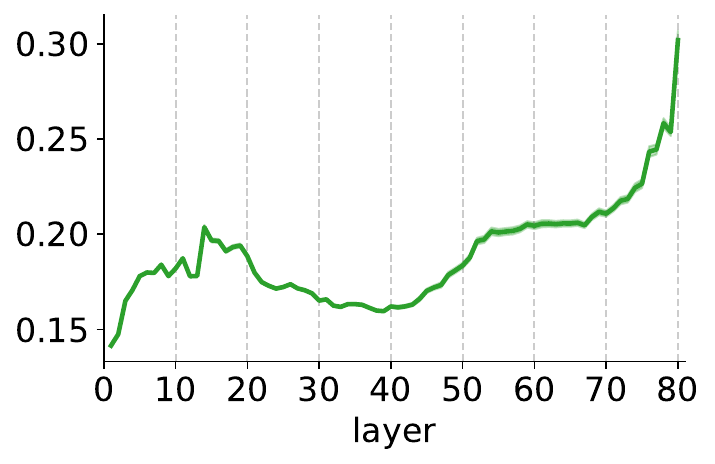}
     \end{subfigure}

     \begin{subfigure}[b]{0.3\textwidth}
         \centering
         \includegraphics[width=\textwidth]{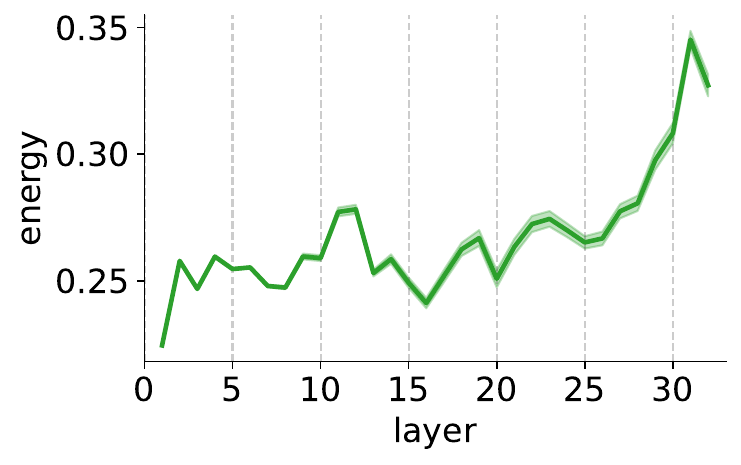}
     \end{subfigure}
     \hfill
     \begin{subfigure}[b]{0.28\textwidth}
         \centering
         \caption{Translation (EN->ZH)}
         \includegraphics[width=\textwidth]{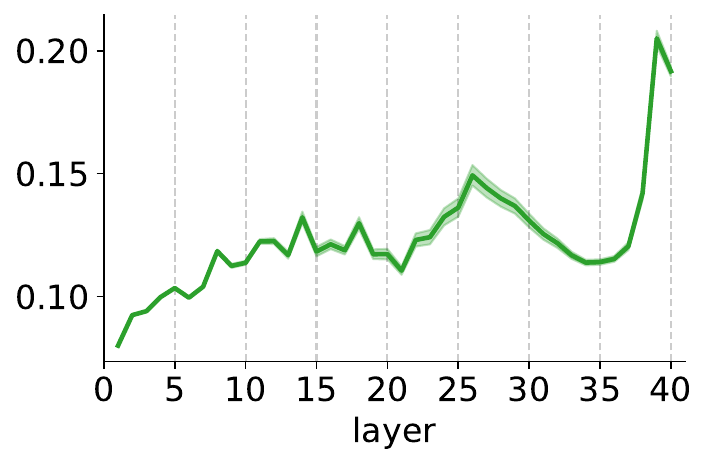}
     \end{subfigure}
     \hfill
     \begin{subfigure}[b]{0.28\textwidth}
         \centering
         \includegraphics[width=\textwidth]{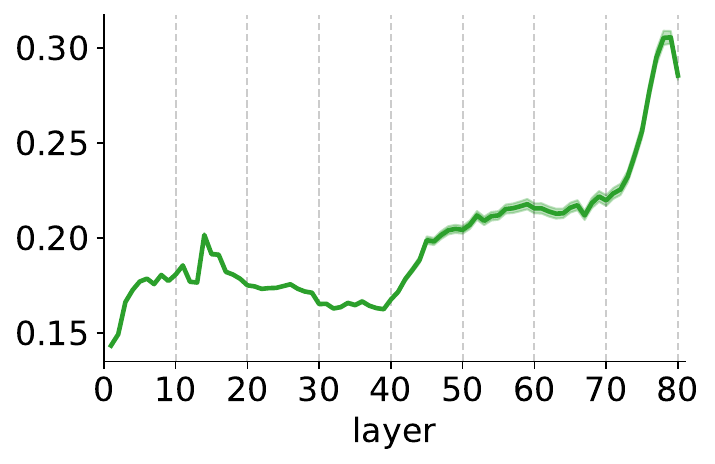}
     \end{subfigure}
     
     \caption{Figures illustrate the translation task where \llama{} 7B, 13B, and 70B are tasked with translating a word from English input language to output language. There is one column per model size. The x-axis shows the layer number of the model, and the y-axis the energy. Means and 95\% Gaussian confidence intervals have been computed over the input examples, numbers in \Appref{app:info}.}
     \label{fig:translation-en-energy}
 \end{figure*}

\begin{figure*}[ht!]
     \centering
     \begin{subfigure}[b]{0.3\textwidth}
         \centering
         \includegraphics[width=\textwidth]{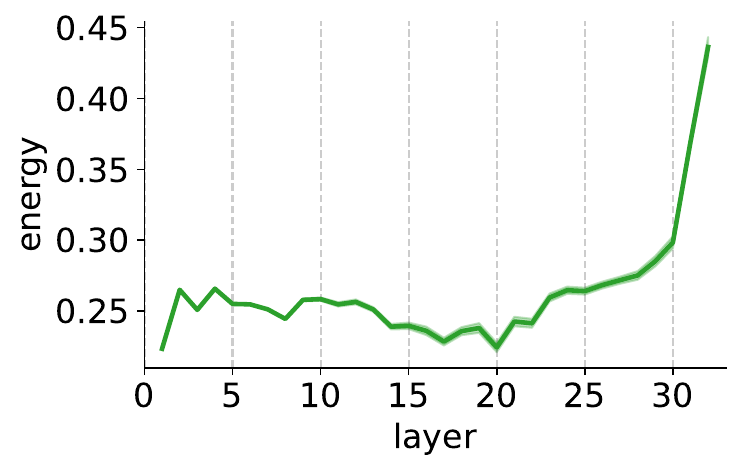}
     \end{subfigure}
     \hfill
     \begin{subfigure}[b]{0.28\textwidth}
         \centering
         \caption{Translation (FR->DE)}
         \includegraphics[width=\textwidth]{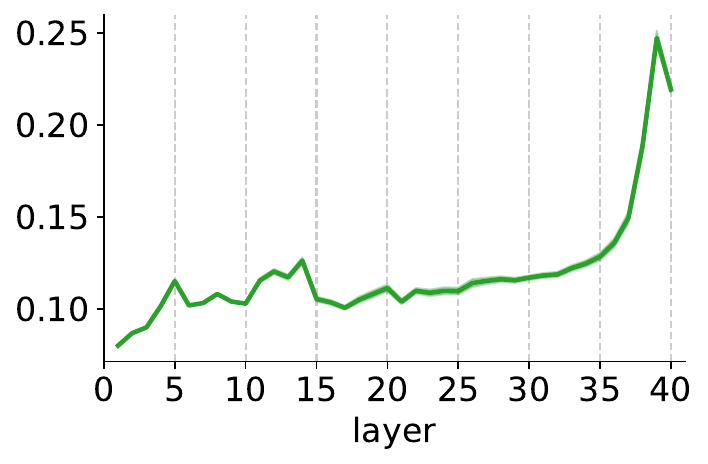}
     \end{subfigure}
     \hfill
     \begin{subfigure}[b]{0.28\textwidth}
         \centering
         \includegraphics[width=\textwidth]{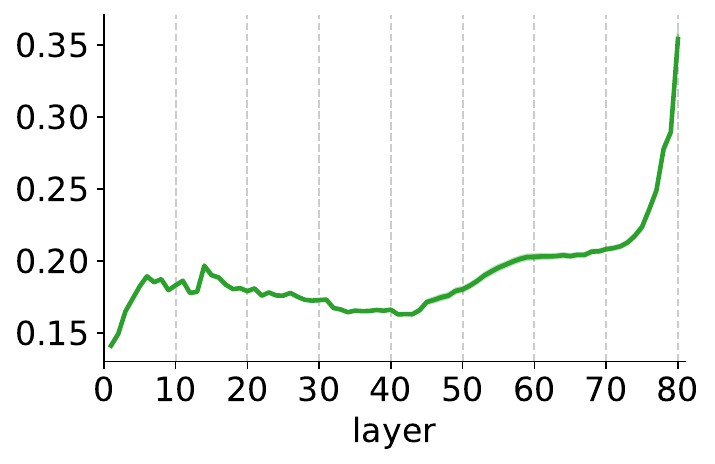}
     \end{subfigure}
     
     \begin{subfigure}[b]{0.3\textwidth}
         \centering
         \includegraphics[width=\textwidth]{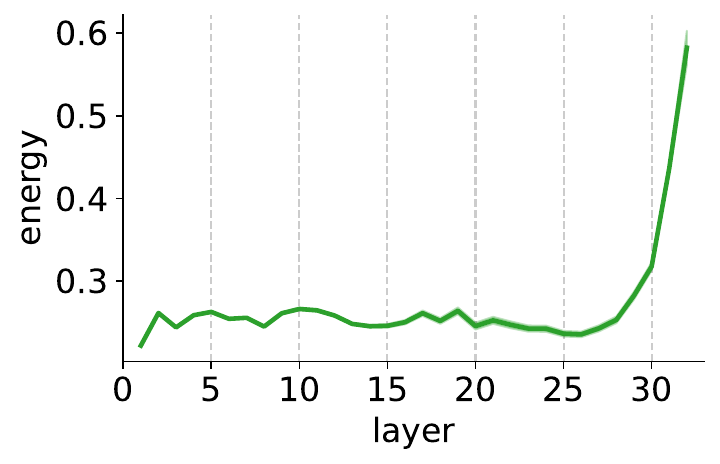}
     \end{subfigure}
     \hfill
     \begin{subfigure}[b]{0.28\textwidth}
         \centering
         \caption{Translation (FR->EN)}
         \includegraphics[width=\textwidth]{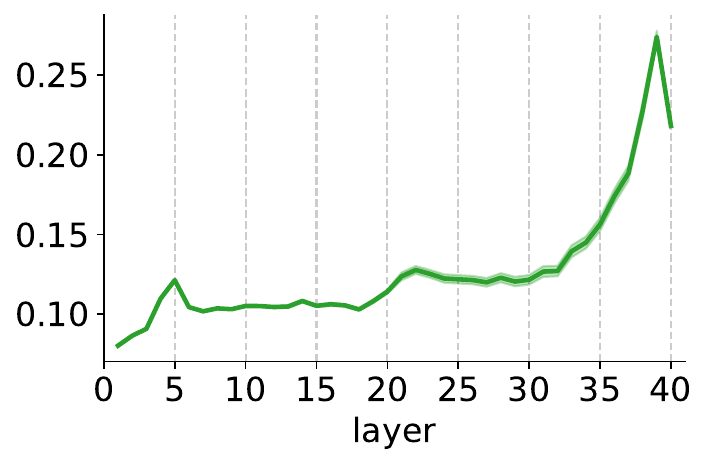}
     \end{subfigure}
     \hfill
     \begin{subfigure}[b]{0.28\textwidth}
         \centering
         \includegraphics[width=\textwidth]{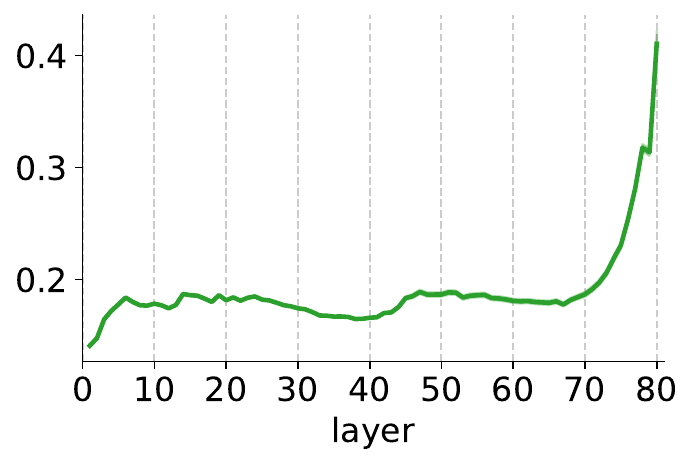}
     \end{subfigure}

     \begin{subfigure}[b]{0.3\textwidth}
         \centering
         \includegraphics[width=\textwidth]{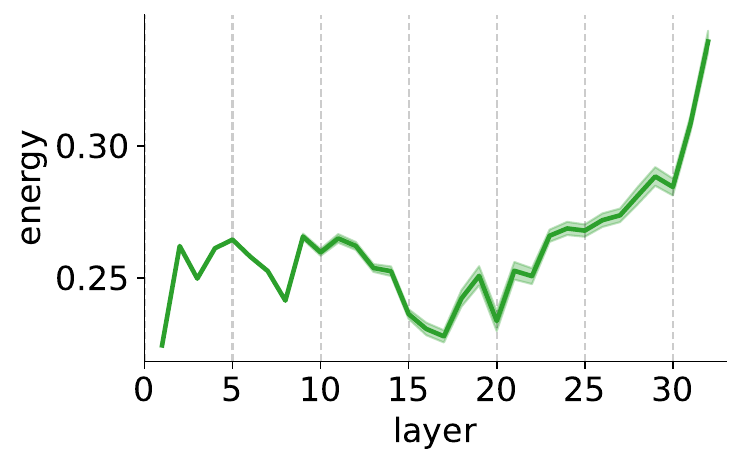}
     \end{subfigure}
     \hfill
     \begin{subfigure}[b]{0.28\textwidth}
         \centering
         \caption{Translation (FR->RU)}
         \includegraphics[width=\textwidth]{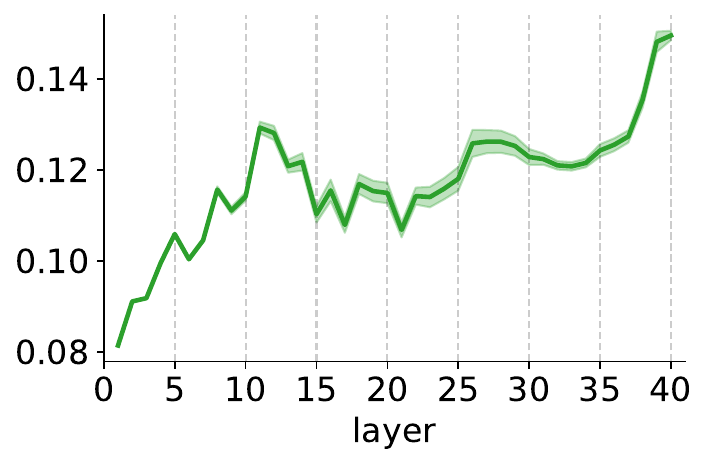}
     \end{subfigure}
     \hfill
     \begin{subfigure}[b]{0.28\textwidth}
         \centering
         \includegraphics[width=\textwidth]{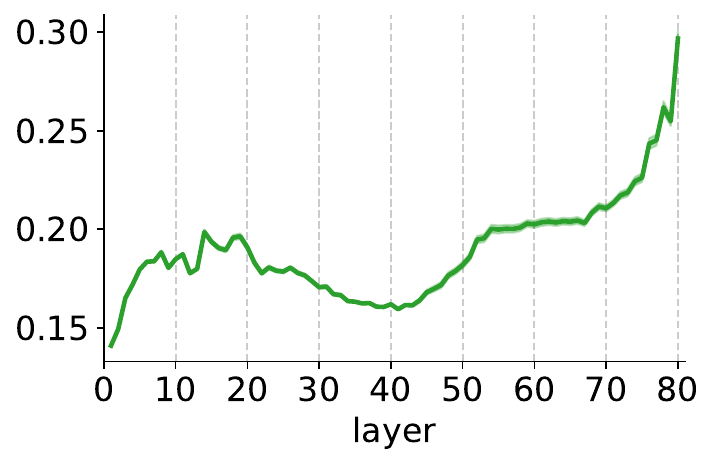}
     \end{subfigure}

     \begin{subfigure}[b]{0.3\textwidth}
         \centering
         \includegraphics[width=\textwidth]{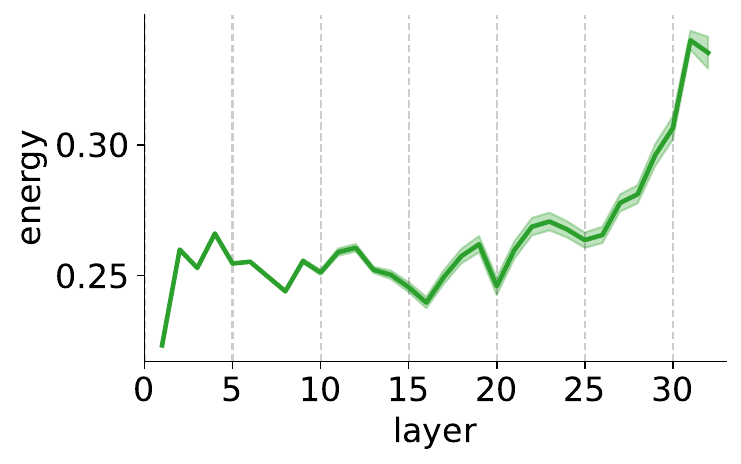}
     \end{subfigure}
     \hfill
     \begin{subfigure}[b]{0.28\textwidth}
         \centering
         \caption{Translation (FR->ZH)}
         \includegraphics[width=\textwidth]{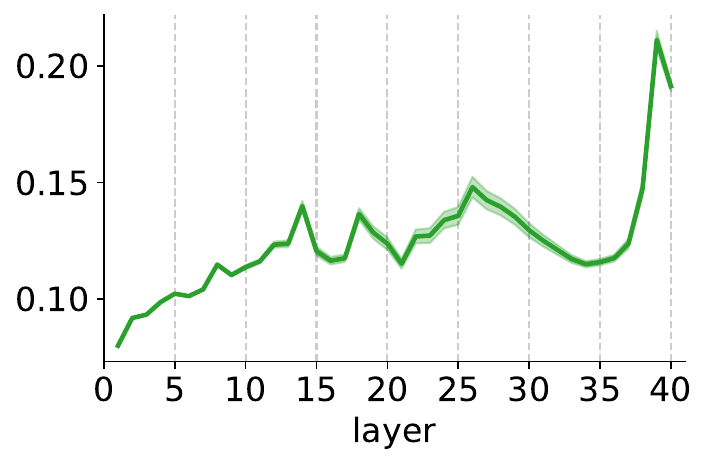}
     \end{subfigure}
     \hfill
     \begin{subfigure}[b]{0.28\textwidth}
         \centering
         \includegraphics[width=\textwidth]{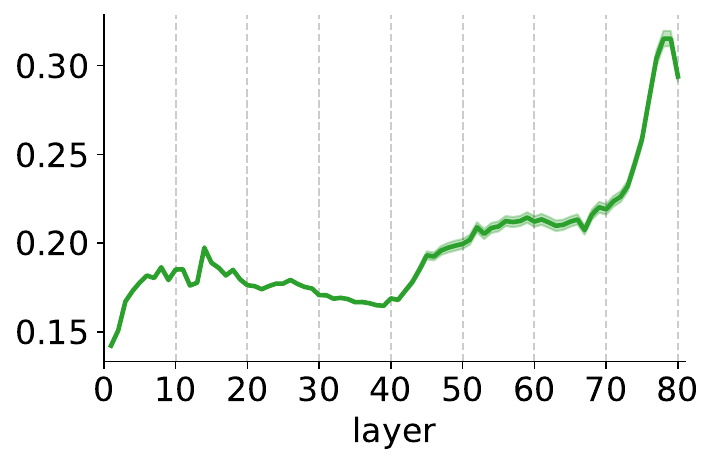}
     \end{subfigure}
     
     \caption{Figures illustrate the translation task where \llama{} 7B, 13B, and 70B are tasked with translating a word from non-English input language to output language. There is one column per model size. The x-axis shows the layer number of the model, and the y-axis the energy.Means and 95\% Gaussian confidence intervals have been computed over the input examples, numbers in \Appref{app:info}.}
     \label{fig:translation-fr-energy}
 \end{figure*}

\begin{figure*}[ht!]
     \centering
     \begin{subfigure}[b]{0.3\textwidth}
         \centering
         \includegraphics[width=\textwidth]{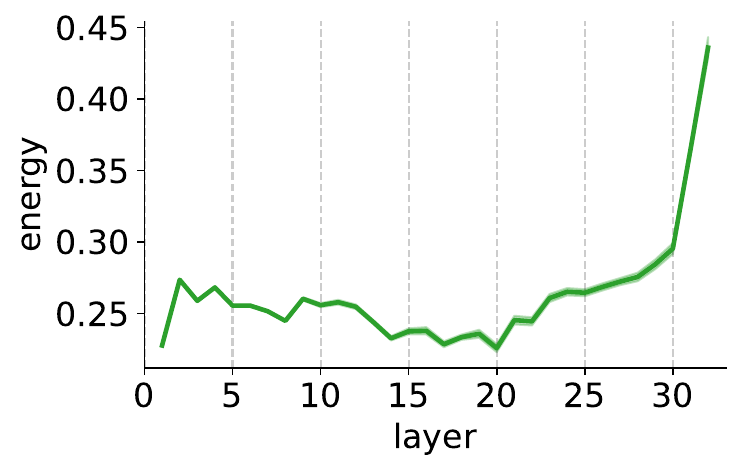}
     \end{subfigure}
     \hfill
     \begin{subfigure}[b]{0.28\textwidth}
         \centering
         \caption{Translation (RU->DE)}
         \includegraphics[width=\textwidth]{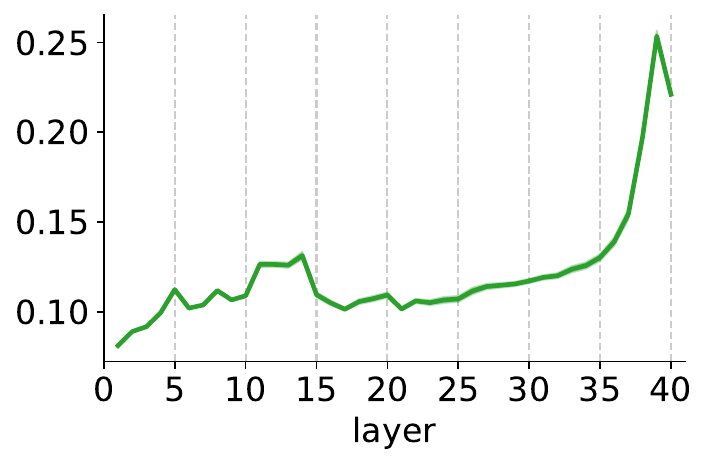}
     \end{subfigure}
     \hfill
     \begin{subfigure}[b]{0.28\textwidth}
         \centering
         \includegraphics[width=\textwidth]{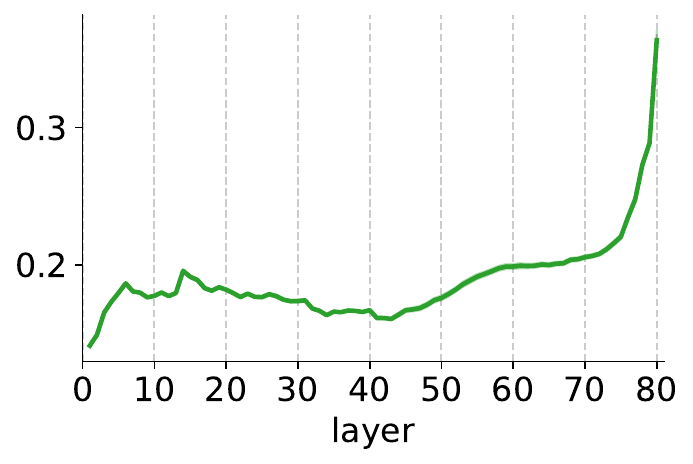}
     \end{subfigure}
     
     \begin{subfigure}[b]{0.3\textwidth}
         \centering
         \includegraphics[width=\textwidth]{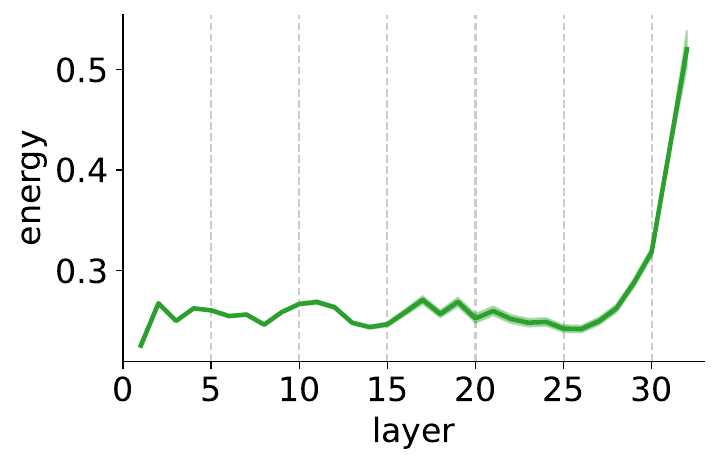}
     \end{subfigure}
     \hfill
     \begin{subfigure}[b]{0.28\textwidth}
         \centering
         \caption{Translation (RU->EN)}
         \includegraphics[width=\textwidth]{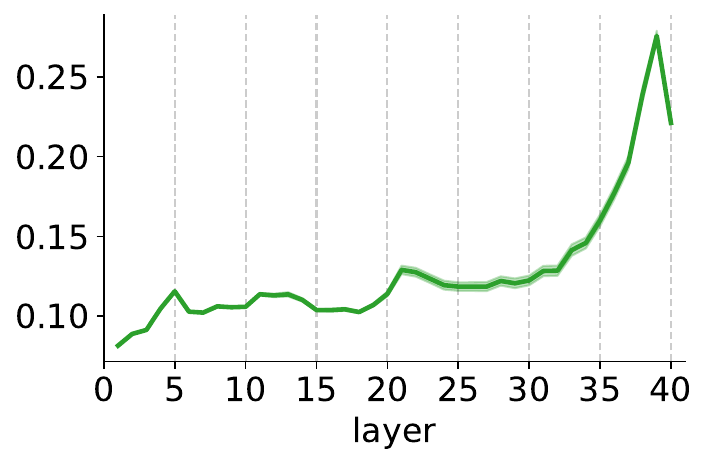}
     \end{subfigure}
     \hfill
     \begin{subfigure}[b]{0.28\textwidth}
         \centering
         \includegraphics[width=\textwidth]{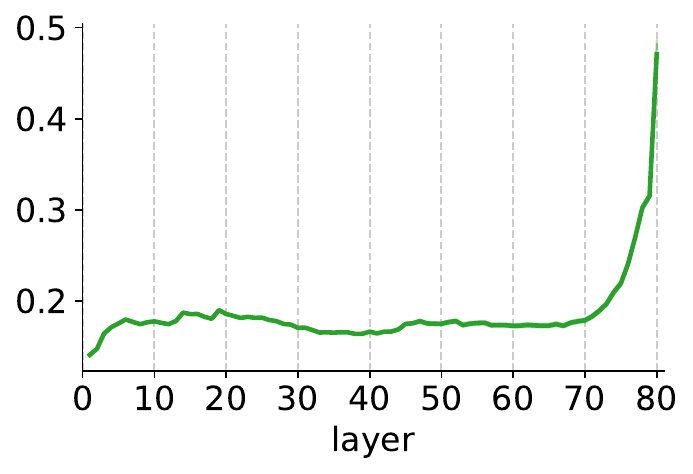}
     \end{subfigure}

     \begin{subfigure}[b]{0.3\textwidth}
         \centering
         \includegraphics[width=\textwidth]{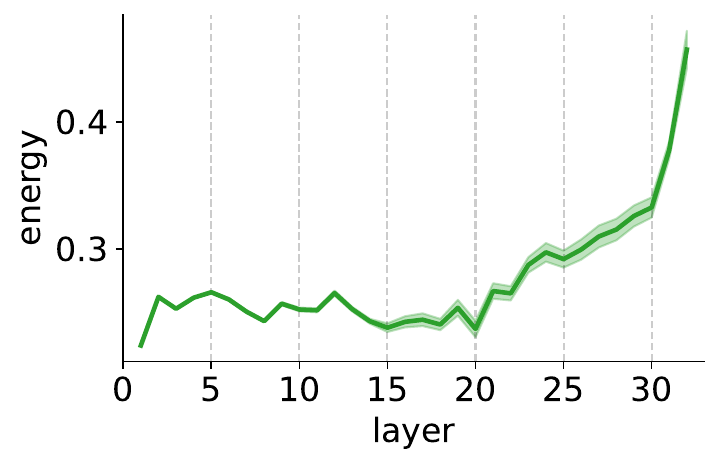}
     \end{subfigure}
     \hfill
     \begin{subfigure}[b]{0.28\textwidth}
         \centering
         \caption{Translation (RU->FR)}
         \includegraphics[width=\textwidth]{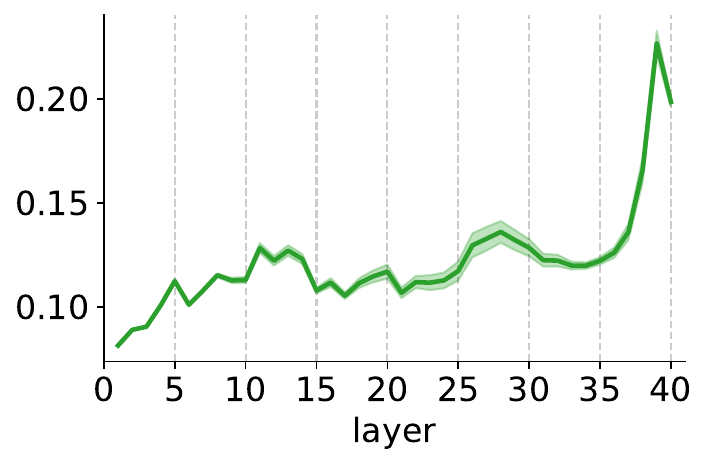}
     \end{subfigure}
     \hfill
     \begin{subfigure}[b]{0.28\textwidth}
         \centering
         \includegraphics[width=\textwidth]{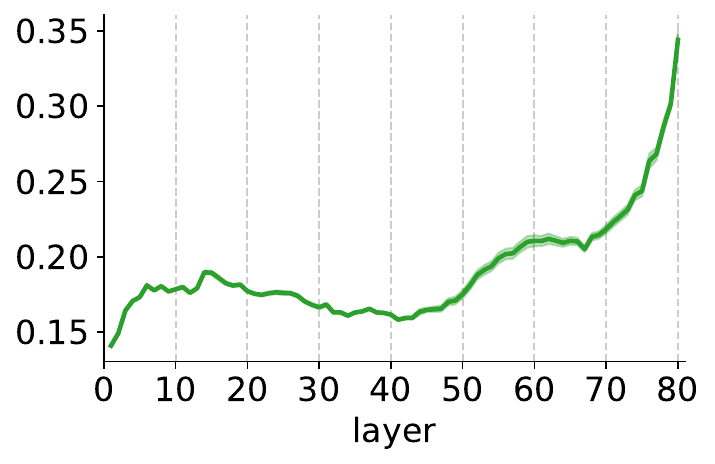}
     \end{subfigure}

     \begin{subfigure}[b]{0.3\textwidth}
         \centering
         \includegraphics[width=\textwidth]{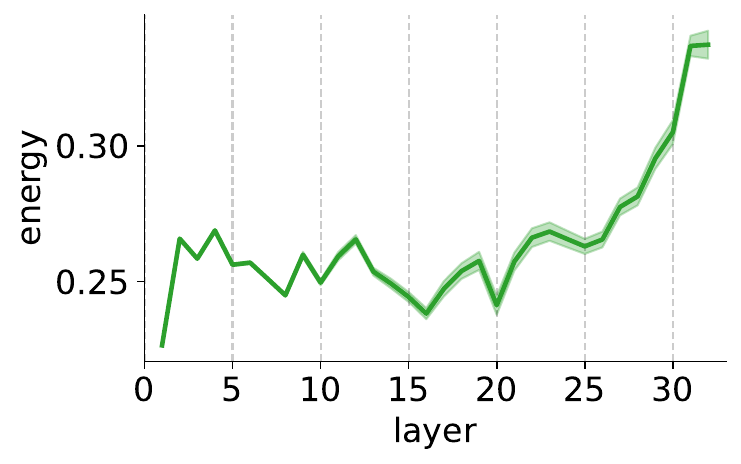}
     \end{subfigure}
     \hfill
     \begin{subfigure}[b]{0.28\textwidth}
         \centering
         \caption{Translation (RU->ZH)}
         \includegraphics[width=\textwidth]{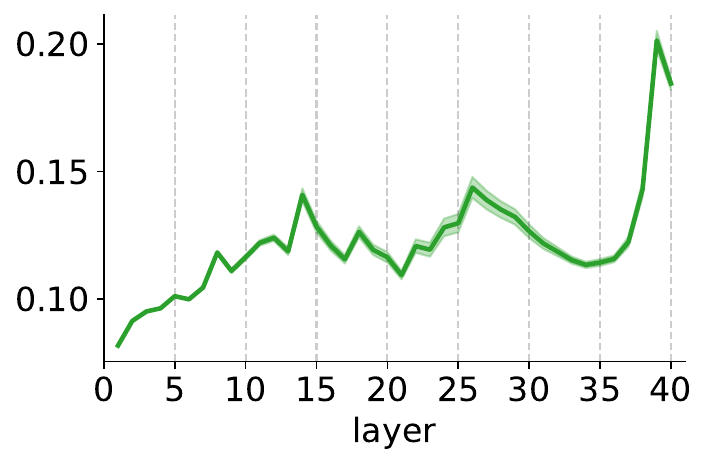}
     \end{subfigure}
     \hfill
     \begin{subfigure}[b]{0.28\textwidth}
         \centering
         \includegraphics[width=\textwidth]{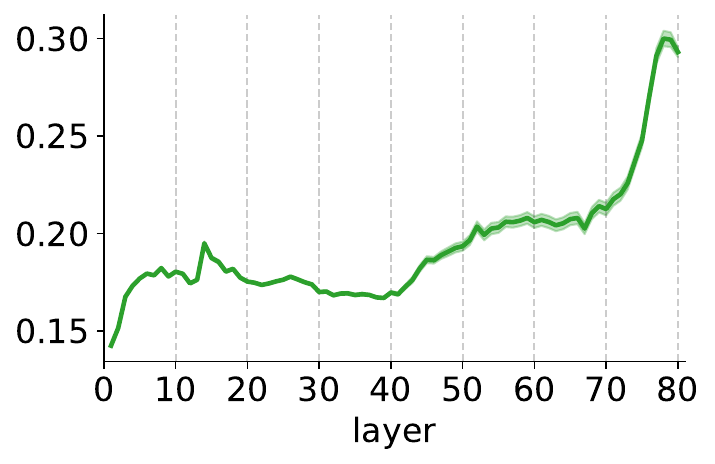}
     \end{subfigure}
     
     \caption{Figures illustrate the translation task where \llama{} 7B, 13B, and 70B are tasked with translating a word from non-English input language to output language. There is one column per model size. The x-axis shows the layer number of the model, and the y-axis the energy. Means and 95\% Gaussian confidence intervals have been computed over the input examples, numbers in \Appref{app:info}.}
     \label{fig:translation-ru-energy}
 \end{figure*}

\begin{figure*}[ht!]
     \centering
     \begin{subfigure}[b]{0.3\textwidth}
         \centering
         \includegraphics[width=\textwidth]{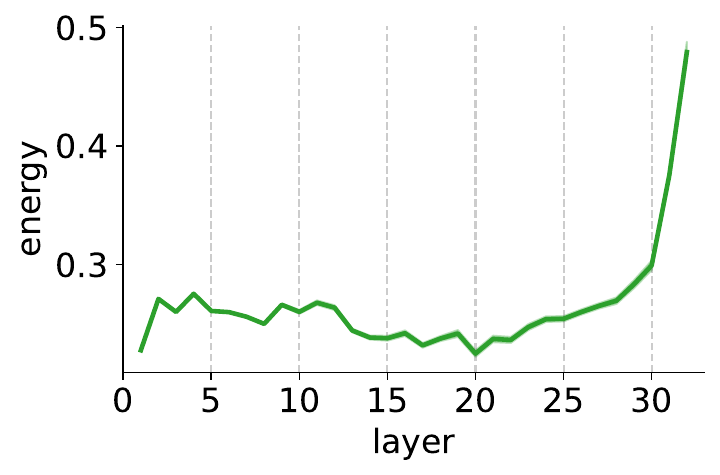}
     \end{subfigure}
     \hfill
     \begin{subfigure}[b]{0.28\textwidth}
         \centering
         \caption{Translation (ZH->DE)}
         \includegraphics[width=\textwidth]{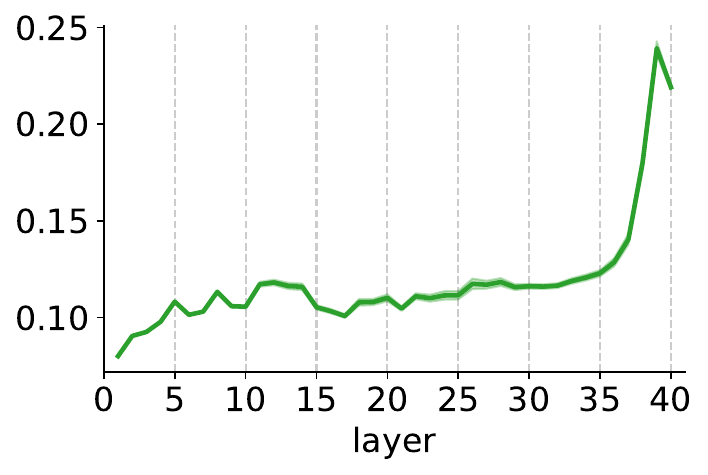}
     \end{subfigure}
     \hfill
     \begin{subfigure}[b]{0.28\textwidth}
         \centering
         \includegraphics[width=\textwidth]{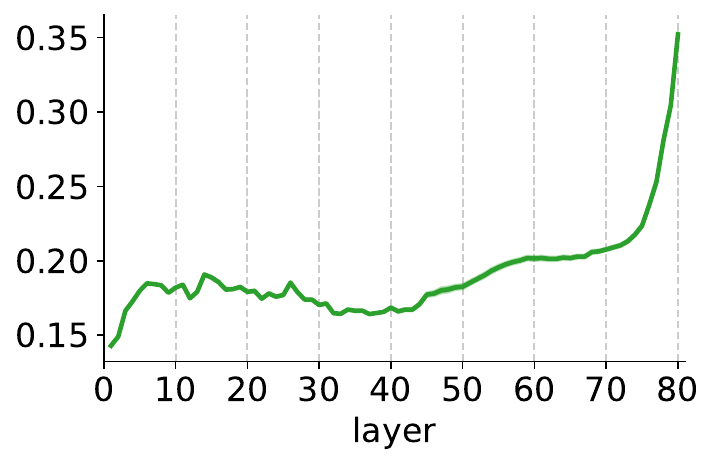}
     \end{subfigure}
     
     \begin{subfigure}[b]{0.3\textwidth}
         \centering
         \includegraphics[width=\textwidth]{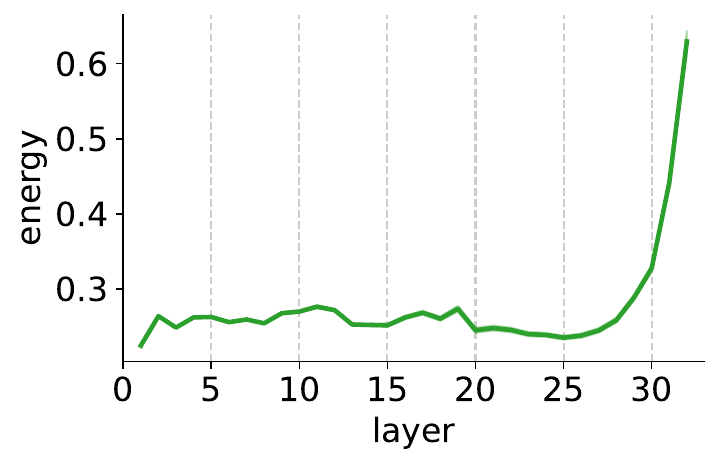}
     \end{subfigure}
     \hfill
     \begin{subfigure}[b]{0.28\textwidth}
         \centering
         \caption{Translation (ZH->EN)}
         \includegraphics[width=\textwidth]{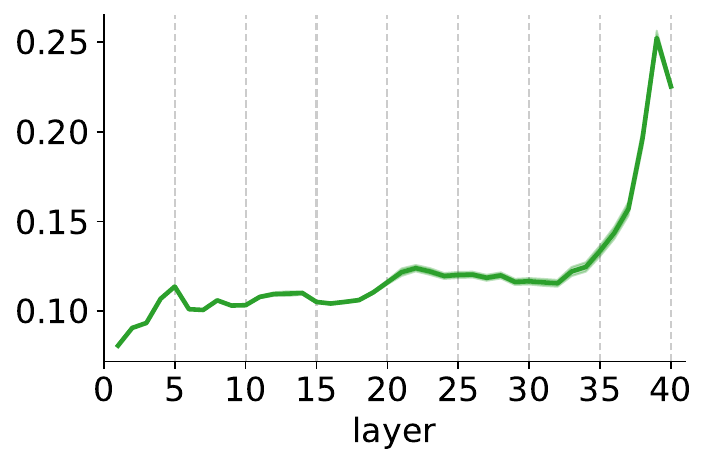}
     \end{subfigure}
     \hfill
     \begin{subfigure}[b]{0.28\textwidth}
         \centering
         \includegraphics[width=\textwidth]{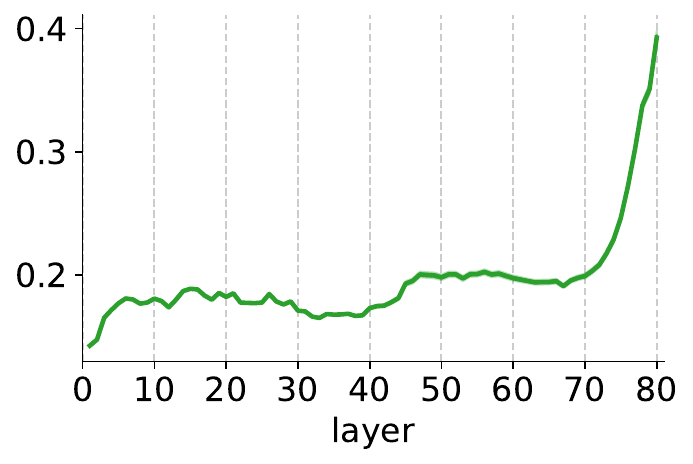}
     \end{subfigure}

     \begin{subfigure}[b]{0.3\textwidth}
         \centering
         \includegraphics[width=\textwidth]{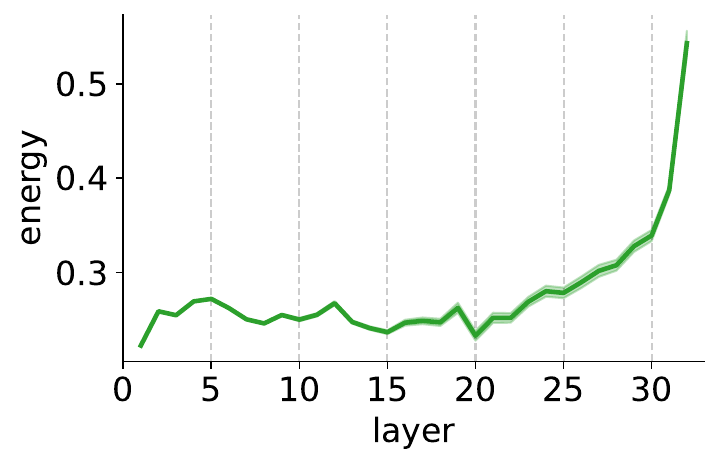}
     \end{subfigure}
     \hfill
     \begin{subfigure}[b]{0.28\textwidth}
         \centering
         \caption{Translation (ZH->FR)}
         \includegraphics[width=\textwidth]{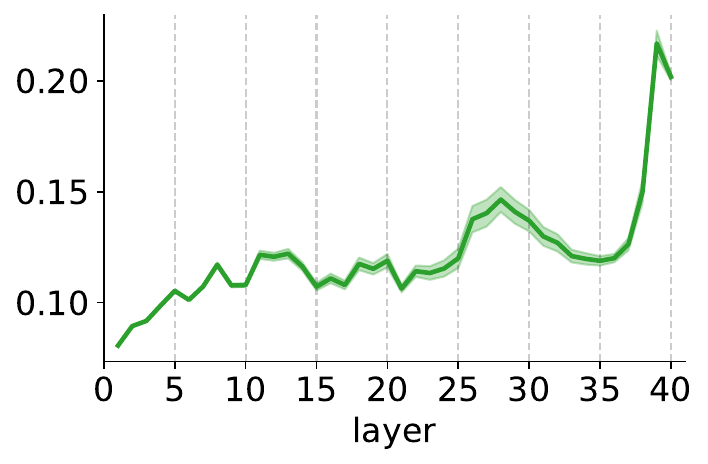}
     \end{subfigure}
     \hfill
     \begin{subfigure}[b]{0.28\textwidth}
         \centering
         \includegraphics[width=\textwidth]{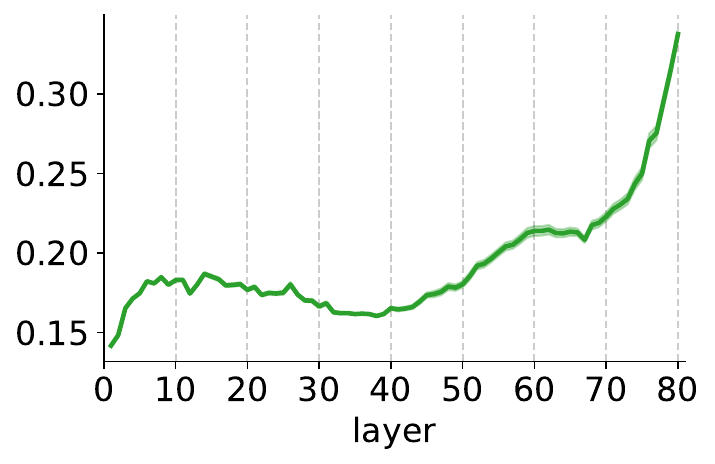}
     \end{subfigure}

     \begin{subfigure}[b]{0.3\textwidth}
         \centering
         \includegraphics[width=\textwidth]{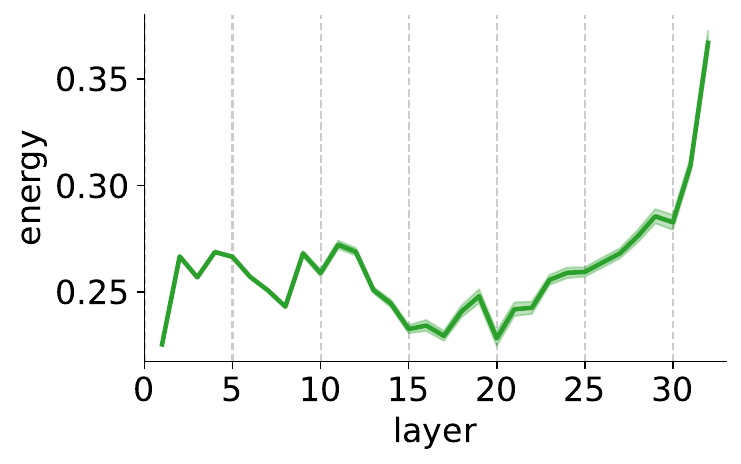}
     \end{subfigure}
     \hfill
     \begin{subfigure}[b]{0.28\textwidth}
         \centering
         \caption{Translation (ZH->RU)}
         \includegraphics[width=\textwidth]{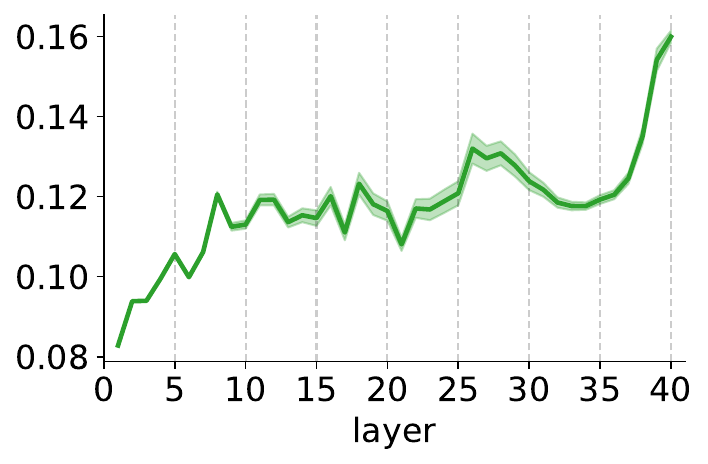}
     \end{subfigure}
     \hfill
     \begin{subfigure}[b]{0.28\textwidth}
         \centering
         \includegraphics[width=\textwidth]{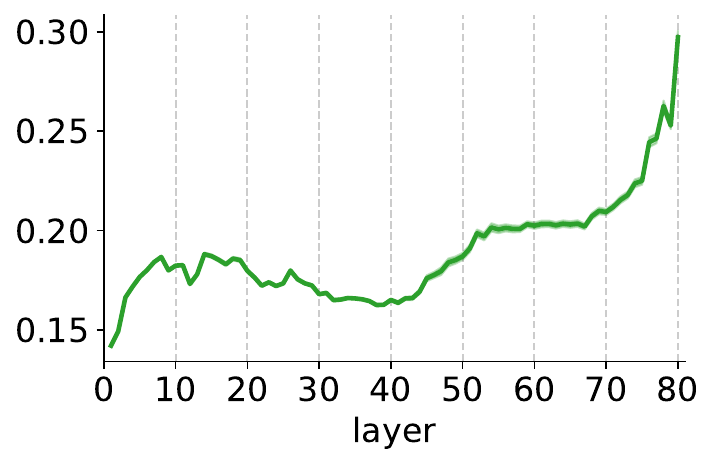}
     \end{subfigure}
     
     \caption{Figures illustrate the translation task where \llama{} 7B, 13B, and 70B are tasked with translating a word from non-English input language to output language. There is one column per model size. The x-axis shows the layer number of the model, and the y-axis the energy. Means and 95\% Gaussian confidence intervals have been computed over the input examples, numbers in \Appref{app:info}.}
     \label{fig:translation-zh-energy}
 \end{figure*}

\xhdr{Energy} Energy~(\Secref{sec:geometric view}) plots for the aggregated translation task are in Fig.~\ref{fig:translation-energy}, for the repetition task in Fig.~\ref{fig:copy-energy}, and, for the cloze task in Fig.~\ref{fig:cloze-energy}. Additionally, we provide the translation task results for individual language pairs in Fig.~\ref{fig:translation-de-energy}, Fig.~\ref{fig:translation-en-energy}, Fig.~\ref{fig:translation-fr-energy}, Fig.~\ref{fig:translation-ru-energy}, Fig.~\ref{fig:translation-zh-energy}.

Energy plots are consistent with the theory outlined in \Secref{sec:theory}.

\subsection{Low-resource language Estonian}
\label{sec:estonian}

\begin{figure*}[ht!]
    \centering
    \begin{subfigure}[b]{0.32\textwidth}
        \centering
        \caption{Copy}
        \includegraphics[width=\textwidth]{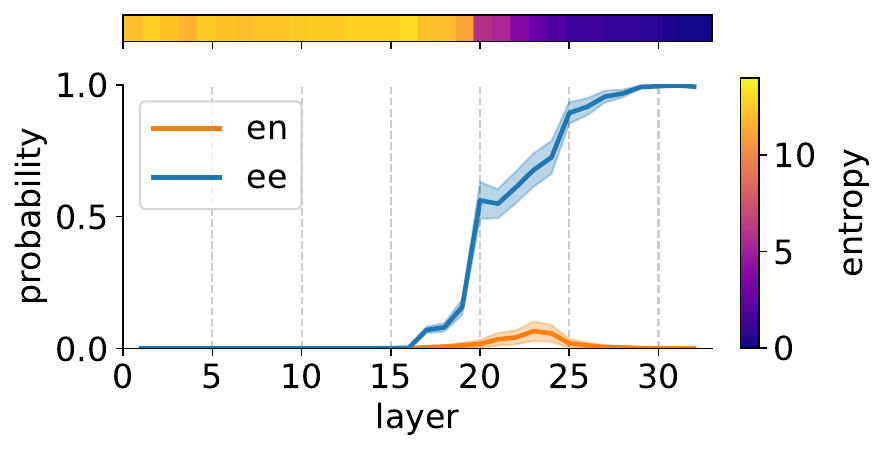}
    \end{subfigure}
    \hfill
    \begin{subfigure}[b]{0.32\textwidth}
        \centering
        \caption{Translation}
        \includegraphics[width=\textwidth]{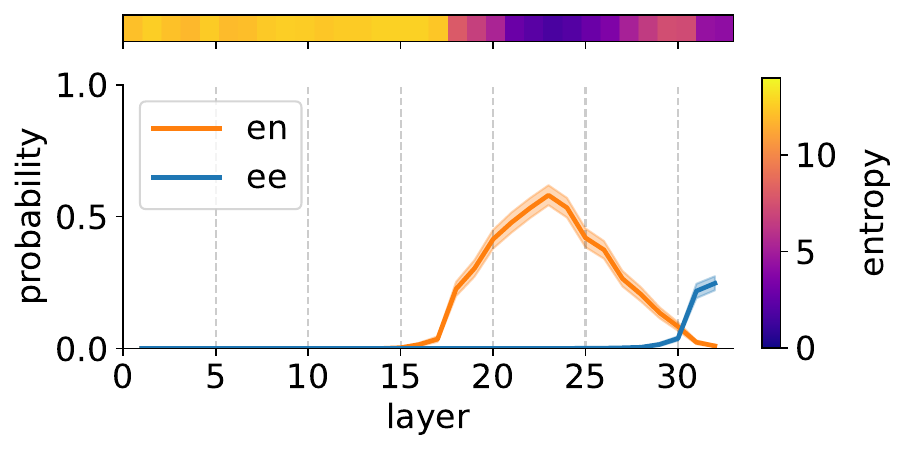}
    \end{subfigure}
    \hfill
    \begin{subfigure}[b]{0.32\textwidth}
        \centering
        \caption{Cloze}
        \includegraphics[width=\textwidth]{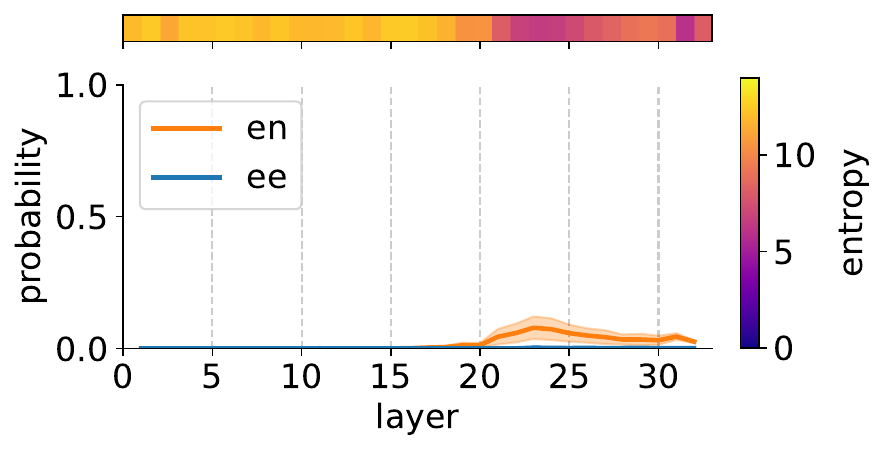}
    \end{subfigure}
    
    \begin{subfigure}[b]{0.28\textwidth}
        \centering
        \includegraphics[width=\textwidth]{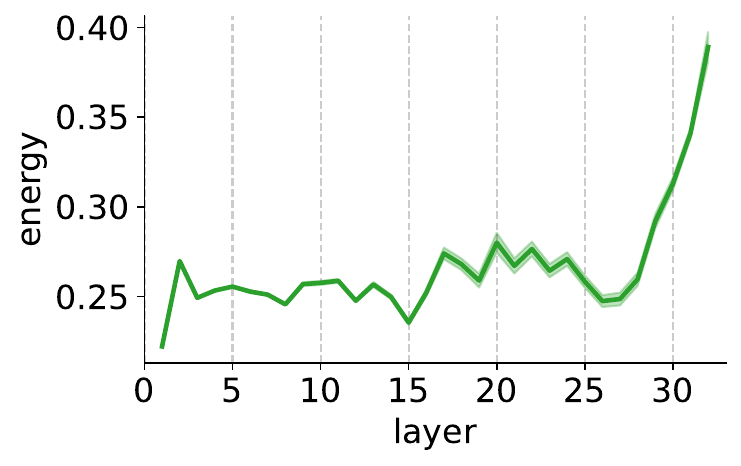}
    \end{subfigure}
    \hfill
    \begin{subfigure}[b]{0.28\textwidth}
        \centering
        \includegraphics[width=\textwidth]{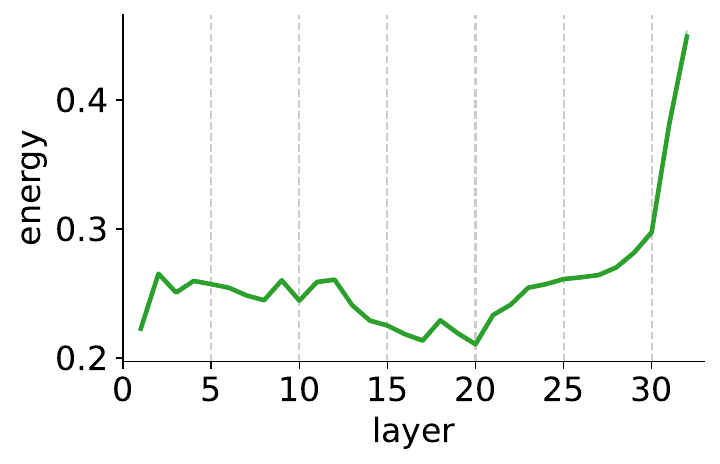}
    \end{subfigure}
    \hfill
    \begin{subfigure}[b]{0.28\textwidth}
        \centering
        \includegraphics[width=\textwidth]{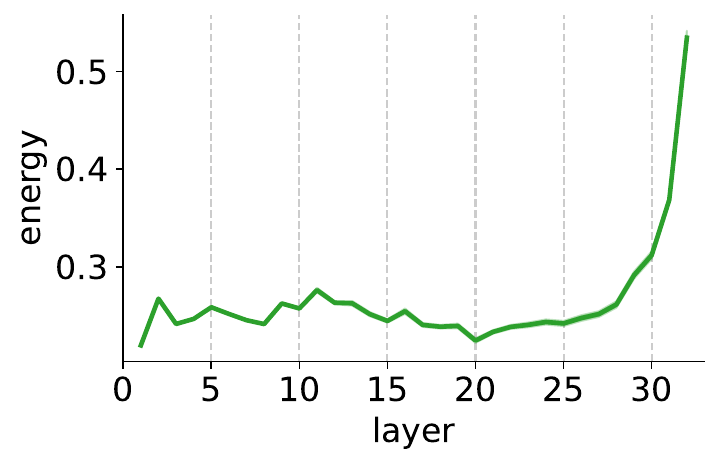}
    \end{subfigure}
    
    \caption{Figures illustrate our analysis of the copy-, translation-, and cloze task for the \textbf{Estonian} language on Llama-2-7B. In the first row, the x-axis shows the layer number of the model, and the y-axis the language probability. In the first row, the x-axis shows the layer number of the model, and the y-axis the token energy. Means and 95\% Gaussian confidence intervals have been computed over the input examples.}
    \label{fig:llama-2-7b-estonian}
\end{figure*}

We also performed our analysis with Llama-2-7B on Estonian, a low-resource language, in \Figref{fig:llama-2-7b-estonian}.
The fact that Estonian is a low-resource language is already evident in the number of single-token words: only one out of our 99 Estonian words can be represented with a single token.

\xhdr{Copy task} In the copy task, Estonian behaves the most similarly to Chinese, with the Estonian probability exceeding the English probability already in the intermediate layers.
% Previously, we speculated that for the case of Chinese this is caused by the availability of the language\hyp specific single tokens, however the result for Estonian makes us rethink this explanation.

\xhdr{Translation task} While the success probability on the translation task after the final layer is significantly smaller than in the languages studied in the main paper, we still observe the same effect as for the other languages: the intermediate next-token distributions decoded via the logit lens concentrate their probability mass on the correct English tokens and only in the final layers transition to Estonian.

\xhdr{Cloze task} The Estonian cloze task seems too hard, possibly due to the extremely low resources of Estonian in the \llama{} training data: \llama{}-7B has a 0\% success probability after the last layer. 
Interestingly, the Estonian success probability is slightly greater than 0\% in the intermediate layers, when the logit lens decodes to English. 
The success probability might increase if we included synonyms of the translated words or used human experts for the creation of the cloze examples instead of GPT-4.

\subsection{Other models: Mistral}
\label{app:mistral}

\begin{figure*}[ht!]
    \centering
    \begin{subfigure}[b]{0.32\textwidth}
        \centering
        \caption{Copy}
        \includegraphics[width=\textwidth]{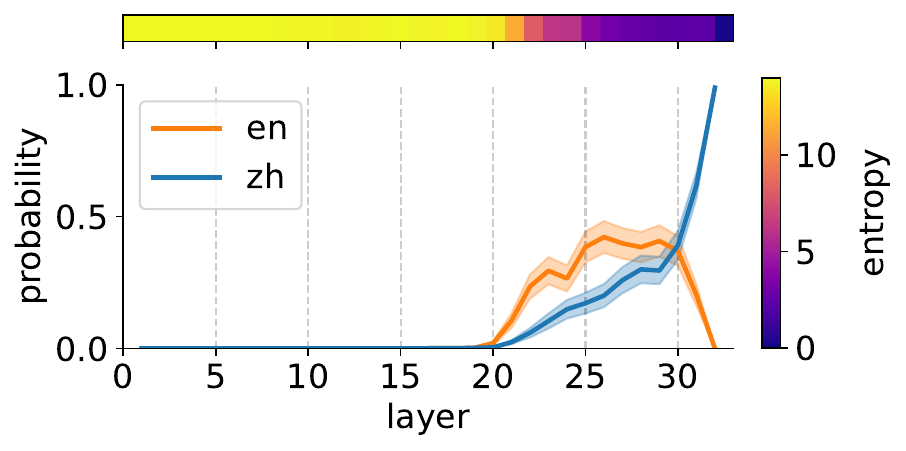}
    \end{subfigure}
    \hfill
    \begin{subfigure}[b]{0.32\textwidth}
        \centering
        \caption{Translation}
        \includegraphics[width=\textwidth]{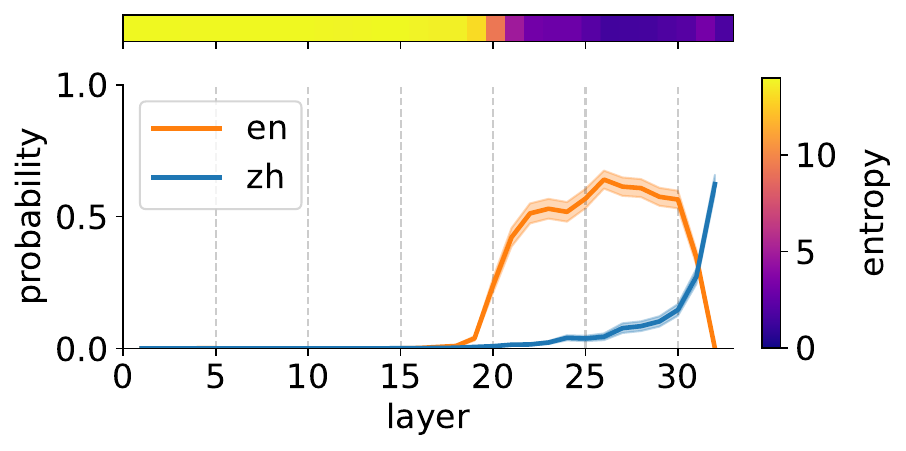}
    \end{subfigure}
    \hfill
    \begin{subfigure}[b]{0.32\textwidth}
        \centering
        \caption{Cloze}
        \includegraphics[width=\textwidth]{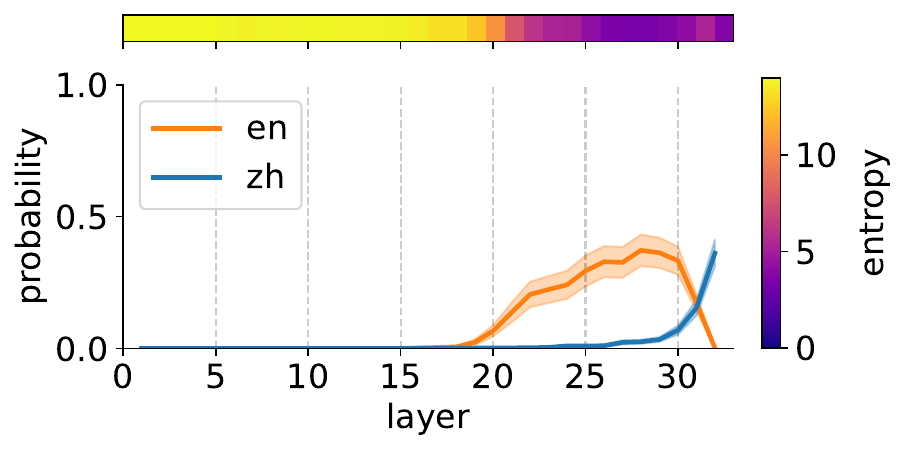}
    \end{subfigure}
    
    \begin{subfigure}[b]{0.28\textwidth}
        \centering
        \includegraphics[width=\textwidth]{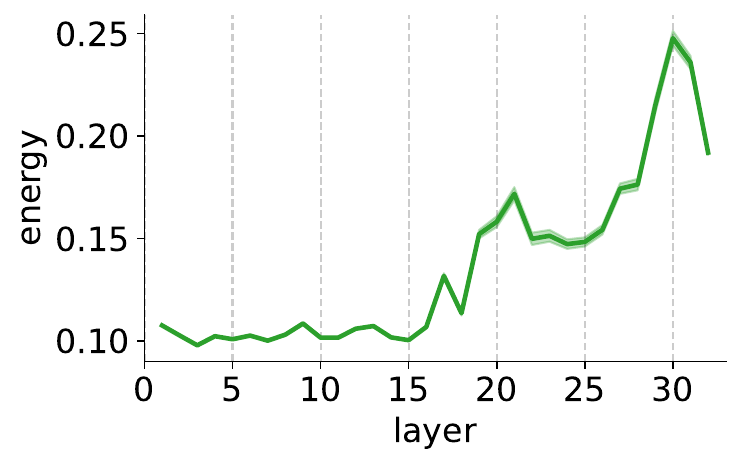}
    \end{subfigure}
    \hfill
    \begin{subfigure}[b]{0.28\textwidth}
        \centering
        \includegraphics[width=\textwidth]{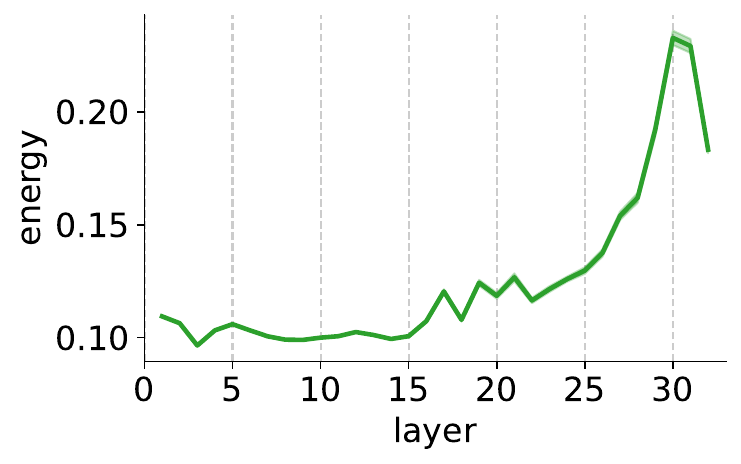}
    \end{subfigure}
    \hfill
    \begin{subfigure}[b]{0.28\textwidth}
        \centering
        \includegraphics[width=\textwidth]{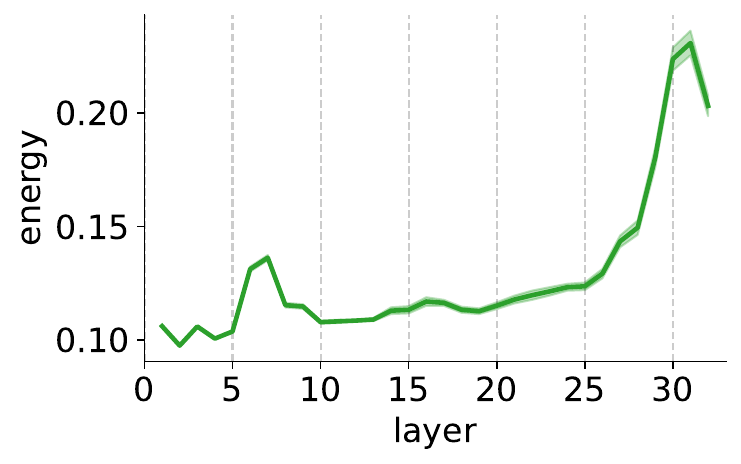}
    \end{subfigure}
    
    \caption{Figures illustrate our analysis of the copy-, translation-, and cloze task for Chinese on \textbf{Mistral-7B}. In the first row, the x-axis shows the layer number of the model, and the y-axis the language probability. In the first row, the x-axis shows the layer number of the model, and the y-axis the token energy. Means and 95\% Gaussian confidence intervals have been computed over the input examples.}
    \label{fig:mistral-7b-chinese}
\end{figure*}

We also performed our analysis on Mistral-7B, a model from outside the Llama model family. The results, shown in \Figref{fig:mistral-7b-chinese}, are consistent with those for \llama{}, pointing at the universality of our findings.

\end{document}